%% file: trs_data_driven_tracking.tex
\documentclass[lettersize,journal]{IEEEtran}
\usepackage{amsmath,amsfonts,amssymb}
\usepackage{algorithmic}
\usepackage{algorithm}
\usepackage{array}
\usepackage[caption=false,font=scriptsize,labelfont=sf,textfont=sf]{subfig}
\usepackage{textcomp}
\usepackage{stfloats}
\usepackage{url}
\usepackage{verbatim}
\usepackage{graphicx}
\usepackage{cite}
\usepackage{acronym}
\usepackage{pgf}
\usepackage{pgfplots}
\usepackage{textcomp}
\usepackage{tikz}
\usepackage{siunitx}
\usepackage{multirow}

\acrodef{cwna}[CWNA]{Constant White Noise Acceleration}
\acrodef{ekf}[EKF]{Extended Kalman Filter}
\acrodef{ew}[EW]{Electromagnetic Warfare}
\acrodef{gct}[GCT]{Gradual Coordinated Turn}
\acrodef{gp}[GP]{Gaussian Process}
\acrodef{gpr}[GPR]{Gaussian Process Regression}
\acrodefplural{gp}[GPs]{Gaussian Processes}
\acrodef{gps}[GPS]{Global Positioning System}
\acrodef{gru}[GRU]{Gated Recurrent Unit}
\acrodef{imm}[IMM]{Interacting Multiple Model}
\acrodef{kf}[KF]{Kalman Filter}
\acrodef{lstm}[LSTM]{Long Short-Term Memory}
\acrodef{map}[MAP]{Maximum A Posteriori}
\acrodef{mkf}[MKF]{Mnemonic Kalman Filter}
\acrodef{ml}[ML]{Machine Learning}
\acrodef{nsim}[NSIM]{Naturally Shift-Invariant Motion}
\acrodef{pf}[PF]{Particle Filter}
\acrodef{rib}[RIB]{Rigid Inflatable Boat}
\acrodef{rmse}[RMSE]{Root Mean Squared Error}
\acrodef{sir}[SIR]{Sampling Importance Resampling}
\acrodef{uav}[UAV]{Unmanned Aerial Vehicle}
\acrodef{ukf}[UKF]{Unscented Kalman Filter}

\pgfplotsset{compat = 1.5}
\pgfdeclarelayer{bg}    
\pgfsetlayers{bg,main}  

\definecolor{blue}{rgb}{0.20, 0.40, 0.80}		
\definecolor{green}{rgb}{0.09, 0.61, 0.49}
\definecolor{lightgreen}{RGB}{177, 200, 0}
\definecolor{orange}{rgb}{0.94, 0.58, 0.00}
\definecolor{red}{rgb}{0.80, 0.00, 0.00}
\definecolor{beige}{RGB}{255, 216, 166}
\definecolor{grey}{RGB}{193, 193, 191}
\definecolor{teal}{RGB}{0, 128, 128}
\definecolor{gold}{RGB}{255, 215, 0}

\usetikzlibrary{calc,fit,plotmarks,patterns,positioning,spy}

\def\bfC{\mathbf{C}}

\def\bfH{\mathbf{H}}
\def\bfk{\mathbf{k}}
\def\bfP{\mathbf{P}}
\def\bfQ{\mathbf{Q}}
\def\bfv{\mathbf{v}}
\def\bfV{\mathbf{V}}
\def\bfx{\mathbf{x}}
\def\bfz{\mathbf{z}}
\def\frakg{\mathfrak{g}}
\def\rmd{\mathrm{d}}

\def\x{{\mathbf x}}
\def\P{{\mathbf{{P}}}}
\def\z{{\mathbf z}}

\def\F{{\mathbf{{F}}}}
\def\Q{{\mathbf{{Q}}}}
\def\H{{\mathbf{{H}}}}
\def\R{{\mathbf{{R}}}}
\def\K{{\mathbf{{K}}}}
\def\I{{\mathbf{{I}}}}
\def\S{{\mathbf{{S}}}}

\def\p{{p}}
\def\w{{\mu}}

\def\lik{{\Lambda}}

\def\params{\boldsymbol{\theta}}

\begin{document}

\title{Data-Driven Approaches for Modelling Target Behaviour}

\author{Isabel Schlangen,~\IEEEmembership{Member,~IEEE}, André Brandenburger, Mengwei Sun, James R. Hopgood,~\IEEEmembership{Senior Member,~IEEE}
\thanks{The first two authors are with Fraunhofer FKIE, Fraunhoferstr. 20, Wachtberg, Germany. 
	The third author is with Cranfield University, Bradford, UK. 
	The last author is with the University of Edinburgh, Edinburgh, UK.
	}
\thanks{Manuscript submitted for review to IEEE Transactions on Signal Processing on October 14, 2024.}
}

\maketitle

\begin{abstract}
	The performance of tracking algorithms strongly depends on the chosen model assumptions regarding the target dynamics.
	If there is a strong mismatch between the chosen model and the true object motion, the track quality may be poor or the track is easily lost.
	Still, the true dynamics might not be known a priori or it is too complex to be expressed in a tractable mathematical formulation.
	This paper provides a comparative study between three different methods that use machine learning to describe the underlying object motion based on training data.
	The first method builds on \acp{gp} for predicting the object motion, the second learns the parameters of an \ac{imm} filter and the third uses a \ac{lstm} network as a motion model.
	All methods are compared against an \ac{ekf} with an analytic motion model as a benchmark and their respective strengths are highlighted in one simulated and two real-world scenarios.
\end{abstract}

\acresetall


\section{Introduction}
Conventional single-target tracking usually builds on the paradigm of Bayesian filtering, which comprises the prediction across time and the correction of the predicted belief using an incoming measurement. 
The performance of the Bayes filter is subject to the chosen mathematical models for the target dynamics and the properties of the sensor in use.
In the majority of applications, the sensor characteristics are known a priori or can be calibrated in preliminary experiments to minimise a possible model mismatch.
The dynamics of the object of interest, on the other hand, might be hard to formalise in advance if the object type is unclear or its physics are not well-understood. 
In many cases, well-known object dynamics such as \ac{cwna} \cite{koch2016tracking} are chosen for convenience, compensating small model inaccuracies by appropriate amounts of process noise.
Unfortunately, a grave mismatch between the true target behaviour and the chosen model leads to poor estimation results or even track loss.

One option for a more versatile description of the target motion is to involve more than one model, for example in an \ac{imm} approach. 
The parameters of the \ac{imm} method are usually predefined but could also be learned as proposed in \cite{brandenburger2023learning}. 
A traditional method for filter parameter optimization is {expectation maximization}, which involves successive calculation of the expected parameter likelihood and its maximization~\cite{lei2006expectation,huang2004imm,huang2005expectation}. 
This typically requires complex analytical calculations, which may need adjustments if the model or sensor architecture changes.
Barrat et al.~\cite{barratt2020fitting} apply convex optimization tools to train Kalman smoother parameters~\cite{barratt2020fitting}. 
While simpler, this method also relies on analytical calculations, which can be challenging for complex architectures. 
Abbeel et al.~\cite{Abbeel2005} demonstrated training an \ac{ekf} using coordinate ascent, and Greenberg et al.~\cite{Greenberg2021} utilized gradient descent for Kalman filter optimization. 
These methods can be trained on ground-truth data or the measurement likelihood, outperforming hand-tuned parameters in real systems~\cite{Abbeel2005}.
Xu et al.~\cite{xu2021ekfnet} introduced {EKFNet}, fitting \ac{ekf} parameters using gradient descent akin to neural network optimization. 
However, this method requires problem-specific parameter regularization functions. 
Furthermore, Coskun et al.~\cite{coskun2017long} proposed generating Kalman filter matrices online using \ac{lstm} networks, outperforming traditional filters and \ac{lstm} networks. 

Instead of optimising the filter parameters with machine learning, it is possible to replace parts of the Bayes recursion with a neural network.
A recent work by Liang and Meyer \cite{liang2023neural} uses belief propagation to enhance measurement association for multi-target tracking problems.
Several research groups have implemented transformer-based data association to include temporal information in the measurement update \cite{pinto2023deep,wei2023transformer,wenna2022multitarget}.
In all of these methods, however, the target dynamics are assumed to follow a \ac{cwna}-like motion, which cannot cope well with highly manoeuvring targets.
In contrast, \cite{jung2019sequential} introduced a particle filter that performs the temporal transition with a trained \ac{lstm} network.
Following this, a Kalman filter has been equipped with a similar \ac{lstm} architecture in the prediction step, leading to the \ac{mkf}.

A parallel approach was developed by Sun et al. \cite{sun2020gaussian, sun2022gaussian}, which builds on \acp{gp} rather than recurrent networks to model the state transition.
All of these methods have in common that the motion model is found based on a set of suitable training data, hence no prior knowledge on the physics of the target of interest is necessary.
This paper, in contrast, focuses on data-driven methods to describe object motion in a single-target Bayesian framework.
The advantage of data-driven models is that they solely depend on the availability of a suitable set of training data, from which the characteristic properties of the underlying target dynamics are derived. 
This is especially advantageous in cases with a rapid change of motion behaviour or very complex dynamics that are not easily described in an analytic formulation. 

To showcase different data-driven motion models, variations of the three approaches in \cite{sun2022gaussian,brandenburger2023learning,jung2020mnemonic} are presented and evaluated.
In Sec.~\ref{sec:bayes}, the Bayesian single-target filter is described as a foundation for the discussed approaches.
After that, the \ac{gp}-based particle filter is presented in Sec.~\ref{sec:gp}, followed by the optimised \ac{imm} filter in Sec.~\ref{sec:imm} and lastly the \ac{lstm}-based \ac{ekf} in Sec.~\ref{sec:mkf}.
The three methods are then compared against a baseline \ac{ekf} in three informative experiments in Sec.~\ref{sec:comparison}: 
The first dataset was created in simulation and hence provides full control over all model parameters, while the second and third datasets are real-world \ac{gps} measurements of two different cooperative targets.
It is demonstrated that each method has its situation-specific advantages and that both analytic and data-driven dynamic models are highly relevant for modern tracking applications (see concluding remarks in Sec.~\ref{sec:conclusion}).

\section{Single-Target Bayesian Filtering}
\label{sec:bayes}
This section summarises the general concept of Bayesian single-target tracking. 
In this paper, a discrete-time framework is assumed, where individual time steps are denoted with integer-valued indices $t\in N$.  
The target state at time $t$ will be written as a $d_x$-dimensional vector $\bfx_t$. 
In a similar manner, the $d_z$-dimensional measurement received at time $t$ is represented by $\bfz_t$, where $\bfz_{1:t}=(\bfz_1,\dots,\bfz_t)$ is the collection of measurements up to time $t$. 
A general method to estimate the temporal evolution of a target state is through the Bayes recursion. 
It consists of two main steps: 

The \textbf{Chapman-Kolmogorov equation} propagates the posterior distribution $p_t(\bfx_t|\bfz_{1:t})$ of the current target state $\bfx_t$ given a sequence of past measurements $\bfz_{1:t}$ to time $t+1$, arriving at the \textit{predicted distribution}
\begin{align}
		&p_{t+1|0:t}(\bfx_{t+1}|\bfz_{1:t})\nonumber\\
		&= \int f_{t+1|0:t} (\bfx_{t+1} |\bfx_{0:t},\bfz_{1:t}) p_t(\bfx_{0:t}|\bfz_{1:t})\rmd\bfx_{0:t} \label{eq:chapman_nonmarkov}\\
		&\approx \int f_{t+1|0:t}(\bfx_{t+1}|\bfx_{0:t},\bfz_{1:t}) p_t(\bfx_t|\bfz_{1:t})\rmd\bfx_{0:t},\label{eq:chapman_nonmarkov_approx}
\end{align}
where $f_{t+1|0:t}$ denotes the transition function to time ${t+1}$ given the previous history $0:t$. 
The approximation in \eqref{eq:chapman_nonmarkov_approx} assumes that the current target distribution $p_t$ is independent of the previous target states $\bfx_0,\dots, \bfx_{t-1}$.

In the update step, a new sensor measurement $\bfz_{t+1}$ is received at time ${t+1}$ and is associated with $\bfx_{t+1}$ according to the likelihood function $g_{t+1} (\bfz_{t+1} |\bfx_{t+1})$. 
The updated state distribution $p_{t+1} (\bfx_{t+1} |\bfz_{1:t+1})$ is calculated via \textbf{Bayes’ rule}:
\begin{equation}
	p_{t+1} (\bfx_{t+1}|\bfz_{1:t+1})\!=\!\frac{p_{t+1|0:t}(\bfx_{t+1}|\bfz_{1:t})g_{t+1} (\bfx_{t+1}|\bfz_{t+1})}{\int p_{t+1|0:t} (\bfx|\bfz_{1:t}) g_{t+1}(\bfx|\bfz_{t+1})\rmd\bfx} .
\end{equation}

In many cases, the transition function is assumed to be Markovian for the sake of simplicity, i.e. the future target state $\bfx_{t+1}$ only depends on the present state $\bfx_t$. 
Under this assumption, the Bayes recursion reduces to
\begin{align}
	&p_{t+1|t}(\bfx_{t+1}|\bfz_{1:t})\!=\!\!\int\!\!\! f_{t+1|t}(\bfx_{t+1}|\bfx_t,\bfz_{1:t}) p_t(\bfx_t|\bfz_{1:t})\rmd\bfx_t,\\
	&p_{t+1}(\bfx_{t+1}|\bfz_{1:t+1})=\frac{p_{t+1|t}(\bfx_{t+1}|\bfz_{1:t})g_{t+1} (\bfz_{t+1}|x_{t+1})}{\int p_{t+1|t}(\bfx|\bfz_{1:t})g_{t+1}(\bfz_{t+1}|\bfx)\rmd\bfx}.
\end{align}

Bayesian single-target tracking can be categorised into two main approaches: model-based filters and data-driven filters. 
Model-based filters, exemplified by the \ac{kf} and the \ac{pf}, rely on a predefined mathematical model to estimate the target state. 
On the other hand, data-driven filters leverage \ac{ml} techniques to directly learn the mapping from observations to target states, without explicit modelling of the system dynamics. 
By incorporating \ac{ml} techniques, data-driven filters have the potential to outperform traditional model-based filters in scenarios with highly non-linear dynamics or complex sensor measurements. 
In the following sections \ref{sec:gp}-\ref{sec:mkf}, we describe three distinct \ac{ml} architectures, i.e., \ac{gp}-, \ac{imm}- and \ac{lstm}-based filters.

\section{The Gaussian Process Model}
\label{sec:gp}
The \ac{gp} model is a popular tool for Bayesian non-linear regression, providing a robust framework for modelling complex relationships in data. 
In a \ac{gp} model, a training dataset and a kernel function are used to generate a joint Gaussian distribution that characterises the values of a function at specified points \cite{koch2016tracking}. 
One of the key advantages of \ac{gp} models is their ability to provide confidence intervals for predictions, offering a reliable estimate of uncertainty. 
In this section, we present an overview of \ac{gpr} theory and introduce a \ac{gp}-based method for learning the motion behaviour of a target, as proposed in \cite{brandenburger2023learning}.

\subsection{Gaussian Process Regression}
Consider a general \ac{gpr} problem involving noisy observations from an unknown function described as:
\begin{equation}
	\bfz=g(\bfx)+\bfv,~~\bfv\sim\mathcal{N}(0,\sigma_\bfv^2I)  
\end{equation}
where $\sigma_\bfv^2$ is the noise variance. 
The training dataset of size $N$ is denoted as $D\!=\!\{X,Z\}$, where $X\hspace{-0.1cm}=\![\bfx_1,\bfx_2,\dots,\bfx_N]$ are the inputs and $Z\!=\![\bfz_1,\bfz_2,\dots,\bfz_N]^T$ the corresponding outputs. 

In \ac{gpr}, the latent distribution of $\frakg(\bfx_t^*)$ at the test point $\bfx_t^*$, denoted as $\frakg_t$, is assumed to follow a Gaussian distribution:
\begin{equation}
	\frakg_t |D \sim \mathcal{N}(\mu(g_t),\Sigma(g_t)),
\end{equation}
where
\begin{subequations}
	\begin{align}
		\mu(\frakg_t)&=m(\bfx_t^*)+\bfk_*^T [K+\sigma^2 I_N]^{-1}[\bfz-m(X)], \label{eq:gp_latentmu}\\
		\Sigma(\frakg_t)&=\bfk_{**}-\bfk_*^T [K+\sigma^2 I_N]^{-1} \bfk_*.  \label{eq:gp_latentsigma}
	\end{align}
\end{subequations}
Here, $K\triangleq k(X,X)$ is the kernel matrix evaluated at training inputs, $\bfk_*\triangleq k(X,\bfx_t^*)$ represents the covariance between training inputs and the test point, and $\bfk_{**}\triangleq k(\bfx_t^*,\bfx_t^*)$ is the covariance at the test point. 
The utilised kernel function $k(\cdot,\cdot)$ is the squared exponential covariance function:
\begin{equation}
	k(\bfx, \bfx')=\sigma_0^2 \exp\left[-\frac{1}{2} \frac{(\bfx-\bfx')^T (\bfx-\bfx')}{l^2} \right]
\end{equation}
where $\sigma_0^2$ and $l^2$ are hyperparameters representing the prior variance of the signal amplitude and the length scale of the kernel function, respectively.

\subsection{GP-based Motion Behaviour Learning — Limited Information of the Training State}
Sun et al. \cite{koch2016tracking} introduced a pioneering \ac{gp}-based learning paradigm to capture the \ac{nsim} characteristics exhibited by targets. 
This method focuses on utilising \acp{gp} to model \ac{nsim} behaviour, specifically targeting Cartesian velocities derived from Cartesian positions. 
In this paper, we consider tracking scenarios where \ac{nsim} characteristics are captured by speed (the Euclidean norm of Cartesian velocities) and heading (computed with the atan2 function).
The hierarchical model is described as 
\begin{subequations}
	\begin{align}
		s_{t+1}&= s_t+ s_t\cos(\phi_t) + v_t^s \label{eq:gp_nsim_xi},\\
		\phi_{t+1}&= \phi_t+ s_t\sin(\phi_t)  + v_t^\phi \label{eq:gp_nsim_eta},\\
		s_t&= f_t^s (\mathbf{u}_{t-1}) \label{eq:gp_nsim_delta_xi},\\
		\phi_t&= f_t^\phi (\mathbf{u}_{t-1}) \label{eq:gp_nsim_delta_eta}.
	\end{align}
\end{subequations}
In this approach, the Cartesian speed and heading, expressed as $s_t\!\!=\!\!f_t^s (\mathbf{u}_{t-1})$ and $\phi_t\!\!=\!\!f_t^\phi (\mathbf{u}_{t-1})$, are learned as two~\acp{gp} $\mathcal{GP}_s$ and $\mathcal{GP}_\phi$, respectively. 
Here, the \ac{gp} input is $\mathbf{u}_{t-1} \triangleq [s_{t-1}, \phi_{t-1}]^{\mathrm{T}}$.
By integrating both Cartesian velocities as inputs for the \acp{gp}, this method aims to capture the intertwined nature of target motion across coordinates, thereby encapsulating vital characteristics of target behaviour during manoeuvres. 

\begin{figure}[!t]
	\centering
	\input{figures/gp.tikz}
	\caption{The factor graph of the proposed joint GP algorithm for state prediction mainly includes offline training and online prediction processes. Legend: Circles – variable nodes; Squares – factor nodes. The process for the prediction on the X-axis is marked in blue.}
	\label{fig:gp}
\end{figure}
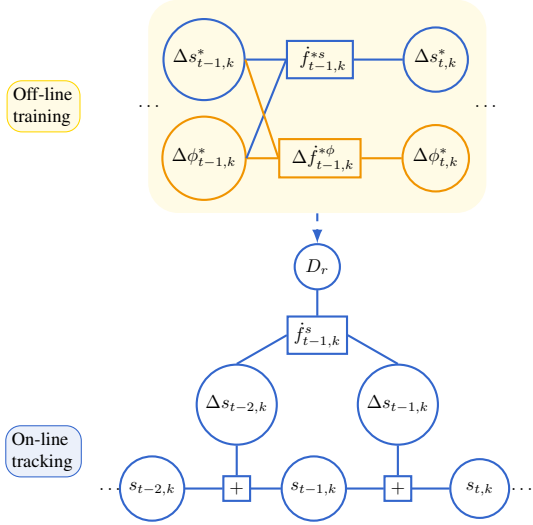

The proposed \ac{gp}-based approach for prediction comprises two pivotal phases: offline training and online prediction.
During training, historical data is harnessed to glean the statistical intricacies of target motion behaviours. 
The proposed \ac{gp}-based X-axis position prediction is shown in Fig.~\ref{fig:gp}, the Y-axis position prediction follows likewise. 
Specifically, the training data sets are $D_s\!=\!\{U,\mathbf{s}\}$ and $D_\phi\!=\!\{U,{\Phi}\}$, where the training input for both \acp{gp} are the same $U\!=\![\mathbf{u}_1, \mathbf{u}_2,\dots\mathbf{u}_{N-1}]$, and the training output for $\mathcal{GP}_s$ and $\mathcal{GP}_\phi$ are $\mathbf{s}\!=\![s_2, s_3,\dots, s_N]^T$ and $\Phi\!=\![\phi_2, \phi_3,\dots, \phi_N]^T$, respectively. 

Subsequently, these insights are extrapolated to facilitate online tracking and state prediction in real-time scenarios by integrating them in the \ac{sir} \ac{pf} framework \cite{doucet2001sequential}. 

\subsection{Target Tracking with Learned NSIM Model}
The \ac{sir} \ac{pf} is used for target tracking, including drawing particles, updating weights, normalising and \ac{map} estimation \cite{doucet2001sequential}. 
Specifically, the particles at time $t-1$ include the position particles $\bfx_{t-1}^{\{m\}}=[\xi_{t-1}^{\{m\}},\eta_{t-1}^{\{m\}}]^T$ and Cartesian velocity particles $\Delta \bfx_{t-2}^{\{m\}}=[\Delta\xi_{t-2}^{\{m\}},\Delta\eta_{t-2}^{\{m\}}]^T$ with weights $w_{t-1}^{\{m\}}, m=1:M$, and $M$ is the size of the particle set. 
At time $t$, the learned \ac{nsim} model is used to draw Cartesian velocity particles $\Delta \bfx_{t-1}^{\{m\}}=[\Delta\xi_{t-1}^{\{m\}},\Delta\eta_{t-1}^{\{m\}}]^T$ with
\begin{subequations}
	\begin{align}
		\Delta\xi_{t-1}^{\{m\}} &\sim p(f_t^\xi (\Delta \bfx_{t-2}^{\{m\}})|D_\xi) \nonumber \\
		&=\mathcal{N}(\mu_t^\xi(\Delta x_{t-2}^{\{m\}}),\Sigma_t^\xi (\Delta \bfx_{t-2}^{\{m\}})), \\
		\Delta\eta_{t-1}^{\{m\}}&\sim p(f_t^\eta (\Delta \bfx_{t-2}^{\{m\}}|D_\eta)\nonumber \\
		&=\mathcal{N}(\mu_t^\eta (\Delta x_{t-2}^{\{m\}}),\Sigma_t^\eta (\Delta \bfx_{t-2}^{\{m\}})).
	\end{align}
\end{subequations}
Here, $\mu_t^\xi (\Delta \bfx_{t-2}^{\{m\}})$ and $\mu_t^\eta(\Delta \bfx_{t-2}^{\{m\}})$ can be calculated via~\eqref{eq:gp_latentmu}, while $\Sigma_t^\xi (\Delta \bfx_{t-2}^{\{m\}})$ and $\Sigma_t^\eta (\Delta \bfx_{t-2}^{\{m\}})$ are found using~\eqref{eq:gp_latentsigma}. 
Then, the position particles $\bfx_t^{\{m\}}\hspace{-0.1cm}=\![\xi_t^{\{m\}},\eta_t^{\{m\}}]^T$ are generated according to the motion model in \eqref{eq:gp_nsim_xi} and \eqref{eq:gp_nsim_eta}, i.e., 
\begin{subequations}
	\begin{align}
		\xi_t^{\{m\}}&=\xi_{t-1}^{\{m\}}+\Delta\xi_{t-1}^{\{m\}}+v_{t-1}^{\xi,\{m\}} \\ 
		\eta_t^{\{m\}}&=\eta_{t-1}^{\{m\}}+\Delta\eta_{t-1}^{\{m\}}+v_{t-1}^{\eta,\{m\}} 
	\end{align}
\end{subequations}
The weights are then updated as $\tilde{w}_t^{\{m\}}\approx w_{t-1}^{\{m\}} p(z_t|\bfx_t^{\{m\}})$, normalised and resampled. 
The \ac{map} estimations of the target’s position and velocity at time $t$ are calculated as:
\begin{subequations}
	\begin{align}
		\hat{\bfx}_t &= \sum_{m=1}^M w_t^{\{m\}} \bfx_t^{\{m\}} \\
		\Delta\hat{\bfx}_{t-1} &= \sum_{m=1}^M w_t^{\{m\}} \Delta \bfx_{t-1}^{\{m\}}  
	\end{align}
\end{subequations}
In conclusion, the proposed \ac{gp}-based method can learn target motion behaviour, particularly suited for scenarios with limited information of the training state and offer a robust framework for prediction and tracking.

\section{Optimization of \acs{imm} Filter Parameters Using Gradient Descent}
\label{sec:imm}
\subsection{\acf{imm} Filter}
The \ac{imm} filter is an advanced form of the Kalman filter designed to efficiently handle multiple manoeuvring models within a unified framework. 
This capability is essential for tracking applications where the target may switch between different modes of motion, each with distinct dynamics. 
The \ac{imm} filter operates by maintaining multiple filter instances, each corresponding to a different motion model, and seamlessly switching between these models based on the likelihood of the observed measurements. 
A detailed explanation of the filter can be found in~\cite{bar2004estimation}. 

Each model in the \ac{imm} framework, referred to as a mode, is equipped with its own set of parameters, including the state transition matrix \( \F^{i} \) and the process noise covariance \( \Q^{i} \). 
The measurement model, however, is typically shared across modes, characterized by the measurement matrix \( \H \) and the measurement noise covariance \( \R \). 
Similar to the \ac{ekf}, the dynamics and measurement matrices may be linearisations of potentially non-linear dynamics \( f \) and measurement function \( h \).
The state dynamics and measurement process for each mode are represented as follows:
\begin{align}
	\x_{t} &= f^{i}(\x_{t-1}) + \mathbf{w}_{t}^{i}, & \mathbf{w}_{t}^{i} \sim \mathcal{N}(0, \Q^{i}), \\
	\z_t &= h(\x_t) + \mathbf{v}_t, & \mathbf{v}_t \sim \mathcal{N}(0, \R),
\end{align}
with mode index \( i \), the mode-specific state transition function \( f^{i} \), process noise \( \mathbf{w}_t^{i} \) with covariance \( \Q^{i} \), measurement function \( h \) and measurement noise \( \mathbf{v}_t \) with covariance \( \R \).

The \ac{imm} filter cycles through a set of operational phases -- model mixing, mode-matched filtering, and mode probability updating -- to compute the combined state estimate. 
These phases are detailed as follows:

\textbf{Mixing Step:} The initial state and covariance estimates for each mode are computed by mixing the estimates from the previous step based on the mode transition probabilities \( \p^{ij} \), which represent the probability of switching from mode \( i \) to~\( j \):
\begin{align}
	\w_{t\vert t-1}^j &= \sum_{i=1}^{m} \p^{ij}\w_{t-1 \vert t-1}^i \label{eq:cbar}\\
	\w_{t-1 \vert t-1}^{i \vert j} &= \p^{ij}\frac{\w_{t-1 \vert t-1}^i}{\w_{t\vert t-1}^j}  \\ 
	\x_{t-1 \vert t-1}^{0j} &= \sum_{i=1}^{m} \w_{t-1 \vert t-1}^{i \vert j}\x_{t-1 \vert t-1}^i 
\end{align}
\begin{gather}
	\x_{t-1}^\text{diff} = \x_{t-1 \vert t-1}^i - \x_{t-1 \vert t-1}^{0j}\\
	\P_{t-1 \vert t-1}^{0j} = \sum_{i=1}^{m} \w_{t-1 \vert t-1}^{i \vert j} \Big[ \P_{t-1 \vert t-1}^i \!+\! \x_{t-1}^\text{diff}\big(\x_{t-1}^\text{diff}\big)^T \Big]
\end{gather}

\textbf{Mode-Matched Filtering:} Each mode now processes the current measurement to update its state estimate and covariance using the standard Kalman filter equations:
\begin{align}
	\F_t^{0j} &= \frac{\partial}{\partial \x} f^j\big(\x_{t-1 \vert t-1}^{0j}\big) \label{eq:F}\\
	\x_{t \vert t-1}^{j} &= f^j\big(\x_{t-1 \vert t-1}^{0j}\big) \\
	\P_{t \vert t-1}^{j} &= \F_t^{0j} \P_{t-1 \vert t-1}^{0j} \big(\F_t^{0j}\big)^T + \Q^j \\
	\H_t^{j} &= \frac{\partial}{\partial \x} h\big(\x_{t \vert t-1}^{j}\big) \label{eq:H}\\
	\S_{t}^{j} &= \H_t^{j} \P_{t \vert t-1}^{j} \big(\H_t^{j}\big)^T + \R \label{eq:S}\\
	\K_{t}^{j} &= \P_{t \vert t-1}^{j}\big(\H_t^{j}\big)^T\big(\S_{t}^{j}\big)^{-1}
\end{align}
\begin{gather}
	\x_{t \vert t}^j = \x_{t \vert t-1}^{j} + \K_{t}^{j}\big(\z_{t}-h(\x_{t \vert t-1}^{j})\big) \\
	\P_{t \vert t}^j = \big(\I - \K_{t}^j\H_t^j\big)\!\P_{t \vert t-1}^j\!\big(\I - \K_{t}^j\H_t^j\big)^T \!\!+ \K_{t} ^j \R \big(\K_{t}^j\big)^T\label{eq:joseph_form}
\end{gather}

\textbf{Mode Probability Update:} The mode probabilities are updated based on the likelihood of the current measurement for each mode:
\begin{gather}
	\lik_t^j = \mathcal{N}\left(\z_t; h(\x^{0j}_{t \vert t-1}), \S^{0j}_t\right) \label{eq:lambda} \\ 
	\w_{t \vert t}^j = \frac{1}{c} \lik_t^j\w_{t\vert t-1}^j \label{eq:wfiltered} \qquad
	c = \sum_{j=1}^{m} \lik_t^j\w_{t\vert t-1}^j
\end{gather}

\textbf{State Combination:} Finally, the \ac{imm} filter combines the estimates from all modes to form a single state estimate and covariance:
\begin{gather}
	\x_{t \vert t} = \sum_{j=1}^{m} \w_{t \vert t}^j \x_{t \vert t}^j \label{eq:matchstate} \\ 
	\begin{split}
		\P_{t \vert t} = \sum_{j=1}^{m} \w_{t \vert t}^j \Big( \P_{t \vert t}^j + \left[\x_{t \vert t}^j - \x_{t \vert t}\right] \left[\x_{t \vert t}^j - \x_{t \vert t}\right]^T\Big)
	\end{split} \label{eq:matchcov}
\end{gather}

This multi-model approach allows the \ac{imm} filter to adaptively manage the uncertainties associated with various motion patterns. 
The next section delves into the gradient descent-based optimization strategy used to refine the \ac{imm} filter parameters, leveraging sensor data to enhance filter performance.

\subsection{Parameter Optimization Using Gradient Descent}
To enhance the performance of the \ac{imm} filter in practical scenarios where ground truth data might not be readily available, it is essential to optimize the filter parameters effectively. 
This subsection discusses the application of gradient descent, a widely used optimization algorithm, to refine the parameters of the \ac{imm} filter based on measurement data alone, as originally described by Brandenburger et al.~\cite{brandenburger2023learning}.

The objective of parameter optimization in this context is to minimize a loss function that quantifies the discrepancy between the predicted measurements by the filter and the actual measurements. 
The chosen loss function for this optimization is the negative log-likelihood of the measurement sequence given the model parameters~\cite{brandenburger2023learning}.

Given the non-linear nature of the measurement function and the system dynamics in certain modes, the \ac{imm} filter parameters are updated using gradient descent with backpropagation, which is capable of handling complex derivatives in a computationally efficient manner. 
For the linear case, the parameters to be optimized, \( \params \), may include the mode transition probabilities \(\p^{ij}\), the mode-specific process noise covariances \(\Q^{i}\), and the measurement noise covariance \(\R\). 
If the dynamics or measurement functions are non-linear, any parameter \( \params \) of the dynamics or measurement model can be optimized as long as the model is differentiable with respect to the chosen \(\params\).

The parameter update rule via gradient descent is defined~as
\begin{equation}
	\params_{k+1} = \params_k - \eta \nabla \mathcal{L}(\params_k),
\end{equation}
where \(\params_k\) denotes the parameter vector at iteration \(k\), \(\eta\) is the learning rate, and \(\nabla \mathcal{L}(\params_k)\) represents the gradient of the loss function with respect to the parameters at iteration \(k\).

The loss function, \(\mathcal{L}\), specifically tailored for \ac{imm} parameter optimization, is formulated as:
\begin{equation}
	\mathcal{L}(\params) = -\sum_{t=1}^{T} \log p(\z_t | \x_{t|t-1}; \params),
\end{equation}
where \(p(\z_t | \x_{t|t-1}; \params)\) is the likelihood of observing measurement \(\z_t\) given the predicted state \(\x_{t|t-1}\) under the parameter set \(\params\). 
Note that the required innovation covariance is calculated as a combination of the innovation covariances in the individual modes, as described in \cite{brandenburger2023learning}.
This formulation effectively captures the fidelity of the state predictions to the measurements, driving the parameter adjustments to enhance filter accuracy.

To compute the gradients required for parameter updates, automatic differentiation tools are employed, which provide an efficient and accurate way to obtain derivatives in complex models. 
These tools facilitate the computation of partial derivatives with respect to each parameter in the model, thus enabling a systematic and robust optimization process.

The optimization is carried out iteratively, where each iteration involves:
\begin{itemize}
	\item Predicting the state and measurement for each mode using the current filter parameters.
	\item Calculating the loss based on the discrepancies between the predicted and actual measurements.
	\item Updating the parameters using the gradient of the loss.
\end{itemize}

Using this procedure, the \ac{imm} filter parameters are refined in a manner that is directly informed by the data, thereby adapting the filter to better reflect the true dynamics of the system it models. 
This adaptability is crucial for applications where model parameters cannot be accurately predetermined and must be learned from operational data~\cite{brandenburger2023learning}. Due to the small number of parameters, the optimisation typically finishes within \SI{30}{\minute} of training time and does not require a GPU.

\section{Dynamic Models Based on Neural Networks}
\label{sec:mkf}
\subsection{\acf{mkf}}

In tracking applications, the target distribution is often approximated as a Gaussian with mean vector $\bfx$ and covariance matrix $\bfP$. 
A Gaussian distribution preserves its Gaussianity under linear combinations and can be fully described by its first two moments $\bfx$ and $\bfP$. 
This implies that the Bayes recursion for Gaussian target distributions can be closed if the dynamic and measurement models are linear, leading to the well-known Kalman filter recursion \cite{kalman1960new}.
This method is an optimal and computationally tractable solution of the Bayes filter for linear Gaussian systems.

Unfortunately, non-linear target dynamics are very common, hence the Kalman filter assumptions are often too restrictive.
For moderate non-linearities it suffices to linearise the motion model like in the Extended or Unscented \aclp{kf} (\acsu{ekf} and \acsu{ukf}) \cite{julier1997new}, however strong manoeuvres still cannot be compensated by those techniques.
In addition, common transition models are formulated as Markov processes, i.e. they only consider the current target state for the prediction of the next time step, hence disregarding most of the target history.

To alleviate these restrictions, the \ac{mkf} has been introduced in \cite{jung2020mnemonic}. 
Instead of the standard Kalman prediction, it uses a recurrent neural network to predict a mean and covariance from a sequence of input data. 
The recurrent structure of the predictor network makes it possible to overcome the Markov assumption. 
Furthermore, the network architecture enforces a Gaussian output, therefore the Kalman recursion can be closed while any type of target dynamics can be modelled.

In particular, the predicted $2d_x$-dimensional mean $\bfx_{t+1|t}=[\bfx_{t+1|t}^\text{pos},\bfx_{t+1|t}^\text{vel}]$ and covariance $\bfP_{t+1|t}$ are calculated as:
\begin{subequations}
	\begin{align}
		\bfx_{t+1|t}^\text{pos} &= \bfx_{t|t}^\text{pos} + \Delta_t \bfv_{t+1|t}^\text{NN},\\
		\bfx_{t+1|t}^\text{vel} &= \bfv_{t+1|t}^\text{NN},\\
		\bfP_{t+1|t} &= \bfP_{t|t} + \bfV^T \bfC_{t+1|t}^\text{NN} (\bfC_{t+1|t}^\text{NN})^T \bfV + \bfQ_{t+1},
	\end{align}
\end{subequations}
where $\bfv_{t+1|t}^\text{NN}$ and $\bfC_{t+1|t}^\text{NN}$ are given by the neural network predictor, standing for the predicted velocity as well as the Cholesky decomposition of its covariance. 
The matrix $\bfV$ denotes the state projection matrix onto the velocity dimensions.
Alternatively, it is possible to formulate a neural network that directly predicts the target position as it was proposed in \cite{jung2020mnemonic}.
However, learning the velocity information instead of absolute positions provides more robustness since the output becomes translation- and rotation-invariant \cite{jung2019sequential}.

Irrespective of the chosen \ac{mkf} prediction, the update step follows the standard (or alternatively, the Extended or Unscented) Kalman equations. 

\subsection{Long Short-Term Memory Architecture}
The original form of the \ac{mkf} uses a so-called \acf{lstm} neural network, as introduced in \cite{hochreiter1997long}. 
It has a recurrent structure, i.e.\ its internal state is recursively fed back as an additional input in every step.
Like this, temporal information is conserved in the internal state. 
To overcome the vanishing gradient problem of standard recurrent networks, \ac{lstm} nodes have several information gates that pass relevant information to the internal memory or specifically forget irrelevant information.

For the architecture of the \ac{mkf}, an input layer of dimension $d_x$ is implemented that passes the input states $\bfx_k$ to an \ac{lstm} layer with $n_\text{HU}^L$ hidden units. 
Its output is then passed to an additional dense layer with $n_\text{HU}^D$ neurons, which is in turn connected to an output layer of dimension $0.5 d_x (d_x+3)$, as proposed in \cite{williams1996using}. 
The output layer hence returns the $d_x$ elements of the predicted mean and $0.5 d_x (d_x+1)$ values filling the Cholesky decomposition of the predicted covariance. 
The Cholesky decomposition always yields a positive definite matrix while involving less parameters than a full covariance, and hence is easier to learn. 
As described in \cite{williams1996using}, the loss can be formulated as the negative log-likelihood between the network estimate and the label. 
In the used implementation, we define the loss based on the measurement $\bfz_t$:
\begin{align}
	\lambda&(\bfx_{t+1|t},\bfC_{t+1|t},\bfz_t) \nonumber \\
	&=\left\lVert\left(\frac{1}{2} \bfC_{t+1|t} (\bfH_t \bfx_{t+1|t}-\bfz_t)-\text{diag}(\bfC_{t+1|t})\right)\right\rVert_1,
\end{align}
where $\bfH_t$ is the measurement matrix as used in the Kalman filter. 
Note that measurement noise will be averaged out by the model during learning if sufficient input data is used to train the model.

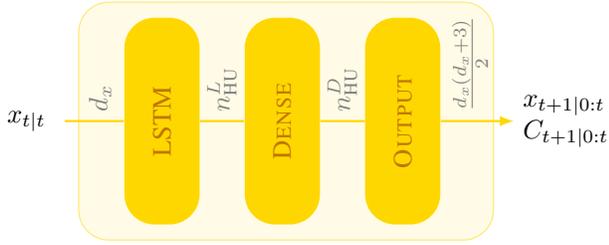
\begin{figure}[!t]
	\centering
	\input{figures/mdlstm.tikz}
	\caption{The \ac{lstm} network architecture used for the \acs{mkf} implementation. 
		The arrows display the number of outputs of the respective layer: 
		$d_x$ is the dimension of the target state space, $n_\text{HU}^L$ and $n_\text{HU}^D$ denote the number of hidden units of the \ac{lstm} and dense layers, respectively, and $d_N=0.5 d_x (d_x+3)$.}
	\label{fig:lstm}
\end{figure}

A graphical representation of the used network architecture is found in Fig.\ \ref{fig:lstm}. 
Note that other neural network architectures could be used in the \ac{mkf} recursion, such as transformer networks \cite{vaswani2017attention} or \acp{gru} \cite{cho2014learning}. 
The \ac{lstm} can be trained on a conventional CPU within \SI{24}{\hour}. 

\section{Comparative Study of the Presented Methods}
\label{sec:comparison}

The capabilities of the three presented methods are evaluated in the following three experiments.
The first experiment is performed on simulated data to have full control over all parameters.
The setup is described in \ref{subsubsec:sim_setup} and the evaluation given in \ref{subsubsec:sim_results}.
Additionally, the methods are tested on the motion of real objects, i.e. a \ac{uav} and a \ac{rib}, to demonstrate the methods' capabilities on real-world scenarios.
Since the focus is on the prediction of target dynamics rather than the sensor model, synthetic measurements are generated from the datasets using a fictional range-bearing sensor to make sure that possible sensing artefacts are excluded from the learned model.
The \ac{uav} dataset is described in \ref{subsubsec:drone_setup} and evaluated in \ref{subsubsec:drone_results}, whereas the \ac{rib} data is introduced in \ref{subsubsec:rib_setup} and analysed in \ref{subsubsec:rib_results}.

For all experiments, the optimised \ac{imm} as well as the \ac{lstm}-based \ac{mkf} were trained on multiple Intel\textsuperscript{\textregistered} Xeon\textsuperscript{\textregistered} Gold 6126 CPU @ 2.60GHz cores with a learning rate of $5 \cdot 10^{-4}$.
The \ac{lstm} network of the \ac{mkf} was equipped with $n_\text{HU}^L=n_\text{HU}^D=32$ hidden units for the \ac{lstm} and dense layers, respectively. 
Training this model required $100000$ iterations, whereas the \ac{imm} parameters were optimised using only $10000$ epochs.
To account for the range-bearing measurements adequately, the \ac{imm} and \ac{mkf} correction step was implemented as an \ac{ekf} update with the measurement model $h$ as defined in \eqref{eq:measurement_model}.
The \ac{gp}-based particle filter, on the other hand, was trained on a MacBook Pro with Apple M2 Pro processor and 16GB RAM.
\begin{table}[h!]
	\centering
	\caption{Training times of the different methods. \label{tab:traintimes}}
	\begin{tabular}{|c|ccc|}
		\hline
		Dataset & \acs{gp}~\cite{sun2022gaussian} & Opt. \acs{imm}~\cite{brandenburger2023learning} & \acs{mkf}~\cite{jung2020mnemonic}\\
		\hline
		Sim. & \SI{111}{\second} & \SI{1998}{\second} (10k steps) & \SI{23964}{\second} (100k steps)\\
		UAV & \SI{495}{\second} & \SI{1488}{\second} (10k steps) & \SI{15912}{\second} (100k steps)\\
		RIB & \SI{1031}{\second} & \SI{2037}{\second} (10k steps) & \SI{22666}{\second} (100k steps)\\
		\hline
	\end{tabular}
\end{table}
Resulting training times are shown in Tab.~\ref{tab:traintimes}.
Note that for the \ac{gct} data, the \ac{gp} is trained on a reduced set size of $50$ trajectories, whereas the \ac{mkf} and optimised \ac{imm} use all of the $512$ trajectories.

In all experiments described below, a standard \ac{ekf} with \ac{cwna} motion \cite{koch2016tracking} is chosen as a benchmark.

\subsection{Experimental Setup}
\subsubsection{Simulated Data}
\label{subsubsec:sim_setup}
\begin{figure}[!t]
	\centering
	\input{figures/trajectory_sample_gt59_test.tikz}
	\caption{ Sample trajectory (\textbf{---}) and synthetic measurements (\textcolor{green}{$\times$}), $100$ time steps.
	The simulated sensor is located at the origin.}
	\label{fig:example_trajectory} 
\end{figure}
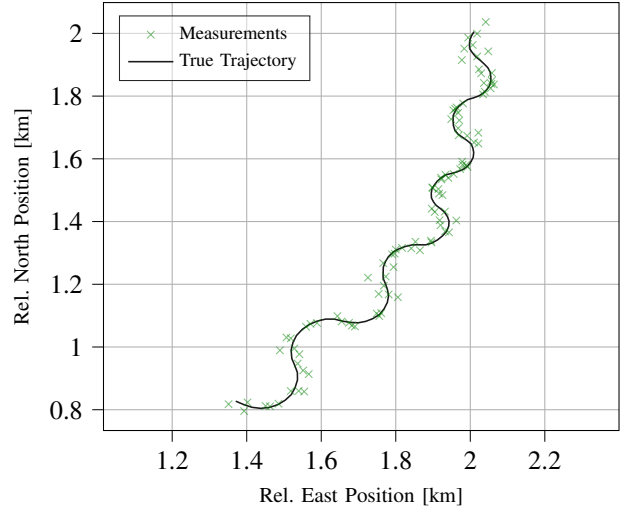
For the first experiment in this paper, trajectories were simulated using a \ac{gct} model similar to the data in \cite{sun2022gaussian}. 
In particular, the \ac{gct} model is defined as a regular succession of left and right coordinated turns with sagittal acceleration $\alpha_{t,k}^t=0$ and lateral accelerations $\alpha_{t,k}^n=\pm \upsilon\si{\degree\per\second}$ with $\upsilon \sim U(10,15)$, where $U(a,b)$ denotes a uniform distribution with bounds $a, b$. 
The starting position is chosen randomly between \SI{2000}{\metre} and \SI{2100}{\metre} in the two Cartesian dimensions, while the speed is randomly initialised with $|v_0| = \pm \SI{10}{\metre\per\second}$.
From the generated trajectories, measurements are extracted in polar coordinates by applying the measurement model:
\begin{equation}
	h(\bfx_k) = \begin{bmatrix}
		r_k \\ \alpha_k
	\end{bmatrix}
	= \begin{bmatrix}
		\sqrt{(x_k^1)^2 + (x_k^2)^2} \\ \text{atan2} (x_k^2, x_k^1)
	\end{bmatrix},
	\label{eq:measurement_model}
\end{equation}
subject to additional white noise with variance $\sigma_r=\SI{25}{\metre}$ and $\sigma_\alpha=\SI{0.01}{\radian}$. 
An example of a test trajectory with ground truth (\textbf{---}) and measurements (\textcolor{green}{$\times$}) is shown in Fig.\ \ref{fig:example_trajectory}. 

The \ac{gp} model already leads to good results on longer test sequences using a short trajectory of $20$ time steps for the learning phase \cite{sun2022gaussian}, which corresponds to one full oscillation of the \ac{gct} model. 
The \ac{lstm} used in the \ac{mkf}, on the other hand, benefits from longer temporal contexts, therefore it is trained on $512$ longer trajectories over $100$ time steps as shown in Fig.\ \ref{fig:example_trajectory}. 
For convenience, the \ac{imm} is also trained on the same trajectories of length $100$. 
The same set of $50$ test runs with trajectories over $100$ time steps is used to compare the three presented filters.

\subsubsection{\acs{uav} Dataset}
\label{subsubsec:drone_setup}
\begin{figure}[!t]
	\centering
	\input{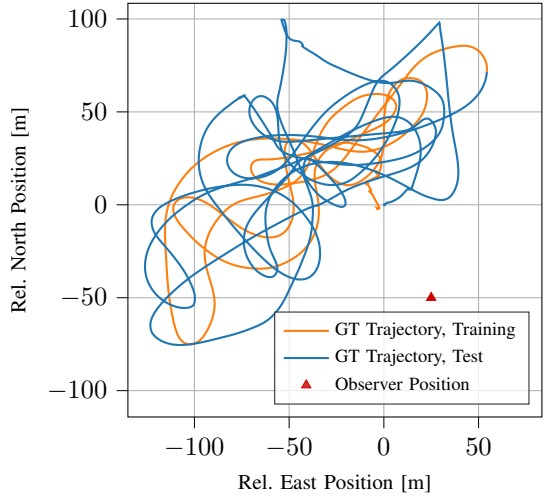}
	\caption{\ac{uav} dataset, relative east/north positions to the start position over the whole flight duration.
	The hypothetical observer (\textcolor{red}{$\blacktriangle$}) is placed at $[\SI{25}{\metre},\SI{-50}{\metre}]$ relative to the start position $[0,0]$.}
	\label{fig:drone_gps}
\end{figure}
The \ac{uav} dataset was created on June 11, 2024 at Fraunhofer FKIE by collecting the \ac{gps} positions at \SI{50}{\hertz} of a DJI M600 hexacopter for a total flight duration of \SI{439}{\second}. 
The dataset was collected under moderate wind conditions of around \SI{13}{\kilo\metre\per\hour}, using an approximate altitude of \SI{30}{\metre} throughout the experiment.
The \ac{uav} was steered manually to include different behavioural patterns into the trajectory. 
The full path is shown in Fig. \ref{fig:drone_gps}.
From this data, measurements were generated synthetically, assuming a range-bearing sensor at position $[\SI{25}{\metre},\SI{-50}{\metre}]$ relative to the trajectory's start point.
Again, the measurement model \eqref{eq:measurement_model} is used, however with $\sigma_r=\SI{0.5}{\metre}$ and $\sigma_\alpha=\SI{5e-5}{\radian}$, respectively.
With a reduced sampling rate of \SI{10}{\hertz}, the training was performed on the last \SI{160}{\second} of the full trajectory since it includes multiple sharp turns that improve the data variety.
The methods were then tested on the remaining first part of the trajectory.
Note that the training and test set were divided into $15$ and $28$ tracklets (minibatches), respectively, having a length $100$ time steps which corresponds to a window-size of approximately \SI{10}{\second}.

\subsubsection{\acs{rib} Dataset}
\label{subsubsec:rib_setup}
\begin{figure}[!t]
	\centering
		\input{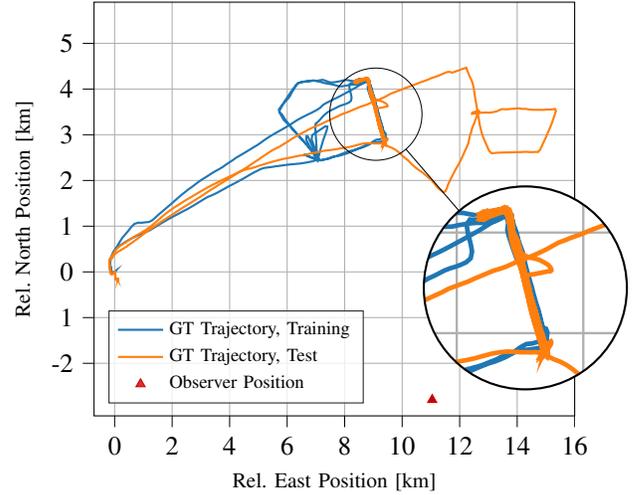}
	\caption{\acs{rib} datasets recorded on two different days of the trial.
	The assumed sensor (\textcolor{red}{$\blacktriangle$}) is placed at the Leonardo headquarters, which is about \SI{11}{\kilo\metre} east and \SI{2.8}{\kilo\metre} south of Port Edgar where the \ac{rib} started on both days.}
	\label{fig:rib_gps}
\end{figure}
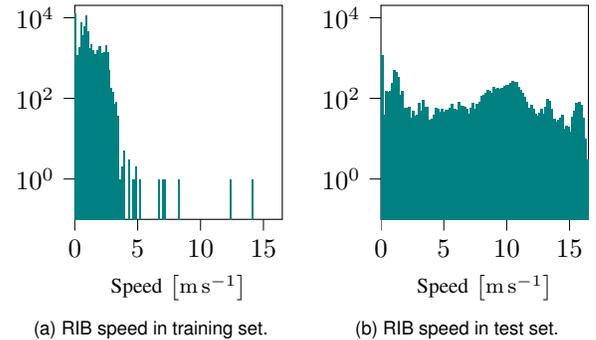
\begin{figure}[!ht]
	\centering
	\subfloat[\acs{rib} speed in training set.]{
		\input{figures/velocities_pred_hist_100bins_log.tikz}
		\label{fig:rib_vel_pre}}
	\subfloat[\acs{rib} speed in test set.]{
		\input{figures/velocities_upd_hist_100bins_log.tikz}
		\label{fig:rib_vel_upd}}
	\caption{Speed distribution for the \acs{rib} data in form of a histogram.}
	\label{fig:rib_velocities}
\end{figure}

The last experiment consists of \ac{gps} measurements from a collaborative \ac{rib} navigating on the Firth of Forth near Edinburgh, UK.
The data was recorded at \SI{5}{\hertz} by Leonardo UK Ltd on November 11 and 13, 2020 as a part of the activities of the NATO research task group SET 278, resulting in a training dataset with a duration of \SI{3.8}{\hour} and a test set of \SI{3.02}{\hour}.
Different behaviours were planned with speeds ranging from \SI{5}{\knot} to \SI{30}{\knot}, however the \ac{rib} effectively reached a maximum speed of \SI{20}{\knot} due to harsh weather conditions during the trial.
On both days of the trial, the \ac{rib} performed several loops of a hockey stick-shaped manoeuvre (see magnified area in Fig.~\ref{fig:rib_gps}) during the cruise to include a repeated pattern in the data, however with different speeds.
The two recorded trajectories are plotted in Fig.~\ref{fig:rib_gps} and the corresponding speeds are shown in Fig.\ \ref{fig:rib_velocities} in histogram form.

Similar to the \ac{uav} scenario \ref{subsubsec:drone_setup}, simulated measurements were generated from the \ac{gps} data.
For this purpose, the sensor position was assumed at the actual Leonardo headquarters, which results in an observer position at around \SI{11}{\kilo\metre} east and \SI{2.8}{\kilo\metre} south of Port Edgar where the \ac{rib} started on both days of the trial.
Again, tracklets were formed from the training and test trajectories with $100$ time steps each, corresponding to a duration of \SI{20}{\second} per tracklet at \SI{5}{\hertz}.

\subsection{Results}
This section summarises the results achieved with the different filtering methods on the three datasets described above.
In the following, the estimated tracks as well as the \ac{rmse} graphs of the \ac{ekf} reference are always shown in red, while the \ac{gp} results are plotted in purple, the optimised \ac{imm} in blue and the \ac{mkf} in green. 
The average measurement error compared to the ground truth is plotted for reference in the form of black crosses.
Furthermore, the overall average \ac{rmse} for each method on each dataset is summarised in Tab.~\ref{tab:pre} and \ref{tab:upd}.
Here, the total errors are denoted with the label \textit{Avg.}, and the relative change with respect to the measurement noise level is labelled with \textit{Rel}.
A relative score below $1.0$ states that the respective method performs better than the noise level, whereas values above $1.0$ suggest a decreased accuracy in relation to the measurement noise.
The respective best method on each dataset is marked in bold.

\begin{table}[h!]
	\centering
	\caption{\acs{rmse} after the prediction.\label{tab:pre}}
	\begin{tabular}{|cc|ccc|c|}
		\hline
		& Metric & \acs{gp}~\cite{sun2022gaussian} & Opt. \acs{imm}~\cite{brandenburger2023learning} & \acs{mkf}~\cite{jung2020mnemonic} & \acs{ekf}\\
		\hline
		\parbox[t]{2mm}{\multirow{2}{*}{\rotatebox[origin=c]{90}{Sim.}}} & Avg. & 19.4747 & 14.9194 & \textbf{13.8068} & 15.5888\\ 
		& Rel. & 1.3491 & 1.0335 & \textbf{0.9565} & 1.0799\\\hline
		\parbox[t]{2mm}{\multirow{2}{*}{\rotatebox[origin=c]{90}{UAV}}} & Avg. & 1.2400 & \textbf{1.0628} & 1.9961 & 2.6594\\
		& Rel. & 0.8492 & \textbf{0.7279} & 1.3671 & 1.8213\\\hline
		\parbox[t]{2mm}{\multirow{2}{*}{\rotatebox[origin=c]{90}{RIB}}} & Avg. & 54.2740 & \textbf{17.4599} & 35.6627 & 19.7488\\
		& Rel. & 1.8148 & \textbf{0.5838} & 1.1923 & 0.6603\\
		\hline
	\end{tabular}
\end{table}
\begin{table}[h!]
	\centering
	\caption{\acs{rmse} after the update. \label{tab:upd}}
	\begin{tabular}{|cc|ccc|c|}
		\hline
		& Metric & \acs{gp}~\cite{sun2022gaussian} & Opt. \acs{imm}~\cite{brandenburger2023learning} & \acs{mkf}~\cite{jung2020mnemonic} & \acs{ekf}\\
		\hline
		\parbox[t]{2mm}{\multirow{2}{*}{\rotatebox[origin=c]{90}{Sim.}}} & Avg. & 14.4378 & \textbf{12.9156} & 13.0762 & 13.4857\\
		& Rel. & 1.0002 & \textbf{0.8947} & 0.9059 & 0.9342\\\hline
		\parbox[t]{2mm}{\multirow{2}{*}{\rotatebox[origin=c]{90}{UAV}}} & Avg. & 1.1824 & \textbf{0.9614} & 1.2899 & 1.4735\\
		& Rel. & 0.8100 & \textbf{0.6584} & 0.8834 & 1.0092\\\hline
		\parbox[t]{2mm}{\multirow{2}{*}{\rotatebox[origin=c]{90}{RIB}}} & Avg. & 51.1044 & \textbf{16.3451} & 33.8194 & 17.3403\\
		& Rel. & 1.7088 & \textbf{0.5465} & 1.1308 & 0.5798\\
		\hline
	\end{tabular}
\end{table}

\subsubsection{Simulated Data}
\label{subsubsec:sim_results}
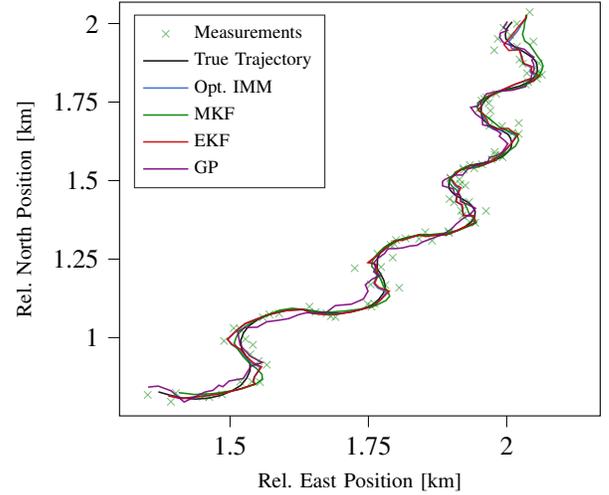
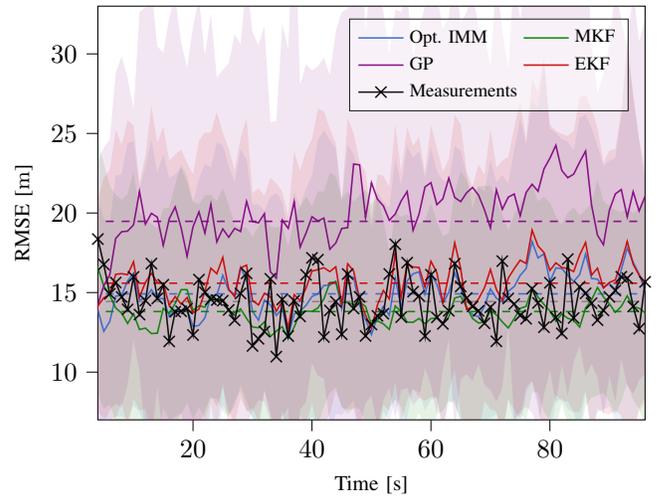
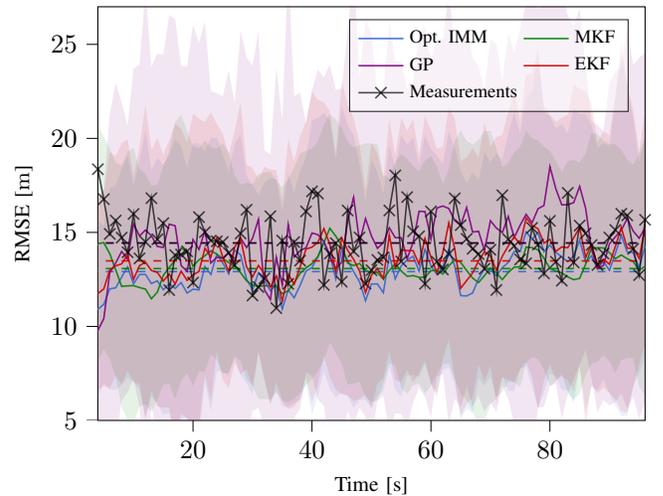
\begin{figure}[h!t]
\centering
\subfloat[Estimated trajectory of a test sample after the filter update.]{
	\input{figures/trajectory_sample59_test.tikz}
	\label{fig:traj_post_gct}}
\hfil
\subfloat[Prediction.]{
	\input{figures/gct_pred_avg_test.tikz}
	\label{fig:avg_pred_gct}}
\hfil
\subfloat[Update.]{
	\input{figures/gct_post_avg_test.tikz}
	\label{fig:avg_upd_gct}}
\caption{Average \ac{rmse} results for the predicted (a) and posterior (b) estimation over $512$ simulated test trajectories with \acs{gct} motion.
	Average errors over time are displayed as dashed lines. 
	Shaded regions correspond to the $2\sigma$ confidence.}
\label{fig:avg_results_gct}
\end{figure}

The first experiment is performed on the simulated data described in Sec.~\ref{subsubsec:sim_setup}.
In this experiment, the underlying \ac{gct} model is exactly the same for the training and the test set, therefore no model mismatch between the training and the evaluation is expected.
Fig.~\ref{fig:avg_results_gct} shows the performance of the different filters on a sample trajectory (Fig.\ \ref{fig:traj_post_gct}) as well as the averaged \ac{rmse} over time after the prediction (Fig.\ \ref{fig:avg_pred_gct}) and update (Fig.\ \ref{fig:avg_upd_gct}). 

After the prediction, most filters achieve average errors above the measurement noise level.
The \ac{gp}-based filter \cite{sun2022gaussian} seems to struggle the most with this dataset, which might be caused by the chosen kernel width.
The optimised \ac{imm} \cite{brandenburger2023learning} and the reference model perform \SI{3.35}{\percent} and \SI{7.99}{\percent} above the measurement noise level, respectively. 
The \ac{mkf} \cite{jung2020mnemonic}, on the other hand, has learned the underlying model well enough from the training set and thus reaches a performance improvement of \SI{4.35}{\percent} with respect to the noise level.
The update naturally improves all results because of the information gained from the measurement.
Here, the optimised \ac{imm} and the \ac{mkf} yield similar improvements of \SI{10.53}{\percent} and \SI{9.41}{\percent}, respectively, followed by the \ac{ekf} which reaches an improvement of \SI{6.58}{\percent} compared to the observation noise.

\subsubsection{\acs{uav} Dataset}
\label{subsubsec:drone_results}

\begin{figure}[t!]
	\centering
	\subfloat[Estimated trajectory of a test sample after the filter update.]{
		\input{figures/drone_trajectory_sample5_test.tikz}
		\label{fig:traj_post_drone}}
	\hfil
	\subfloat[Prediction.]{
		\input{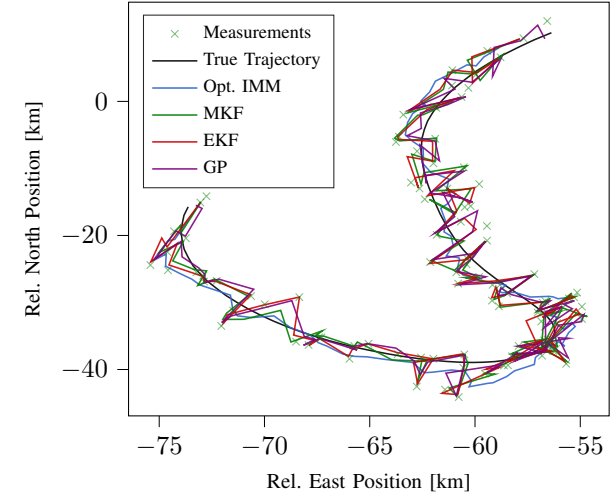}
		\label{fig:avg_pred_drone}}
	\hfill
	\subfloat[Update.]{
		\input{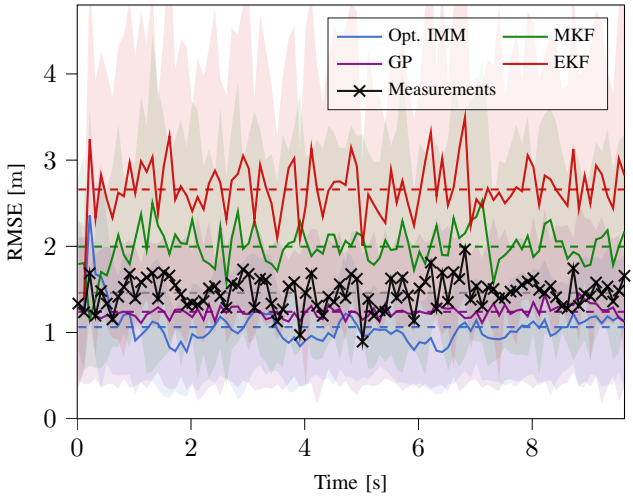}
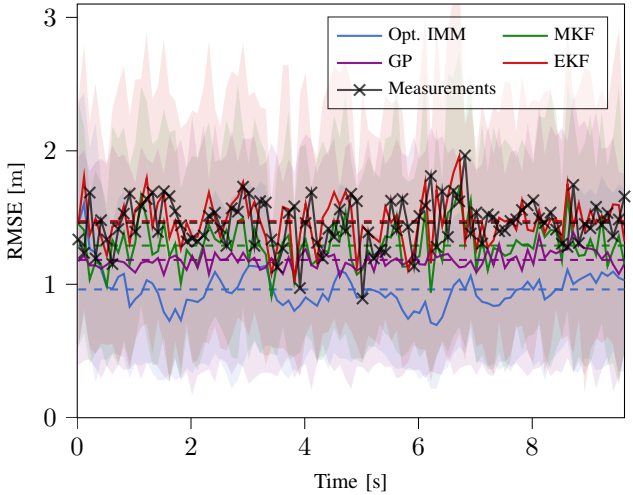
		\label{fig:avg_upd_drone}}
	\caption{Average \ac{rmse} results for the predicted (a) and posterior (b) estimation over $28$ test trajectories of the \ac{uav} dataset.
		Average errors over time are displayed as dashed lines. 
		Shaded regions correspond to the $2\sigma$ confidence.}
	\label{fig:avg_results_drone}
\end{figure}

As described in section \ref{subsubsec:drone_setup}, the second dataset consists of real-world data captured from a manually controlled multicopter. 
Fig.~\ref{fig:avg_results_drone} shows the performance of each method in terms of a sample trajectory (Fig.\ \ref{fig:traj_post_drone}) and the average \ac{rmse} results (Fig.\ \ref{fig:avg_pred_drone} and \ref{fig:avg_upd_drone}). 
With respect to the \ac{rmse}, the \ac{ekf} performs worst, which is especially visible for the target prediction (Fig.~\ref{fig:avg_pred_drone}). 
In addition, the quality of the posterior estimate of the \ac{ekf} closely matches the quality of the measurements themselves (Fig.~\ref{fig:avg_upd_drone}). 
This indicates a substantial model mismatch between the target dynamics utilized by the \ac{ekf} and the actual dynamics.
A similar, but less grave mismatch can be seen for the \ac{mkf}~\cite{jung2020mnemonic}, which is able to show a significant improvement of the posterior \ac{rmse} compared to the measurements. 
A possible explanation for this imbalance is the small size of the dataset. 
Since the training of neural networks can require a significant amount of data to prevent overfitting, the performance of such methods can degrade when training data is sparse. 
This stands in contrast to the simulated dataset shown in section~\ref{subsubsec:sim_results}, where there is not only a significant amount of training data available, but it is also drawn from the same distribution as the test data. 
Since the train and test data of the \ac{uav} dataset is a split at an arbitrary point in time, there can be a mismatch between the training and test data, which can further reduce the performance of learning approaches.
The \ac{gp}~\cite{sun2022gaussian} is able to further improve upon the results of the \ac{mkf} for both the prediction and posterior \ac{rmse}. 
This improvement is most significant for the prediction, which indicates that the \ac{gp} is able to sufficiently reflect the target dynamics. 
In contrast to the \ac{mkf}, the \ac{gp} analyses the overall statistics of the target motion, therefore the comparably small dataset suffices for regression of the model.
Of the compared methods, the optimized \ac{imm} filter~\cite{brandenburger2023learning} yields the best average \ac{rmse} for the \ac{uav} dataset. 
Since it directly models the physics of the target and is parametrized only by a small set of variables, the small dataset is sufficient to optimize the filter. 
Furthermore, as seen in Fig.~\ref{fig:drone_gps}, the target trajectory features straight and curved sections, which is well suited for the \ac{imm} filter.

\subsubsection{\acs{rib} Dataset}
\label{subsubsec:rib_results}
\begin{figure}[t!]
	\centering
	\subfloat[Estimated trajectory of a test sample after the filter update.]{
		\input{figures/rib_trajectory_sample72_test.tikz}
		\label{fig:traj_post_rib}}
	\hfil
	\subfloat[Prediction.]{
		\input{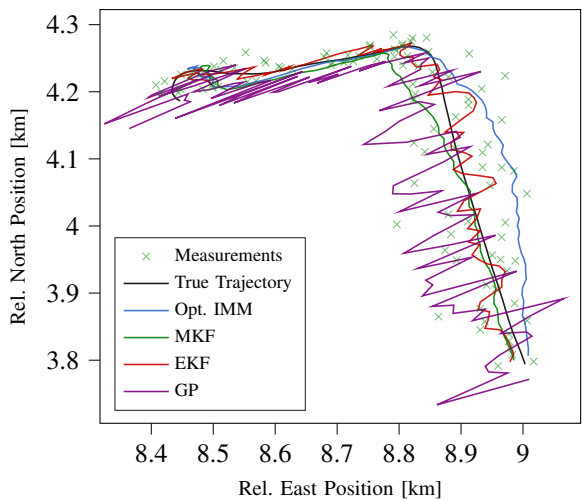}
		\label{fig:avg_pred_rib}}
	\hfil
	\subfloat[Update.]{
		\input{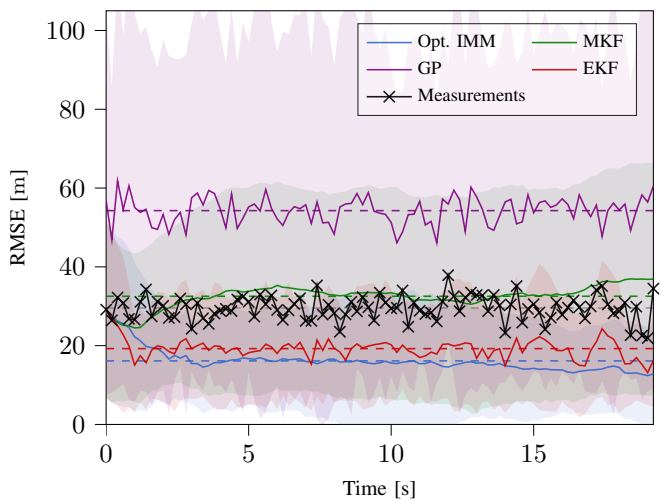}
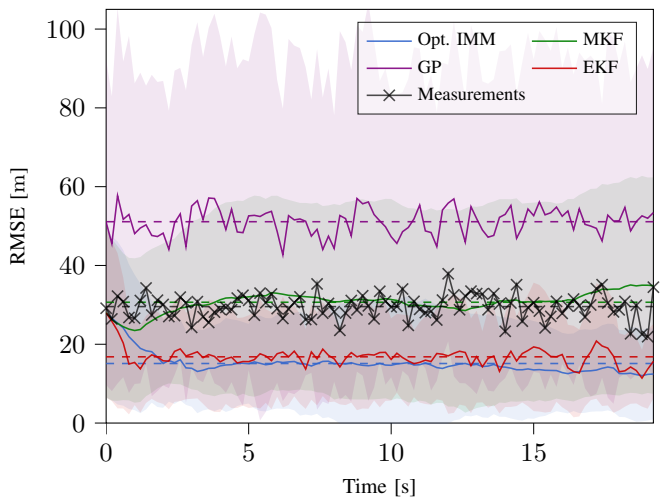
		\label{fig:avg_upd_rib}}
	\caption{Average \ac{rmse} results for the predicted (a) and posterior (b) estimation over $109$ test trajectories of the \acs{rib} dataset.
		Average errors over time are displayed as dashed lines. 
		Shaded regions correspond to the $2\sigma$ confidence.}
	\label{fig:avg_results_rib}
\end{figure}

The last experiment is performed on the trajectory of a cooperative \ac{rib} as explained in Sec.~\ref{subsubsec:rib_setup}. 
A sample of the estimated trajectory and the obtained \ac{rmse} results are shown in Fig.\ \ref{fig:avg_results_rib}.
For the neural network, this dataset is particularly challenging: 
Although the training and test set contain similar trajectories (see Fig.~\ref{fig:rib_gps}), the underlying velocities are drastically dissimilar, as Fig.~\ref{fig:rib_velocities} illustrates.
In particular, the training dataset barely contains any velocities over \SI{4}{\metre\per\second}, whereas the majority of the test data ranges between \SI{5}{\metre\per\second} and \SI{12}{\metre\per\second}.
As expected, the \ac{lstm}-based \ac{mkf}~\cite{jung2020mnemonic} suffers from the model mismatch between the training and test sets, resulting in a prediction error that is \SI{19.23}{\percent} higher after training and \SI{13.08}{\percent} higher after testing when compared to the measurement noise. 
The \ac{gp}-based filter \cite{sun2022gaussian} results in even higher errors of \SI{81.48}{\percent} above measurement noise level after training and \SI{70.88}{\percent} after testing.
A more heterogeneous test set with higher similarity to the training set is expected to bring better results in both cases.
The \ac{ekf}, on the other hand, already reaches an improvement of \SI{33.97}{\percent} and \SI{42.02}{\percent} in comparison to the noise level after training and testing, respectively, while the optimised \ac{imm} \cite{brandenburger2023learning} is able to outperform the baseline even further.
Like in the \ac{uav} dataset, the \ac{imm} benefits from its knowledge of a \ac{cwna} motion, so that the parameter mismatch between the training and the test set can still be compensated even though the higher velocities have not been seen during training.

\acresetall

\section{Conclusion}
\label{sec:conclusion}
This paper presented and compared three different Bayesian single-target trackers that model object motion with a data-driven approach.
The first method builds on a \ac{gp} that learns statistical patterns in the provided data, together with the underlying uncertainty. 
This model is incorporated in a particle filter approach to deal with non-linear range-bearing measurements. 
The second algorithm is based on an \ac{imm} filter, optimising parameters such as the mode transition probabilities as well as the sensor and process noise.
Finally, the third method uses a recurrent neural network architecture to estimate a state and covariance from a stream of input data, hence making it possible to close the (extended) Kalman filter recursion. 
It could be shown that the learned motion models clearly outperform a baseline \ac{ekf} in cases where the target shows strong manoeuvrability, which cannot be modelled with linear transition functions. 
Still, learned models have a similar issue in cases where the test data differs greatly from the training data with respect to the underlying motion.
Here, prior knowledge about a physical motion model can be beneficial to overcome the issues of overly homogeneous or insufficiently large training sets.

This work shows the suitability of different \ac{ml} architectures to model target dynamics. 
As prior publications have shown, a notable advantage over analytic methods is their robustness against low detection rates and target occlusion, which are common challenges in radar signal processing. 
Moreover, no prior knowledge about the physics of the target dynamics is required, allowing for the encapsulation of various target behaviours within a single model. 
The portrayed methods thus serve as a guideline for the development of future tracking and reconnaissance systems.
Overall, the presented study shows that both analytic and data-driven models are relevant depending on the application.
Future studies shall investigate possibilities for hybrid approaches, which combine the strengths of both paradigms to yield even more robust results on complex scenarios.

\section*{Acknowledgments}
The authors would like to thank Leonardo UK Ltd, Edinburgh and the NATO SET-278 RTG for providing access to the \acs{rib} dataset and for many fruitful discussions.

\bibliographystyle{IEEEtran}
\bibliography{biblio}

\newpage

\section{Author Biographies}

\begin{IEEEbiography}[{\includegraphics[width=1in,height=1.25in,clip,keepaspectratio]{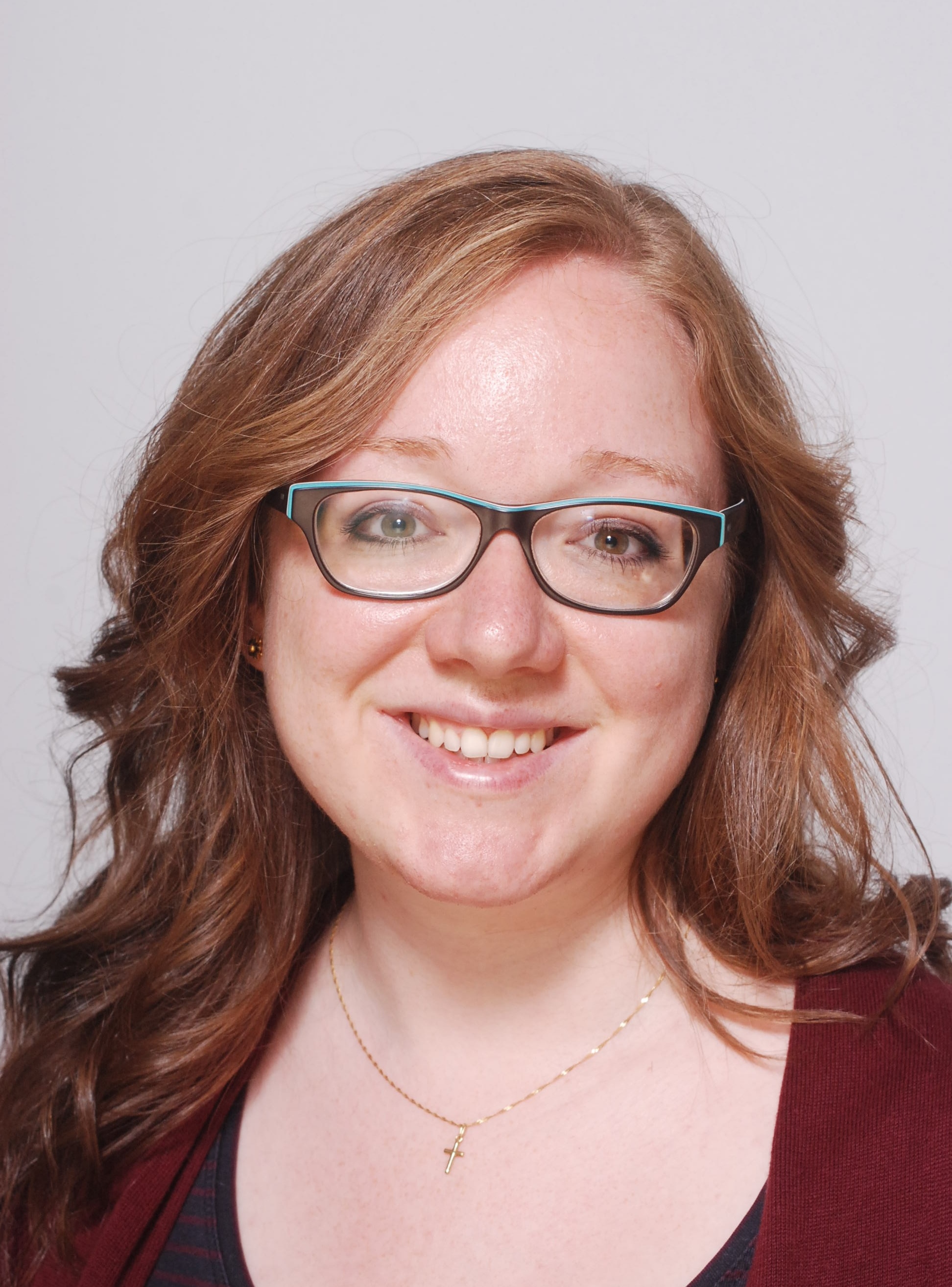}}]{Isabel Schlangen}
holds a German Diploma in mathematics from the University of Bonn (Germany) and a joint M.Sc.\ degree in vision and robotics from the Universities of Burgundy (France), Girona (Spain) and Heriot-Watt (Edinburgh, UK). 
In 2017, she received the Ph.D.\ degree from Heriot-Watt University for her research on multi-object filtering with second-order moment statistics. 
She is a research associate with Fraunhofer FKIE, Wachtberg, Germany, working on multi-object estimation, resource management, and signal intelligence applications.
\end{IEEEbiography}

\begin{IEEEbiography}[{\includegraphics[width=1in,height=1.25in,clip,keepaspectratio]{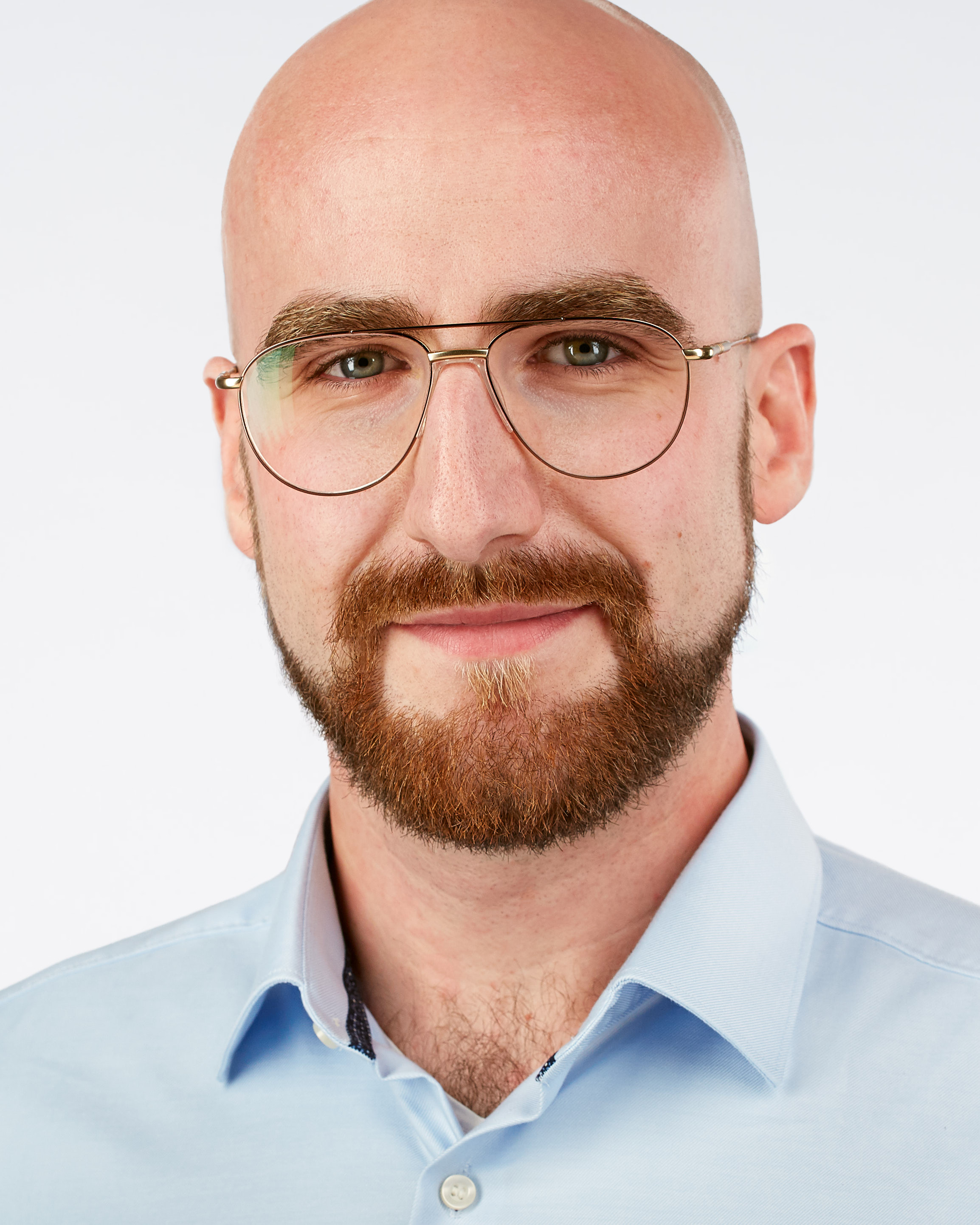}}]{Andr{\'e} Brandenburger}
	received his Master's degree in Computer Science from the University of Bonn, Germany in 2020. 
	While earning his degree, he was part of the Autonomous Intelligent Systems lab at the same university and contributed significantly to winning numerous international robotics competitions. 
	Since graduating, Andr\'e has been a research associate in the Department of Sensor Data and Information Fusion at Fraunhofer FKIE, Germany. 
	In this role, he has engaged in both publicly funded and industry research projects. 
	His research, which primarily revolves around reinforcement learning, path planning, sensor data fusion and numerical optimization, aims to enable automation in challenging applications.
\end{IEEEbiography}

\begin{IEEEbiography}[{\includegraphics[width=1in,height=1.25in,clip,keepaspectratio]{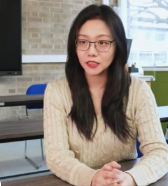}}]{Mengwei Sun}
	is a Lecturer in Sensor Fusion at the School of Aerospace, Transport, and Manufacturing at Cranfield University, UK. She previously served as a research associate at the Institute of Digital Communications at the University of Edinburgh from 2019 to 2023. Her research primarily focuses on advanced algorithms for inference in distributed and modular sensor networks. She is dedicated to developing cutting-edge algorithmic solutions significantly benefit sensor fusion problems in academic research and industry.
\end{IEEEbiography}

\begin{IEEEbiography}[{\includegraphics[width=1in,height=1.25in,clip,keepaspectratio]{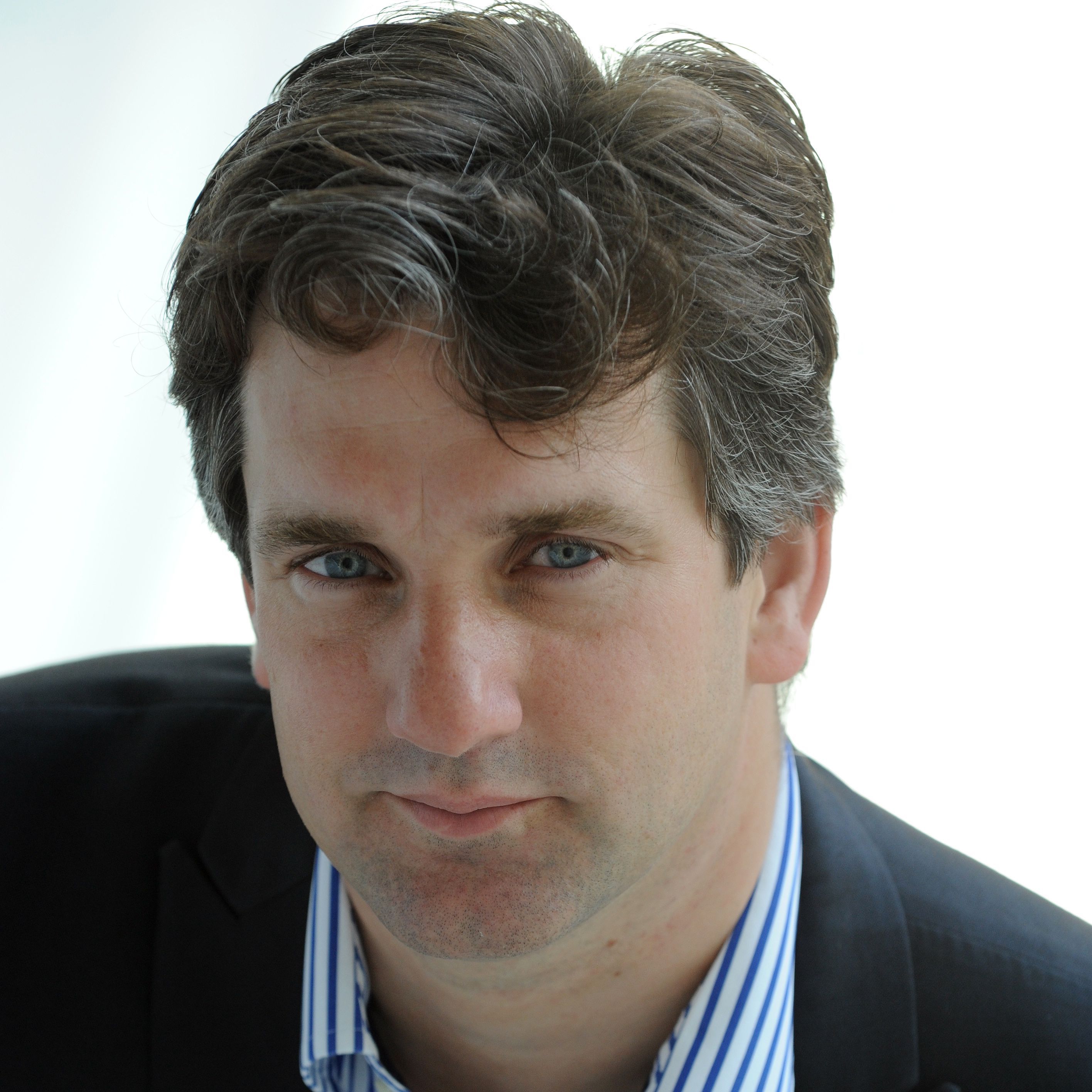}}]{James Hopgood}
	is Director of Electronics \& Electrical Engineering (EEE) in the School of Engineering (SoE) at the University of Edinburgh, UK. 
	His research specialism is statistical signal processing and spans a diverse range of applications, from medical to underwater imaging, to multi-modal sensor-fusion and multi-target tracking. 
	He is supported by $>$\pounds 21M of research funding, is PI on the EPSRC and MoD funded Centre for Doctoral Training in Sensing, Processing, and AI for Defence and Security, and the “DASA: Look Out!” competition for developing future concepts of early warning for the Royal Navy, and is a WP lead on UDRC Phase 3.
\end{IEEEbiography}

\vfill

\end{document}

%% file: figures/gp.tikz
\begin{tikzpicture}[node distance=2.2cm,every node/.style={scale=0.7}] 

\node[black,circle,draw=blue,thick](xipast){$\Delta s_{t-1,k}^*$};
\node[black,rectangle,draw=blue,right of=xipast,thick](xipresent){$~\dot{f}_{t-1,k}^{*s}$};	
\node[black,circle,draw=blue,right of=xipresent,thick](xifuture){$\Delta s_{t,k}^*$};
\node[black,circle,draw=orange,below=0.2cm of xipast,thick](etapast){$\Delta\phi_{t-1,k}^*$};
\node[black,rectangle,draw=orange,right of=etapast,thick](etapresent){$~\Delta\dot{f}_{t-1,k}^{*\phi}$};	
\node[black,circle,draw=orange,right of=etapresent,thick](etafuture){$\Delta\phi_{t,k}^*$};

\draw[blue,-,thick] (xipast.east) -- (xipresent.west) node[gray,pos=0.6,anchor=south,minimum height = 0cm] {};
\draw[blue,-,thick] (xipresent.east) -- (xifuture.west) node[gray,pos=0.5,anchor=south,minimum height = 0cm] {};
\draw[orange,-,thick] (etapast.east) -- (etapresent.west) node[gray,pos=0.6,anchor=south,minimum height = 0cm] {};
\draw[orange,-,thick] (etapresent.east) -- (etafuture.west) node[gray,pos=0.5,anchor=south,minimum height = 0cm] {};
\draw[blue,-,thick] (etapast.east) -- (xipresent.west) node[gray,pos=0.6,anchor=south,minimum height = 0cm] {};
\draw[orange,-,thick] (xipast.east) -- (etapresent.west) node[gray,pos=0.5,anchor=south,minimum height = 0cm] {};

\begin{pgfonlayer}{bg}
	\node[fit={($(xipast.north west)+(-1.2,0.9)$)  ($(etafuture.south east)+(1.2,-0.9)$)},gold!60,fill=gold!10,rounded corners=0.4cm] (rect){};
\end{pgfonlayer}

\node[black,draw=gold,fill=gold!10,rounded corners=0.2cm, left of=rect,text width=1.15cm, anchor=east,node distance=4.5cm](offline){Off-line training};
\node[black,draw=blue,fill=blue!10,rounded corners=0.2cm, below=3.9cm of offline,text width=1.2cm](online){On-line tracking};
\node[left of=rect,black,node distance=3.2cm]{${\huge\dots}$};
\node[right of=rect,black,node distance=3.2cm]{${\huge\dots}$};

\node[black,draw=blue,circle,below=0.4cm of rect,thick](dxi){$D_r$};
\node[black,draw=blue,rectangle,below=0.6cm of dxi,anchor=center,thick](fxi){$\dot{f}_{t-1,k}^s$};
\node[black,draw=blue,circle,below left=0.4cm of fxi,thick](deltaxipast){$\Delta s_{t-2,k}$};
\node[black,draw=blue,circle,below right=0.4cm of fxi,thick](deltaxipresent){$\Delta s_{t-1,k}$};
\node[black,draw=blue,rectangle,below=0.4cm of deltaxipast,thick](plus1){$+$};
\node[black,draw=blue,rectangle,below=0.4cm of deltaxipresent,thick](plus2){$+$};
\node[black,draw=blue,circle,right=0.4cm of plus1,thick](xit){$s_{t-1,k}$};
\node[black,draw=blue,circle,left=0.5cm of plus1,thick](xitminusone){$s_{t-2,k}$};
\node[black,draw=blue,circle,right=0.5cm of plus2,thick](xitplusone){$~s_{t,k}~$};

\draw[blue,-latex,dashed,thick] (rect.south) -- (dxi.north) node[gray,pos=0.6,anchor=south,minimum height = 0cm] {};
\draw[blue,-,thick] (dxi.south) -- (fxi.north) node[gray,pos=0.6,anchor=south,minimum height = 0cm] {};
\draw[blue,-,thick] (deltaxipast.north) -- (fxi.west) node[gray,pos=0.6,anchor=south,minimum height = 0cm] {};
\draw[blue,-,thick] (fxi.east) -- (deltaxipresent.north) node[gray,pos=0.6,anchor=south,minimum height = 0cm] {};
\node[left of=xitminusone,black,node distance=0.8cm]{${\huge\dots}$};
\node[right of=xitplusone,black,node distance=0.8cm]{${\huge\dots}$};

\draw[blue,-,thick] (xitminusone.east) -- (plus1.west) node[gray,pos=0.6,anchor=south,minimum height = 0cm] {};
\draw[blue,-,thick] (plus1.east) -- (xit.west) node[gray,pos=0.6,anchor=south,minimum height = 0cm] {};
\draw[blue,-,thick] (xit.east) -- (plus2.west) node[gray,pos=0.6,anchor=south,minimum height = 0cm] {};
\draw[blue,-,thick] (plus2.east) -- (xitplusone.west) node[gray,pos=0.6,anchor=south,minimum height = 0cm] {};
\draw[blue,-,thick] (deltaxipast.south) -- (plus1.north) node[gray,pos=0.6,anchor=south,minimum height = 0cm] {};
\draw[blue,-,thick] (deltaxipresent.south) -- (plus2.north) node[gray,pos=0.6,anchor=south,minimum height = 0cm] {};
\end{tikzpicture}

%% file: figures/mdlstm.tikz
\begin{tikzpicture}[node distance=1.6cm,minimum width=1.0cm,minimum height=2.75cm,rounded corners=.4cm]

\node(input){$x_{t|t}$};	
\node[black!20!orange,fill=gold,right of=input,node distance=1.8cm](lstm){\rotatebox{90}{\acs{lstm}}};	
\node[black!20!orange,fill=gold,right of=lstm](dense){\rotatebox{90}{\textsc{Dense}}};
\node[black!20!orange,fill=gold,right of=dense](id){\rotatebox{90}{\textsc{Output}}};
\node[right of=id,node distance=2.1cm,text width=1cm](output){${x}_{t+1|0:t}$ ${C}_{t+1|0:t}$};	

\draw[gold,thick,-] (input.east) -- (lstm.west) node[gray,pos=0.6,anchor=south,minimum height = 0cm] {\rotatebox{90}{$d_x$}};
\draw[gold,thick,-] (lstm.east) -- (dense.west) node[gray,pos=0.5,anchor=south,minimum height = 0cm] {\rotatebox{90}{$n_\text{HU}^L$}};
\draw[gold,thick,-] (dense.east) -- (id.west) node[gray,pos=0.5,anchor=south,minimum height = 0cm] {\rotatebox{90}{$n_\text{HU}^D$}};
\draw[gold,thick,-latex] (id.east) -- (output.west) node[gray,pos=0.4,anchor=south,minimum height = 0cm] {\rotatebox{90}{$\frac{d_x(d_x+3)}{2}$}};

\begin{pgfonlayer}{bg}
	\draw[gold!60,fill=gold!10] ($(lstm.north west)+(-0.6,0.2)$) rectangle ($(id.south east)+(0.7,-0.2)$);
\end{pgfonlayer}

\end{tikzpicture}

%% file: figures/trajectory_sample_gt59_test.tikz
\begin{tikzpicture}

\definecolor{darkgray176}{RGB}{176,176,176}
\definecolor{green}{RGB}{0,128,0}
\definecolor{lightgray204}{RGB}{204,204,204}

\begin{axis}[
	legend cell align={left},
	legend style={fill opacity=0.8, draw opacity=1, text opacity=1, draw=lightgray204},
	tick align=outside,
	tick pos=left,
	x grid style={darkgray176},
	xmajorgrids,
	xlabel={\footnotesize Rel. East Position [km]},
	xmin=1015.59849243164, xmax=2399.14479370117,
	xtick style={color=black},
	xtick={1200,1400,1600,1800,2000,2200,2400},
	xticklabels={1.2,1.4,1.6,1.8,2,2.2,2.4},
	y grid style={darkgray176},
	ymajorgrids,
	ylabel={\footnotesize Rel. North Position [km]},
	ymin=733.698699951172, ymax=2098.22842407227,
	ytick style={color=black},
	ytick={600,800,1000,1200,1400,1600,1800,2000,2200},
	yticklabels={0.6,0.8,1,1.2,1.4,1.6,1.8,2,2.2},
	legend style={at={(0.03,0.97)},anchor=north west,legend cell align=left,draw=white!15!black,fill opacity=0.5,draw opacity=1,text opacity=1,font=\scriptsize} 
	]
\addplot [draw=green, fill=green, mark=x, only marks, opacity=0.5]
table{%
x  y
2041.49072265625 2036.20434570312
1994.41174316406 1986.5615234375
2018.38012695312 1999.10900878906
1983.87414550781 1951.42785644531
2006.52575683594 1962.98779296875
1977.35021972656 1914.90307617188
2018.52124023438 1926.49890136719
2048.23852539062 1942.57800292969
2023.24072265625 1885.34814453125
2029.05639648438 1872.44555664062
2057.98510742188 1873.66882324219
2036.57702636719 1841.94006347656
2057.5234375 1848.62060546875
2063.52905273438 1836.95678710938
2038.76086425781 1813.91918945312
2054.5537109375 1824.6513671875
2034.30041503906 1805.58264160156
2056.54174804688 1839.16357421875
1980.87475585938 1776.40185546875
1959.51428222656 1759.6806640625
1963.10168457031 1764.08862304688
1954.34228515625 1754.69396972656
1967.59997558594 1746.43908691406
1948.90612792969 1726.58557128906
1970.64111328125 1722.63659667969
1967.36743164062 1697.697265625
1968.99462890625 1675.03210449219
1991.67736816406 1673.67456054688
2021.67260742188 1683.20288085938
2010.24768066406 1652.23522949219
2021.73413085938 1648.74487304688
1978.26318359375 1591.12353515625
1979.40563964844 1581.37890625
1988.71142578125 1580.59777832031
1972.95678710938 1567.31616210938
1992.65734863281 1574.11889648438
1941.13122558594 1539.29187011719
1955.45043945312 1551.56213378906
1933.34118652344 1549.20495605469
1922.4072265625 1534.10546875
1921.45153808594 1539.68249511719
1898.28723144531 1509.24108886719
1898.57641601562 1506.72338867188
1914.55883789062 1503.50036621094
1916.18762207031 1489.10754394531
1925.22937011719 1484.43823242188
1896.60266113281 1441.19702148438
1904.70446777344 1430.59655761719
1931.96997070312 1431.9091796875
1917.65905761719 1403.9833984375
1920.17529296875 1388.30981445312
1962.48962402344 1403.26953125
1932.56335449219 1368.60791015625
1942.94079589844 1365.61511230469
1895.74340820312 1333.48327636719
1865.02172851562 1308.45056152344
1894.30139160156 1339.04907226562
1841.65124511719 1314.4267578125
1852.78820800781 1335.57690429688
1797.05102539062 1297.71594238281
1799.92456054688 1310.7021484375
1816.91259765625 1316.12353515625
1790.18469238281 1295.41540527344
1766.34326171875 1267.36828613281
1725.35400390625 1221.03540039062
1793.96667480469 1254.74877929688
1773.00451660156 1224.70959472656
1767.76330566406 1195.31481933594
1754.20166015625 1169.11083984375
1780.83117675781 1167.52661132812
1805.72729492188 1158.62927246094
1748.68237304688 1106.70263671875
1760.82385253906 1107.98559570312
1755.76025390625 1098.4970703125
1690.45385742188 1065.44226074219
1682.23901367188 1068.32043457031
1674.42565917969 1077.18286132812
1654.61755371094 1081.38549804688
1643.38757324219 1098.46691894531
1588.66174316406 1076.7490234375
1559.29016113281 1064.39965820312
1571.23828125 1073.79541015625
1506.65466308594 1030.18908691406
1518.16882324219 1027.591796875
1489.11157226562 989.350524902344
1527.85412597656 995.283813476562
1541.28491210938 977.044677734375
1536.74426269531 947.114990234375
1551.49731445312 925.126159667969
1566.43469238281 913.577575683594
1518.28771972656 859.593322753906
1539.40881347656 860.048645019531
1554.28076171875 857.866638183594
1485.41333007812 819.2783203125
1462.52880859375 810.693969726562
1450.48608398438 812.016235351562
1393.06860351562 795.722778320312
1402.09692382812 823.46240234375
1351.21423339844 817.619018554688
};
\addlegendentry{Measurements}
\addplot [semithick, black, opacity=0.9]
table {%
2009.90441894531 2005.38220214844
2003.88806152344 1993.24584960938
1999.18041992188 1980.61291503906
1997.56640625 1965.32250976562
1998.34436035156 1951.33483886719
2005.33374023438 1938.61450195312
2014.14562988281 1925.84875488281
2025.28674316406 1915.41101074219
2036.58837890625 1903.34094238281
2047.17272949219 1890.49890136719
2053.33129882812 1876.42651367188
2055.09326171875 1861.34130859375
2055.25390625 1847.57653808594
2051.64697265625 1833.38012695312
2044.23083496094 1820.40966796875
2034.85412597656 1809.9208984375
2023.54956054688 1800.92700195312
2008.99487304688 1794.43420410156
1993.48132324219 1789.16796875
1980.662109375 1779.43090820312
1969.09191894531 1767.56091308594
1960.41906738281 1754.18615722656
1954.07421875 1737.98864746094
1953.8662109375 1723.01147460938
1954.5927734375 1706.4111328125
1958.56750488281 1691.53771972656
1967.919921875 1677.60778808594
1980.44213867188 1667.59436035156
1991.34545898438 1658.43737792969
2002.03503417969 1647.87939453125
2006.63940429688 1635.45385742188
2009.18615722656 1620.64208984375
2007.79858398438 1604.80517578125
2001.56298828125 1590.08435058594
1991.95971679688 1576.00720214844
1979.54284667969 1565.37231445312
1963.68041992188 1558.03576660156
1949.72668457031 1553.59753417969
1934.74645996094 1548.97741699219
1922.70373535156 1540.31640625
1911.51538085938 1528.94567871094
1903.52648925781 1516.44067382812
1897.08642578125 1504.32653808594
1895.259765625 1490.00964355469
1895.83374023438 1475.09191894531
1901.25524902344 1462.7001953125
1908.2158203125 1451.78234863281
1920.14074707031 1441.16198730469
1930.56237792969 1430.05126953125
1938.87524414062 1415.41015625
1942.92797851562 1401.49597167969
1942.14819335938 1385.19567871094
1936.33056640625 1368.61730957031
1925.66333007812 1354.52160644531
1911.88684082031 1341.29138183594
1896.67761230469 1331.44519042969
1879.90222167969 1325.66137695312
1860.92358398438 1325.96899414062
1843.63549804688 1325.61682128906
1824.93078613281 1320.20910644531
1809.16613769531 1313.32702636719
1793.42687988281 1303.22973632812
1780.4755859375 1288.19677734375
1772.42504882812 1272.67504882812
1766.55920410156 1254.60229492188
1766.1474609375 1236.34204101562
1766.04919433594 1219.06945800781
1772.51257324219 1201.11779785156
1777.86437988281 1182.59692382812
1780.53344726562 1162.06665039062
1777.36767578125 1142.11462402344
1768.80029296875 1121.94604492188
1756.58361816406 1103.86950683594
1738.48425292969 1090.4912109375
1719.48352050781 1081.732421875
1699.06262207031 1077.14477539062
1676.86828613281 1078.54077148438
1656.68395996094 1082.64367675781
1635.39147949219 1088.83996582031
1611.58569335938 1088.94006347656
1588.20910644531 1082.20092773438
1567.30517578125 1071.38110351562
1548.99829101562 1056.10827636719
1533.46789550781 1036.51794433594
1524.16418457031 1014.56231689453
1519.181640625 990.139221191406
1520.96472167969 963.453552246094
1528.771484375 941.506042480469
1535.94018554688 918.9306640625
1536.22619628906 894.749389648438
1529.61096191406 869.798583984375
1519.60815429688 848.162353515625
1503.10205078125 830.268798828125
1484.37939453125 816.672912597656
1461.80322265625 807.482116699219
1437.76892089844 804.311340332031
1413.8291015625 807.784973144531
1391.32922363281 816.758361816406
1370.52014160156 827.239868164062
};
\addlegendentry{True Trajectory}
\end{axis}

\end{tikzpicture}

%% file: figures/velocities_pred_hist_100bins_log.tikz
\begin{tikzpicture}

\definecolor{darkgray176}{RGB}{176,176,176}

\begin{axis}[
	width=0.49\linewidth,
	height=0.5\linewidth,
	log basis y={10},
	tick align=outside,
	tick pos=left,
	x grid style={darkgray176},
	xlabel={\footnotesize Speed $\left[\si{\metre\per\second}\right]$},
	xmin=0, xmax=16.5, 
	xtick style={color=black},
	y grid style={darkgray176},
	ymin=0.1, ymax=20252.2611111641,
	ymode=log,
	ytick style={color=black}
]
\draw[draw=none,fill=teal] (axis cs:0,0.00001) rectangle (axis cs:0.141675382852554,12630);
\draw[draw=none,fill=teal] (axis cs:0.141675382852554,0.00001) rectangle (axis cs:0.283350765705109,1162);
\draw[draw=none,fill=teal] (axis cs:0.283350765705109,0.00001) rectangle (axis cs:0.425026118755341,1827);
\draw[draw=none,fill=teal] (axis cs:0.425026118755341,0.00001) rectangle (axis cs:0.566701531410217,7697);
\draw[draw=none,fill=teal] (axis cs:0.566701531410217,0.00001) rectangle (axis cs:0.708376884460449,3619);
\draw[draw=none,fill=teal] (axis cs:0.708376884460449,0.00001) rectangle (axis cs:0.850052237510681,6304);
\draw[draw=none,fill=teal] (axis cs:0.850052237510681,0.00001) rectangle (axis cs:0.991727650165558,11419);
\draw[draw=none,fill=teal] (axis cs:0.991727590560913,0.00001) rectangle (axis cs:1.13340306282043,4514);
\draw[draw=none,fill=teal] (axis cs:1.13340306282043,0.00001) rectangle (axis cs:1.27507841587067,1789);
\draw[draw=none,fill=teal] (axis cs:1.27507829666138,0.00001) rectangle (axis cs:1.41675364971161,2171);
\draw[draw=none,fill=teal] (axis cs:1.4167537689209,0.00001) rectangle (axis cs:1.55842912197113,1548);
\draw[draw=none,fill=teal] (axis cs:1.55842900276184,0.00001) rectangle (axis cs:1.70010435581207,1274);
\draw[draw=none,fill=teal] (axis cs:1.70010447502136,0.00001) rectangle (axis cs:1.84177994728088,1558);
\draw[draw=none,fill=teal] (axis cs:1.84177994728088,0.00001) rectangle (axis cs:1.98345530033112,1954);
\draw[draw=none,fill=teal] (axis cs:1.98345518112183,0.00001) rectangle (axis cs:2.12513065338135,2002);
\draw[draw=none,fill=teal] (axis cs:2.12513065338135,0.00001) rectangle (axis cs:2.26680612564087,1289);
\draw[draw=none,fill=teal] (axis cs:2.26680612564087,0.00001) rectangle (axis cs:2.40848135948181,1421);
\draw[draw=none,fill=teal] (axis cs:2.40848135948181,0.00001) rectangle (axis cs:2.55015683174133,2040);
\draw[draw=none,fill=teal] (axis cs:2.55015707015991,0.00001) rectangle (axis cs:2.69183230400085,1412);
\draw[draw=none,fill=teal] (axis cs:2.69183206558228,0.00001) rectangle (axis cs:2.8335075378418,496);
\draw[draw=none,fill=teal] (axis cs:2.8335075378418,0.00001) rectangle (axis cs:2.97518301010132,180);
\draw[draw=none,fill=teal] (axis cs:2.97518301010132,0.00001) rectangle (axis cs:3.11685824394226,148);
\draw[draw=none,fill=teal] (axis cs:3.11685824394226,0.00001) rectangle (axis cs:3.25853371620178,74);
\draw[draw=none,fill=teal] (axis cs:3.25853395462036,0.00001) rectangle (axis cs:3.4002091884613,82);
\draw[draw=none,fill=teal] (axis cs:3.40020895004272,0.00001) rectangle (axis cs:3.54188442230225,37);
\draw[draw=none,fill=teal] (axis cs:3.54188442230225,0.00001) rectangle (axis cs:3.68355989456177,1);
\draw[draw=none,fill=teal] (axis cs:3.68355989456177,0.00001) rectangle (axis cs:3.82523512840271,2);
\draw[draw=none,fill=teal] (axis cs:3.82523512840271,0.00001) rectangle (axis cs:3.96691060066223,5);
\draw[draw=none,fill=teal] (axis cs:3.96691083908081,0.00001) rectangle (axis cs:4.10858631134033,0.00001);
\draw[draw=none,fill=teal] (axis cs:4.10858535766602,0.00001) rectangle (axis cs:4.25026082992554,0.00001);
\draw[draw=none,fill=teal] (axis cs:4.2502613067627,0.00001) rectangle (axis cs:4.39193677902222,3);
\draw[draw=none,fill=teal] (axis cs:4.39193630218506,0.00001) rectangle (axis cs:4.53361177444458,0.00001);
\draw[draw=none,fill=teal] (axis cs:4.53361225128174,0.00001) rectangle (axis cs:4.6752872467041,1);
\draw[draw=none,fill=teal] (axis cs:4.6752872467041,0.00001) rectangle (axis cs:4.81696271896362,1);
\draw[draw=none,fill=teal] (axis cs:4.81696224212646,0.00001) rectangle (axis cs:4.95863771438599,2);
\draw[draw=none,fill=teal] (axis cs:4.95863819122314,0.00001) rectangle (axis cs:5.10031366348267,0.00001);
\draw[draw=none,fill=teal] (axis cs:5.10031318664551,0.00001) rectangle (axis cs:5.24198865890503,1);
\draw[draw=none,fill=teal] (axis cs:5.24198913574219,0.00001) rectangle (axis cs:5.38366413116455,0.00001);
\draw[draw=none,fill=teal] (axis cs:5.38366413116455,0.00001) rectangle (axis cs:5.52533960342407,0.00001);
\draw[draw=none,fill=teal] (axis cs:5.52533912658691,0.00001) rectangle (axis cs:5.66701459884644,0.00001);
\draw[draw=none,fill=teal] (axis cs:5.66701507568359,0.00001) rectangle (axis cs:5.80869054794312,0.00001);
\draw[draw=none,fill=teal] (axis cs:5.80869007110596,0.00001) rectangle (axis cs:5.95036554336548,0.00001);
\draw[draw=none,fill=teal] (axis cs:5.95036602020264,0.00001) rectangle (axis cs:6.092041015625,0.00001);
\draw[draw=none,fill=teal] (axis cs:6.092041015625,0.00001) rectangle (axis cs:6.23371648788452,0.00001);
\draw[draw=none,fill=teal] (axis cs:6.23371601104736,0.00001) rectangle (axis cs:6.37539148330688,0.00001);
\draw[draw=none,fill=teal] (axis cs:6.37539196014404,0.00001) rectangle (axis cs:6.51706743240356,0.00001);
\draw[draw=none,fill=teal] (axis cs:6.51706695556641,0.00001) rectangle (axis cs:6.65874242782593,0.00001);
\draw[draw=none,fill=teal] (axis cs:6.65874290466309,0.00001) rectangle (axis cs:6.80041790008545,1);
\draw[draw=none,fill=teal] (axis cs:6.80041790008545,0.00001) rectangle (axis cs:6.94209337234497,0.00001);
\draw[draw=none,fill=teal] (axis cs:6.94209289550781,0.00001) rectangle (axis cs:7.08376836776733,1);
\draw[draw=none,fill=teal] (axis cs:7.08376884460449,0.00001) rectangle (axis cs:7.22544431686401,1);
\draw[draw=none,fill=teal] (axis cs:7.22544384002686,0.00001) rectangle (axis cs:7.36711931228638,0.00001);
\draw[draw=none,fill=teal] (axis cs:7.36711978912354,0.00001) rectangle (axis cs:7.5087947845459,0.00001);
\draw[draw=none,fill=teal] (axis cs:7.5087947845459,0.00001) rectangle (axis cs:7.65047025680542,0.00001);
\draw[draw=none,fill=teal] (axis cs:7.65046977996826,0.00001) rectangle (axis cs:7.79214525222778,0.00001);
\draw[draw=none,fill=teal] (axis cs:7.79214572906494,0.00001) rectangle (axis cs:7.93382120132446,0.00001);
\draw[draw=none,fill=teal] (axis cs:7.9338207244873,0.00001) rectangle (axis cs:8.07549667358398,0.00001);
\draw[draw=none,fill=teal] (axis cs:8.07549667358398,0.00001) rectangle (axis cs:8.21717166900635,0.00001);
\draw[draw=none,fill=teal] (axis cs:8.21717166900635,0.00001) rectangle (axis cs:8.35884761810303,1);
\draw[draw=none,fill=teal] (axis cs:8.35884857177734,0.00001) rectangle (axis cs:8.50052356719971,0.00001);
\draw[draw=none,fill=teal] (axis cs:8.50052261352539,0.00001) rectangle (axis cs:8.64219760894775,0.00001);
\draw[draw=none,fill=teal] (axis cs:8.64219760894775,0.00001) rectangle (axis cs:8.78387355804443,0.00001);
\draw[draw=none,fill=teal] (axis cs:8.78387451171875,0.00001) rectangle (axis cs:8.92554950714111,0.00001);
\draw[draw=none,fill=teal] (axis cs:8.9255485534668,0.00001) rectangle (axis cs:9.06722450256348,0.00001);
\draw[draw=none,fill=teal] (axis cs:9.06722450256348,0.00001) rectangle (axis cs:9.20889949798584,0.00001);
\draw[draw=none,fill=teal] (axis cs:9.20890045166016,0.00001) rectangle (axis cs:9.35057544708252,0.00001);
\draw[draw=none,fill=teal] (axis cs:9.3505744934082,0.00001) rectangle (axis cs:9.49225044250488,0.00001);
\draw[draw=none,fill=teal] (axis cs:9.49225044250488,0.00001) rectangle (axis cs:9.63392543792725,0.00001);
\draw[draw=none,fill=teal] (axis cs:9.63392543792725,0.00001) rectangle (axis cs:9.77560138702393,0.00001);
\draw[draw=none,fill=teal] (axis cs:9.77560234069824,0.00001) rectangle (axis cs:9.91727733612061,0.00001);
\draw[draw=none,fill=teal] (axis cs:9.91727638244629,0.00001) rectangle (axis cs:10.0589513778687,0.00001);
\draw[draw=none,fill=teal] (axis cs:10.0589513778687,0.00001) rectangle (axis cs:10.2006273269653,0.00001);
\draw[draw=none,fill=teal] (axis cs:10.2006282806396,0.00001) rectangle (axis cs:10.342303276062,0.00001);
\draw[draw=none,fill=teal] (axis cs:10.3423023223877,0.00001) rectangle (axis cs:10.4839782714844,0.00001);
\draw[draw=none,fill=teal] (axis cs:10.4839782714844,0.00001) rectangle (axis cs:10.6256532669067,0.00001);
\draw[draw=none,fill=teal] (axis cs:10.6256542205811,0.00001) rectangle (axis cs:10.7673292160034,0.00001);
\draw[draw=none,fill=teal] (axis cs:10.7673282623291,0.00001) rectangle (axis cs:10.9090042114258,0.00001);
\draw[draw=none,fill=teal] (axis cs:10.9090042114258,0.00001) rectangle (axis cs:11.0506792068481,0.00001);
\draw[draw=none,fill=teal] (axis cs:11.0506792068481,0.00001) rectangle (axis cs:11.1923551559448,0.00001);
\draw[draw=none,fill=teal] (axis cs:11.1923561096191,0.00001) rectangle (axis cs:11.3340311050415,0.00001);
\draw[draw=none,fill=teal] (axis cs:11.3340301513672,0.00001) rectangle (axis cs:11.4757051467896,0.00001);
\draw[draw=none,fill=teal] (axis cs:11.4757051467896,0.00001) rectangle (axis cs:11.6173810958862,0.00001);
\draw[draw=none,fill=teal] (axis cs:11.6173820495605,0.00001) rectangle (axis cs:11.7590570449829,0.00001);
\draw[draw=none,fill=teal] (axis cs:11.7590560913086,0.00001) rectangle (axis cs:11.9007320404053,0.00001);
\draw[draw=none,fill=teal] (axis cs:11.9007320404053,0.00001) rectangle (axis cs:12.0424070358276,0.00001);
\draw[draw=none,fill=teal] (axis cs:12.042407989502,0.00001) rectangle (axis cs:12.1840829849243,0.00001);
\draw[draw=none,fill=teal] (axis cs:12.18408203125,0.00001) rectangle (axis cs:12.3257579803467,0.00001);
\draw[draw=none,fill=teal] (axis cs:12.3257579803467,0.00001) rectangle (axis cs:12.467432975769,1);
\draw[draw=none,fill=teal] (axis cs:12.467432975769,0.00001) rectangle (axis cs:12.6091089248657,0.00001);
\draw[draw=none,fill=teal] (axis cs:12.60910987854,0.00001) rectangle (axis cs:12.7507848739624,0.00001);
\draw[draw=none,fill=teal] (axis cs:12.7507839202881,0.00001) rectangle (axis cs:12.8924589157104,0.00001);
\draw[draw=none,fill=teal] (axis cs:12.8924589157104,0.00001) rectangle (axis cs:13.0341348648071,0.00001);
\draw[draw=none,fill=teal] (axis cs:13.0341358184814,0.00001) rectangle (axis cs:13.1758108139038,0.00001);
\draw[draw=none,fill=teal] (axis cs:13.1758098602295,0.00001) rectangle (axis cs:13.3174858093262,0.00001);
\draw[draw=none,fill=teal] (axis cs:13.3174858093262,0.00001) rectangle (axis cs:13.4591608047485,0.00001);
\draw[draw=none,fill=teal] (axis cs:13.4591617584229,0.00001) rectangle (axis cs:13.6008367538452,0.00001);
\draw[draw=none,fill=teal] (axis cs:13.6008358001709,0.00001) rectangle (axis cs:13.7425117492676,0.00001);
\draw[draw=none,fill=teal] (axis cs:13.7425117492676,0.00001) rectangle (axis cs:13.8841867446899,0.00001);
\draw[draw=none,fill=teal] (axis cs:13.8841867446899,0.00001) rectangle (axis cs:14.0258626937866,0.00001);
\draw[draw=none,fill=teal] (axis cs:14.0258636474609,0.00001) rectangle (axis cs:14.1675386428833,1);
\end{axis}

\end{tikzpicture}

%% file: figures/velocities_upd_hist_100bins_log.tikz
\begin{tikzpicture}

\definecolor{darkgray176}{RGB}{176,176,176}

\begin{axis}[
	width=0.49\linewidth,
	height=0.5\linewidth,
	log basis y={10},
	tick align=outside,
	tick pos=left,
	x grid style={darkgray176},
	xlabel={\footnotesize Speed $\left[\si{\metre\per\second}\right]$},
	xmin=0, xmax=16.5, 
	xtick style={color=black},
	y grid style={darkgray176},
	ymin=0.1, ymax=20252.26, 
	ymode=log,
	ytick style={color=black},
	legend style={at={(0.03,0.03)},anchor=south west,legend cell align=left,draw=white!15!black,fill opacity=0.5,draw opacity=1,text opacity=1,font=\scriptsize}
]
\draw[draw=none,fill=teal] (axis cs:0,0.00001) rectangle (axis cs:0.164919167757034,1205);
\draw[draw=none,fill=teal] (axis cs:0.164919167757034,0.00001) rectangle (axis cs:0.329838335514069,39);
\draw[draw=none,fill=teal] (axis cs:0.329838335514069,0.00001) rectangle (axis cs:0.494757503271103,152);
\draw[draw=none,fill=teal] (axis cs:0.494757533073425,0.00001) rectangle (axis cs:0.659676671028137,148);
\draw[draw=none,fill=teal] (axis cs:0.659676671028137,0.00001) rectangle (axis cs:0.824595808982849,154);
\draw[draw=none,fill=teal] (axis cs:0.824595808982849,0.00001) rectangle (axis cs:0.989515006542206,245);
\draw[draw=none,fill=teal] (axis cs:0.989515066146851,0.00001) rectangle (axis cs:1.15443420410156,503);
\draw[draw=none,fill=teal] (axis cs:1.15443420410156,0.00001) rectangle (axis cs:1.31935334205627,449);
\draw[draw=none,fill=teal] (axis cs:1.31935334205627,0.00001) rectangle (axis cs:1.48427248001099,331);
\draw[draw=none,fill=teal] (axis cs:1.48427248001099,0.00001) rectangle (axis cs:1.6491916179657,121);
\draw[draw=none,fill=teal] (axis cs:1.6491916179657,0.00001) rectangle (axis cs:1.8141108751297,152);
\draw[draw=none,fill=teal] (axis cs:1.8141108751297,0.00001) rectangle (axis cs:1.97903001308441,58);
\draw[draw=none,fill=teal] (axis cs:1.97902989387512,0.00001) rectangle (axis cs:2.14394903182983,60);
\draw[draw=none,fill=teal] (axis cs:2.14394950866699,0.00001) rectangle (axis cs:2.3088686466217,62);
\draw[draw=none,fill=teal] (axis cs:2.30886840820312,0.00001) rectangle (axis cs:2.47378754615784,34);
\draw[draw=none,fill=teal] (axis cs:2.47378778457642,0.00001) rectangle (axis cs:2.63870692253113,48);
\draw[draw=none,fill=teal] (axis cs:2.63870668411255,0.00001) rectangle (axis cs:2.80362582206726,38);
\draw[draw=none,fill=teal] (axis cs:2.80362606048584,0.00001) rectangle (axis cs:2.96854519844055,39);
\draw[draw=none,fill=teal] (axis cs:2.96854496002197,0.00001) rectangle (axis cs:3.13346409797668,79);
\draw[draw=none,fill=teal] (axis cs:3.13346433639526,0.00001) rectangle (axis cs:3.29838347434998,49);
\draw[draw=none,fill=teal] (axis cs:3.2983832359314,0.00001) rectangle (axis cs:3.46330261230469,93);
\draw[draw=none,fill=teal] (axis cs:3.46330261230469,0.00001) rectangle (axis cs:3.6282217502594,62);
\draw[draw=none,fill=teal] (axis cs:3.62822198867798,0.00001) rectangle (axis cs:3.79314112663269,61);
\draw[draw=none,fill=teal] (axis cs:3.79314088821411,0.00001) rectangle (axis cs:3.95806002616882,30);
\draw[draw=none,fill=teal] (axis cs:3.9580602645874,0.00001) rectangle (axis cs:4.12297916412354,31);
\draw[draw=none,fill=teal] (axis cs:4.12297916412354,0.00001) rectangle (axis cs:4.28789854049683,39);
\draw[draw=none,fill=teal] (axis cs:4.28789854049683,0.00001) rectangle (axis cs:4.45281744003296,57);
\draw[draw=none,fill=teal] (axis cs:4.45281791687012,0.00001) rectangle (axis cs:4.61773729324341,55);
\draw[draw=none,fill=teal] (axis cs:4.61773681640625,0.00001) rectangle (axis cs:4.78265571594238,46);
\draw[draw=none,fill=teal] (axis cs:4.78265571594238,0.00001) rectangle (axis cs:4.94757509231567,54);
\draw[draw=none,fill=teal] (axis cs:4.94757509231567,0.00001) rectangle (axis cs:5.11249399185181,46);
\draw[draw=none,fill=teal] (axis cs:5.11249446868896,0.00001) rectangle (axis cs:5.27741384506226,52);
\draw[draw=none,fill=teal] (axis cs:5.2774133682251,0.00001) rectangle (axis cs:5.44233226776123,52);
\draw[draw=none,fill=teal] (axis cs:5.44233226776123,0.00001) rectangle (axis cs:5.60725164413452,58);
\draw[draw=none,fill=teal] (axis cs:5.60725212097168,0.00001) rectangle (axis cs:5.77217149734497,73);
\draw[draw=none,fill=teal] (axis cs:5.77217102050781,0.00001) rectangle (axis cs:5.93708992004395,56);
\draw[draw=none,fill=teal] (axis cs:5.93708992004395,0.00001) rectangle (axis cs:6.10200929641724,52);
\draw[draw=none,fill=teal] (axis cs:6.10200929641724,0.00001) rectangle (axis cs:6.26692819595337,82);
\draw[draw=none,fill=teal] (axis cs:6.26692867279053,0.00001) rectangle (axis cs:6.43184804916382,65);
\draw[draw=none,fill=teal] (axis cs:6.43184757232666,0.00001) rectangle (axis cs:6.59676647186279,65);
\draw[draw=none,fill=teal] (axis cs:6.59676647186279,0.00001) rectangle (axis cs:6.76168584823608,61);
\draw[draw=none,fill=teal] (axis cs:6.76168632507324,0.00001) rectangle (axis cs:6.92660570144653,56);
\draw[draw=none,fill=teal] (axis cs:6.92660522460938,0.00001) rectangle (axis cs:7.09152412414551,41);
\draw[draw=none,fill=teal] (axis cs:7.09152412414551,0.00001) rectangle (axis cs:7.2564435005188,58);
\draw[draw=none,fill=teal] (axis cs:7.2564435005188,0.00001) rectangle (axis cs:7.42136240005493,76);
\draw[draw=none,fill=teal] (axis cs:7.42136287689209,0.00001) rectangle (axis cs:7.58628225326538,55);
\draw[draw=none,fill=teal] (axis cs:7.58628177642822,0.00001) rectangle (axis cs:7.75120067596436,73);
\draw[draw=none,fill=teal] (axis cs:7.75120067596436,0.00001) rectangle (axis cs:7.91612005233765,82);
\draw[draw=none,fill=teal] (axis cs:7.9161205291748,0.00001) rectangle (axis cs:8.08103942871094,113);
\draw[draw=none,fill=teal] (axis cs:8.08103942871094,0.00001) rectangle (axis cs:8.24595832824707,125);
\draw[draw=none,fill=teal] (axis cs:8.24595832824707,0.00001) rectangle (axis cs:8.4108772277832,133);
\draw[draw=none,fill=teal] (axis cs:8.4108772277832,0.00001) rectangle (axis cs:8.57579708099365,161);
\draw[draw=none,fill=teal] (axis cs:8.57579708099365,0.00001) rectangle (axis cs:8.74071598052979,149);
\draw[draw=none,fill=teal] (axis cs:8.74071598052979,0.00001) rectangle (axis cs:8.90563488006592,139);
\draw[draw=none,fill=teal] (axis cs:8.90563488006592,0.00001) rectangle (axis cs:9.07055377960205,192);
\draw[draw=none,fill=teal] (axis cs:9.07055282592773,0.00001) rectangle (axis cs:9.23547267913818,174);
\draw[draw=none,fill=teal] (axis cs:9.2354736328125,0.00001) rectangle (axis cs:9.40039253234863,177);
\draw[draw=none,fill=teal] (axis cs:9.40039253234863,0.00001) rectangle (axis cs:9.56531143188477,176);
\draw[draw=none,fill=teal] (axis cs:9.56531143188477,0.00001) rectangle (axis cs:9.73023128509521,178);
\draw[draw=none,fill=teal] (axis cs:9.73023128509521,0.00001) rectangle (axis cs:9.89515018463135,201);
\draw[draw=none,fill=teal] (axis cs:9.89515018463135,0.00001) rectangle (axis cs:10.0600690841675,210);
\draw[draw=none,fill=teal] (axis cs:10.0600690841675,0.00001) rectangle (axis cs:10.2249879837036,213);
\draw[draw=none,fill=teal] (axis cs:10.2249870300293,0.00001) rectangle (axis cs:10.3899068832397,242);
\draw[draw=none,fill=teal] (axis cs:10.3899078369141,0.00001) rectangle (axis cs:10.5548267364502,269);
\draw[draw=none,fill=teal] (axis cs:10.5548267364502,0.00001) rectangle (axis cs:10.7197456359863,259);
\draw[draw=none,fill=teal] (axis cs:10.7197456359863,0.00001) rectangle (axis cs:10.8846645355225,250);
\draw[draw=none,fill=teal] (axis cs:10.8846645355225,0.00001) rectangle (axis cs:11.0495843887329,197);
\draw[draw=none,fill=teal] (axis cs:11.0495843887329,0.00001) rectangle (axis cs:11.214503288269,159);
\draw[draw=none,fill=teal] (axis cs:11.214503288269,0.00001) rectangle (axis cs:11.3794221878052,137);
\draw[draw=none,fill=teal] (axis cs:11.3794212341309,0.00001) rectangle (axis cs:11.5443410873413,112);
\draw[draw=none,fill=teal] (axis cs:11.5443420410156,0.00001) rectangle (axis cs:11.7092609405518,84);
\draw[draw=none,fill=teal] (axis cs:11.7092609405518,0.00001) rectangle (axis cs:11.8741798400879,57);
\draw[draw=none,fill=teal] (axis cs:11.8741798400879,0.00001) rectangle (axis cs:12.039098739624,70);
\draw[draw=none,fill=teal] (axis cs:12.039098739624,0.00001) rectangle (axis cs:12.2040185928345,54);
\draw[draw=none,fill=teal] (axis cs:12.2040185928345,0.00001) rectangle (axis cs:12.3689374923706,42);
\draw[draw=none,fill=teal] (axis cs:12.3689374923706,0.00001) rectangle (axis cs:12.5338563919067,37);
\draw[draw=none,fill=teal] (axis cs:12.5338554382324,0.00001) rectangle (axis cs:12.6987752914429,44);
\draw[draw=none,fill=teal] (axis cs:12.6987762451172,0.00001) rectangle (axis cs:12.8636951446533,54);
\draw[draw=none,fill=teal] (axis cs:12.8636951446533,0.00001) rectangle (axis cs:13.0286140441895,41);
\draw[draw=none,fill=teal] (axis cs:13.0286140441895,0.00001) rectangle (axis cs:13.1935329437256,63);
\draw[draw=none,fill=teal] (axis cs:13.1935329437256,0.00001) rectangle (axis cs:13.358452796936,95);
\draw[draw=none,fill=teal] (axis cs:13.358452796936,0.00001) rectangle (axis cs:13.5233716964722,86);
\draw[draw=none,fill=teal] (axis cs:13.5233716964722,0.00001) rectangle (axis cs:13.6882905960083,55);
\draw[draw=none,fill=teal] (axis cs:13.688289642334,0.00001) rectangle (axis cs:13.8532094955444,32);
\draw[draw=none,fill=teal] (axis cs:13.8532104492188,0.00001) rectangle (axis cs:14.0181293487549,26);
\draw[draw=none,fill=teal] (axis cs:14.0181293487549,0.00001) rectangle (axis cs:14.183048248291,30);
\draw[draw=none,fill=teal] (axis cs:14.183048248291,0.00001) rectangle (axis cs:14.3479671478271,33);
\draw[draw=none,fill=teal] (axis cs:14.3479671478271,0.00001) rectangle (axis cs:14.5128870010376,40);
\draw[draw=none,fill=teal] (axis cs:14.5128870010376,0.00001) rectangle (axis cs:14.6778059005737,18);
\draw[draw=none,fill=teal] (axis cs:14.6778059005737,0.00001) rectangle (axis cs:14.8427248001099,21);
\draw[draw=none,fill=teal] (axis cs:14.8427248001099,0.00001) rectangle (axis cs:15.007643699646,20);
\draw[draw=none,fill=teal] (axis cs:15.0076427459717,0.00001) rectangle (axis cs:15.1725625991821,15);
\draw[draw=none,fill=teal] (axis cs:15.1725635528564,0.00001) rectangle (axis cs:15.3374824523926,36);
\draw[draw=none,fill=teal] (axis cs:15.3374824523926,0.00001) rectangle (axis cs:15.5024013519287,50);
\draw[draw=none,fill=teal] (axis cs:15.5024013519287,0.00001) rectangle (axis cs:15.6673212051392,78);
\draw[draw=none,fill=teal] (axis cs:15.6673212051392,0.00001) rectangle (axis cs:15.8322401046753,84);
\draw[draw=none,fill=teal] (axis cs:15.8322401046753,0.00001) rectangle (axis cs:15.9971590042114,69);
\draw[draw=none,fill=teal] (axis cs:15.9971580505371,0.00001) rectangle (axis cs:16.1620788574219,34);
\draw[draw=none,fill=teal] (axis cs:16.1620788574219,0.00001) rectangle (axis cs:16.326997756958,10);
\draw[draw=none,fill=teal] (axis cs:16.3269996643066,0.00001) rectangle (axis cs:16.4919185638428,3);
\end{axis}

\end{tikzpicture}

%% file: figures/trajectory_sample59_test.tikz
\begin{tikzpicture}

\definecolor{darkgray176}{RGB}{176,176,176}
\definecolor{green}{RGB}{0,128,0}
\definecolor{lightgray204}{RGB}{204,204,204}
\definecolor{purple}{RGB}{128,0,128}

\begin{axis}[
	width=0.9\linewidth,
	height=0.8\linewidth,
	legend cell align={left},
	legend style={fill opacity=0.8, draw opacity=1, text opacity=1, draw=lightgray204},
	tick align=outside,
	tick pos=left,
	x grid style={darkgray176},
	xlabel={\footnotesize Rel. East Position [km]},
	xmin=1300.280151367188, xmax=2168.88317871094,
	xtick style={color=black},
	xtick={-250,0,250,500,750,1000,1250,1500,1750,2000,2250},
	xticklabels={-0.25,0,0.25,0.5,0.75,1,1.25,1.5,1.75,2,2.25},
	y grid style={darkgray176},
	ylabel={\footnotesize Rel. North Position [km]},
	ymin=750.810217285156, ymax=2068.01456298828,
	ytick style={color=black},
	ytick={-250,0,250,500,750,1000,1250,1500,1750,2000,2250},
	yticklabels={-0.25,0,0.25,0.5,0.75,1,1.25,1.5,1.75,2,2.25},
	legend style={at={(0.03,0.97)},anchor=north west,legend cell align=left,draw=white!15!black,fill opacity=0.5,draw opacity=1,text opacity=1,font=\scriptsize}
]
\addplot [draw=green, fill=green, mark=x, only marks, opacity=0.5]
table{%
x  y
2041.49072265625 2036.20434570312
1994.41174316406 1986.5615234375
2018.38012695312 1999.10900878906
1983.87414550781 1951.42785644531
2006.52575683594 1962.98779296875
1977.35021972656 1914.90307617188
2018.52124023438 1926.49890136719
2048.23852539062 1942.57800292969
2023.24072265625 1885.34814453125
2029.05639648438 1872.44555664062
2057.98510742188 1873.66882324219
2036.57702636719 1841.94006347656
2057.5234375 1848.62060546875
2063.52905273438 1836.95678710938
2038.76086425781 1813.91918945312
2054.5537109375 1824.6513671875
2034.30041503906 1805.58264160156
2056.54174804688 1839.16357421875
1980.87475585938 1776.40185546875
1959.51428222656 1759.6806640625
1963.10168457031 1764.08862304688
1954.34228515625 1754.69396972656
1967.59997558594 1746.43908691406
1948.90612792969 1726.58557128906
1970.64111328125 1722.63659667969
1967.36743164062 1697.697265625
1968.99462890625 1675.03210449219
1991.67736816406 1673.67456054688
2021.67260742188 1683.20288085938
2010.24768066406 1652.23522949219
2021.73413085938 1648.74487304688
1978.26318359375 1591.12353515625
1979.40563964844 1581.37890625
1988.71142578125 1580.59777832031
1972.95678710938 1567.31616210938
1992.65734863281 1574.11889648438
1941.13122558594 1539.29187011719
1955.45043945312 1551.56213378906
1933.34118652344 1549.20495605469
1922.4072265625 1534.10546875
1921.45153808594 1539.68249511719
1898.28723144531 1509.24108886719
1898.57641601562 1506.72338867188
1914.55883789062 1503.50036621094
1916.18762207031 1489.10754394531
1925.22937011719 1484.43823242188
1896.60266113281 1441.19702148438
1904.70446777344 1430.59655761719
1931.96997070312 1431.9091796875
1917.65905761719 1403.9833984375
1920.17529296875 1388.30981445312
1962.48962402344 1403.26953125
1932.56335449219 1368.60791015625
1942.94079589844 1365.61511230469
1895.74340820312 1333.48327636719
1865.02172851562 1308.45056152344
1894.30139160156 1339.04907226562
1841.65124511719 1314.4267578125
1852.78820800781 1335.57690429688
1797.05102539062 1297.71594238281
1799.92456054688 1310.7021484375
1816.91259765625 1316.12353515625
1790.18469238281 1295.41540527344
1766.34326171875 1267.36828613281
1725.35400390625 1221.03540039062
1793.96667480469 1254.74877929688
1773.00451660156 1224.70959472656
1767.76330566406 1195.31481933594
1754.20166015625 1169.11083984375
1780.83117675781 1167.52661132812
1805.72729492188 1158.62927246094
1748.68237304688 1106.70263671875
1760.82385253906 1107.98559570312
1755.76025390625 1098.4970703125
1690.45385742188 1065.44226074219
1682.23901367188 1068.32043457031
1674.42565917969 1077.18286132812
1654.61755371094 1081.38549804688
1643.38757324219 1098.46691894531
1588.66174316406 1076.7490234375
1559.29016113281 1064.39965820312
1571.23828125 1073.79541015625
1506.65466308594 1030.18908691406
1518.16882324219 1027.591796875
1489.11157226562 989.350524902344
1527.85412597656 995.283813476562
1541.28491210938 977.044677734375
1536.74426269531 947.114990234375
1551.49731445312 925.126159667969
1566.43469238281 913.577575683594
1518.28771972656 859.593322753906
1539.40881347656 860.048645019531
1554.28076171875 857.866638183594
1485.41333007812 819.2783203125
1462.52880859375 810.693969726562
1450.48608398438 812.016235351562
1393.06860351562 795.722778320312
1402.09692382812 823.46240234375
1351.21423339844 817.619018554688
};
\addlegendentry{Measurements}
\addplot [semithick, black, opacity=0.9]
table {%
2009.90441894531 2005.38220214844
2003.88806152344 1993.24584960938
1999.18041992188 1980.61291503906
1997.56640625 1965.32250976562
1998.34436035156 1951.33483886719
2005.33374023438 1938.61450195312
2014.14562988281 1925.84875488281
2025.28674316406 1915.41101074219
2036.58837890625 1903.34094238281
2047.17272949219 1890.49890136719
2053.33129882812 1876.42651367188
2055.09326171875 1861.34130859375
2055.25390625 1847.57653808594
2051.64697265625 1833.38012695312
2044.23083496094 1820.40966796875
2034.85412597656 1809.9208984375
2023.54956054688 1800.92700195312
2008.99487304688 1794.43420410156
1993.48132324219 1789.16796875
1980.662109375 1779.43090820312
1969.09191894531 1767.56091308594
1960.41906738281 1754.18615722656
1954.07421875 1737.98864746094
1953.8662109375 1723.01147460938
1954.5927734375 1706.4111328125
1958.56750488281 1691.53771972656
1967.919921875 1677.60778808594
1980.44213867188 1667.59436035156
1991.34545898438 1658.43737792969
2002.03503417969 1647.87939453125
2006.63940429688 1635.45385742188
2009.18615722656 1620.64208984375
2007.79858398438 1604.80517578125
2001.56298828125 1590.08435058594
1991.95971679688 1576.00720214844
1979.54284667969 1565.37231445312
1963.68041992188 1558.03576660156
1949.72668457031 1553.59753417969
1934.74645996094 1548.97741699219
1922.70373535156 1540.31640625
1911.51538085938 1528.94567871094
1903.52648925781 1516.44067382812
1897.08642578125 1504.32653808594
1895.259765625 1490.00964355469
1895.83374023438 1475.09191894531
1901.25524902344 1462.7001953125
1908.2158203125 1451.78234863281
1920.14074707031 1441.16198730469
1930.56237792969 1430.05126953125
1938.87524414062 1415.41015625
1942.92797851562 1401.49597167969
1942.14819335938 1385.19567871094
1936.33056640625 1368.61730957031
1925.66333007812 1354.52160644531
1911.88684082031 1341.29138183594
1896.67761230469 1331.44519042969
1879.90222167969 1325.66137695312
1860.92358398438 1325.96899414062
1843.63549804688 1325.61682128906
1824.93078613281 1320.20910644531
1809.16613769531 1313.32702636719
1793.42687988281 1303.22973632812
1780.4755859375 1288.19677734375
1772.42504882812 1272.67504882812
1766.55920410156 1254.60229492188
1766.1474609375 1236.34204101562
1766.04919433594 1219.06945800781
1772.51257324219 1201.11779785156
1777.86437988281 1182.59692382812
1780.53344726562 1162.06665039062
1777.36767578125 1142.11462402344
1768.80029296875 1121.94604492188
1756.58361816406 1103.86950683594
1738.48425292969 1090.4912109375
1719.48352050781 1081.732421875
1699.06262207031 1077.14477539062
1676.86828613281 1078.54077148438
1656.68395996094 1082.64367675781
1635.39147949219 1088.83996582031
1611.58569335938 1088.94006347656
1588.20910644531 1082.20092773438
1567.30517578125 1071.38110351562
1548.99829101562 1056.10827636719
1533.46789550781 1036.51794433594
1524.16418457031 1014.56231689453
1519.181640625 990.139221191406
1520.96472167969 963.453552246094
1528.771484375 941.506042480469
1535.94018554688 918.9306640625
1536.22619628906 894.749389648438
1529.61096191406 869.798583984375
1519.60815429688 848.162353515625
1503.10205078125 830.268798828125
1484.37939453125 816.672912597656
1461.80322265625 807.482116699219
1437.76892089844 804.311340332031
1413.8291015625 807.784973144531
1391.32922363281 816.758361816406
1370.52014160156 827.239868164062
};
\addlegendentry{True Trajectory}
\addplot [semithick, blue, opacity=0.9]
table {%
2035.33142089844 2027.498046875
2034.87902832031 2017.73840332031
2025.53576660156 1994.04382324219
2019.19116210938 1975.392578125
2003.95678710938 1942.23791503906
2009.06372070312 1920.31628417969
2023.4501953125 1917.16247558594
2027.12719726562 1891.69763183594
2030.05554199219 1872.33959960938
2043.5234375 1860.72265625
2042.58081054688 1844.95922851562
2048.60595703125 1838.91906738281
2056.1826171875 1830.84411621094
2049.21899414062 1820.45825195312
2049.34619140625 1818.57592773438
2041.37072753906 1811.02099609375
2041.88708496094 1822.564453125
2013.767578125 1805.70532226562
1986.19946289062 1784.71704101562
1969.07775878906 1770.45288085938
1955.42822265625 1756.6533203125
1956.17565917969 1740.56896972656
1948.87915039062 1725.61804199219
1955.6572265625 1712.13037109375
1960.85083007812 1693.70422363281
1965.91235351562 1673.32421875
1977.89880371094 1661.62390136719
1997.3076171875 1661.34545898438
2007.34521484375 1650.19262695312
2017.22424316406 1643.8642578125
2007.32055664062 1615.6171875
1997.53063964844 1594.48266601562
1992.181640625 1581.75939941406
1980.39721679688 1571.01904296875
1981.44262695312 1565.80090332031
1960.42333984375 1552.45886230469
1951.46691894531 1547.970703125
1935.04296875 1547.89196777344
1924.30419921875 1538.3203125
1916.20385742188 1535.16259765625
1905.61071777344 1517.880859375
1898.04968261719 1506.28283691406
1901.87915039062 1495.53247070312
1906.81604003906 1482.98889160156
1913.87365722656 1475.21374511719
1909.185546875 1452.10656738281
1908.16064453125 1433.2255859375
1918.23767089844 1421.92297363281
1919.77124023438 1405.31750488281
1921.39086914062 1389.06811523438
1938.73986816406 1385.83801269531
1939.09387207031 1372.59338378906
1942.05493164062 1364.13403320312
1922.69641113281 1349.97924804688
1895.9677734375 1329.76684570312
1886.19641113281 1330.98901367188
1858.33984375 1324.62841796875
1844.14990234375 1328.72009277344
1816.83654785156 1313.90490722656
1798.69567871094 1309.45202636719
1797.94311523438 1305.1328125
1788.66662597656 1294.89587402344
1775.35107421875 1275.32446289062
1751.71545410156 1241.41137695312
1762.30078125 1232.8544921875
1763.47875976562 1218.0595703125
1764.80480957031 1195.04309082031
1759.91162109375 1172.35876464844
1766.73950195312 1158.03637695312
1783.103515625 1145.08508300781
1771.58081054688 1120.17272949219
1764.80578613281 1108.2529296875
1757.24169921875 1098.40100097656
1723.61767578125 1083.93994140625
1695.38317871094 1075.63586425781
1672.86682128906 1075.13195800781
1650.75048828125 1078.03356933594
1631.75207519531 1089.28820800781
1601.54455566406 1086.44763183594
1572.51379394531 1075.2021484375
1560.41027832031 1068.07995605469
1528.51672363281 1045.65197753906
1514.27294921875 1026.55249023438
1495.85766601562 995.485900878906
1501.85986328125 978.888916015625
1515.80102539062 961.890380859375
1525.63830566406 940.626098632812
1539.53833007812 918.576232910156
1554.67260742188 905.372497558594
1543.76611328125 874.273681640625
1542.349609375 859.901000976562
1545.4765625 852.079406738281
1517.185546875 835.477111816406
1484.931640625 822.243530273438
1457.41735839844 815.263549804688
1413.94763183594 806.624755859375
1388.87426757812 814.673767089844
};
\addlegendentry{Opt. IMM}
\addplot [semithick, green, opacity=0.9]
table {%
2036.07873535156 2028.25122070312
2035.57995605469 2019.80102539062
2032.31201171875 2001.79675292969
2031.54797363281 1986.53869628906
2028.44812011719 1965.97180175781
2032.47717285156 1944.73889160156
2040.95922851562 1933.72985839844
2048.7548828125 1914.009765625
2055.2822265625 1896.52209472656
2062.5751953125 1879.31909179688
2065.60302734375 1865.27661132812
2063.81640625 1852.37133789062
2059.22509765625 1835.81469726562
2052.83422851562 1823.76696777344
2039.45446777344 1810.35803222656
2023.59692382812 1795.82409667969
2009.12158203125 1791.1318359375
1991.91967773438 1785.08337402344
1973.35070800781 1773.90051269531
1958.53283691406 1760.62634277344
1947.57446289062 1747.45495605469
1947.09521484375 1731.81616210938
1949.48400878906 1723.04809570312
1956.88159179688 1712.05395507812
1967.40808105469 1699.3408203125
1977.6416015625 1683.83154296875
1987.41650390625 1669.88879394531
1997.86975097656 1661.44201660156
2010.39916992188 1653.07934570312
2018.94140625 1644.83056640625
2021.58044433594 1628.23352050781
2015.76904296875 1609.52563476562
1999.61853027344 1589.30993652344
1980.3974609375 1571.98620605469
1968.65612792969 1557.7353515625
1956.71032714844 1547.88427734375
1943.1572265625 1540.93701171875
1926.09057617188 1539.5224609375
1917.03833007812 1534.90795898438
1907.45910644531 1527.52941894531
1902.61706542969 1515.7265625
1897.4765625 1503.96960449219
1897.12292480469 1491.080078125
1902.49145507812 1479.42004394531
1908.95874023438 1470.69555664062
1914.46130371094 1456.38488769531
1918.05969238281 1441.06188964844
1923.5048828125 1426.78015136719
1928.10034179688 1412.45861816406
1931.40307617188 1397.3056640625
1937.72375488281 1385.9912109375
1942.97497558594 1375.95642089844
1945.1923828125 1366.35900878906
1937.25170898438 1358.99060058594
1914.3212890625 1342.60888671875
1889.59680175781 1332.94775390625
1862.23413085938 1327.97998046875
1841.19677734375 1327.62890625
1822.822265625 1320.02502441406
1800.46301269531 1307.97229003906
1787.56909179688 1297.92651367188
1777.14770507812 1285.56750488281
1767.71411132812 1269.15502929688
1757.44201660156 1245.19555664062
1755.88977050781 1227.19140625
1761.30725097656 1215.71142578125
1768.89282226562 1198.326171875
1773.20251464844 1181.255859375
1776.81982421875 1164.88684082031
1785.12426757812 1147.55419921875
1788.39587402344 1131.44567871094
1781.63793945312 1118.13330078125
1763.14782714844 1102.25573730469
1735.03234863281 1090.63684082031
1707.62109375 1084.77734375
1681.45458984375 1081.2275390625
1657.0244140625 1082.03198242188
1633.41223144531 1089.42858886719
1612.11682128906 1093.41284179688
1589.00012207031 1087.00622558594
1569.806640625 1073.56127929688
1546.77124023438 1056.31774902344
1528.07080078125 1036.3720703125
1512.5986328125 1007.02972412109
1507.47338867188 982.490112304688
1514.673828125 961.376831054688
1523.67993164062 939.768493652344
1535.77819824219 916.946411132812
1547.95874023438 901.427368164062
1557.77172851562 882.671569824219
1558.330078125 867.943054199219
1546.82824707031 852.451843261719
1522.35974121094 837.775329589844
1494.92150878906 828.305541992188
1467.52453613281 821.479309082031
1435.89379882812 818.842224121094
1407.2783203125 824.785583496094
};
\addlegendentry{MKF}
\addplot [semithick, red, opacity=0.9]
table {%
2035.0673828125 2027.2275390625
2033.92785644531 2016.0498046875
2020.13635253906 1987.92199707031
2012.18225097656 1968.24609375
1994.98083496094 1933.02624511719
2003.79931640625 1914.23522949219
2022.61267089844 1916.43347167969
2027.43615722656 1891.37048339844
2030.71655273438 1872.94665527344
2046.01452636719 1863.03149414062
2043.95861816406 1846.97985839844
2050.65893554688 1841.65698242188
2058.875 1833.42504882812
2049.90014648438 1821.884765625
2050.1708984375 1820.13171386719
2041.13513183594 1811.33703613281
2042.97924804688 1824.71691894531
2011.12451171875 1803.86694335938
1980.90307617188 1779.6640625
1963.94860839844 1765.32385253906
1951.31652832031 1752.45849609375
1954.88073730469 1737.95715332031
1948.18579101562 1724.67272949219
1957.41870117188 1712.96459960938
1963.47729492188 1695.16760253906
1968.50952148438 1674.98449707031
1981.3154296875 1664.50512695312
2002.5078125 1666.24475097656
2011.92919921875 1654.12426757812
2021.01611328125 1647.31750488281
2006.90515136719 1615.0400390625
1994.25842285156 1592.158203125
1988.43225097656 1579.45812988281
1976.33056640625 1568.76110839844
1979.80261230469 1564.69592285156
1957.22119140625 1550.56188964844
1949.60400390625 1546.90588378906
1933.5615234375 1547.65234375
1923.69934082031 1537.34033203125
1916.55932617188 1535.16345214844
1905.55603027344 1516.86962890625
1897.97326660156 1505.82958984375
1903.423828125 1496.0908203125
1909.27648925781 1484.27990722656
1917.0166015625 1477.60290527344
1910.40600585938 1452.65539550781
1908.40283203125 1433.2265625
1919.68676757812 1422.9130859375
1920.58312988281 1405.970703125
1921.74768066406 1389.40393066406
1941.59338378906 1388.04479980469
1940.68395996094 1373.99914550781
1943.21374511719 1365.33081054688
1920.09252929688 1349.13903808594
1890.08996582031 1326.12805175781
1882.06030273438 1328.88073730469
1853.35791015625 1321.95007324219
1841.58679199219 1327.345703125
1813.27331542969 1310.73229980469
1796.19262695312 1307.43054199219
1798.79382324219 1304.88879394531
1790.03723144531 1295.33715820312
1775.76477050781 1275.00561523438
1749.13879394531 1238.77795410156
1763.50964355469 1233.34423828125
1765.68994140625 1219.58129882812
1767.294921875 1196.18054199219
1761.18579101562 1173.17651367188
1769.06726074219 1159.71838378906
1787.60534667969 1147.73059082031
1772.46142578125 1120.98937988281
1764.24206542969 1108.82275390625
1756.41711425781 1098.57092285156
1718.97399902344 1082.05969238281
1690.25756835938 1073.04699707031
1669.20874023438 1073.25817871094
1648.66259765625 1077.05615234375
1631.708984375 1089.81469726562
1600.88232421875 1085.81140136719
1570.85119628906 1073.28125
1560.38391113281 1067.26245117188
1526.62854003906 1044.01184082031
1513.32385253906 1025.34448242188
1495.05236816406 994.251586914062
1503.17150878906 979.424987792969
1518.50073242188 963.269104003906
1528.25854492188 942.04931640625
1542.18884277344 919.943176269531
1556.94702148438 907.067626953125
1543.89538574219 874.392333984375
1541.55065917969 860.036987304688
1545.3994140625 852.525695800781
1514.82019042969 834.75390625
1482.11315917969 821.078186035156
1455.42736816406 814.416076660156
1411.76245117188 805.782958984375
1388.42797851562 814.801574707031
};
\addlegendentry{EKF}
\addplot [semithick, purple, opacity=0.9]
table[row sep=crcr]{%
	2001.26935060438	2006.55414650492\\
	1993.9694678546	1987.76219773964\\
	2001.45160239941	1978.426663182\\
	1988.25476991453	1969.32275935234\\
	1987.14627230589	1955.85688800287\\
	1993.98074438839	1939.39774849712\\
	2013.04204259787	1926.5167147457\\
	2031.93335614191	1917.97841957898\\
	2043.46666480384	1899.77391129433\\
	2049.18789896419	1893.00323000486\\
	2045.27024032908	1883.18488602791\\
	2048.5005279003	1871.72194305135\\
	2049.64127344329	1854.43942131253\\
	2045.37779453657	1839.46873244843\\
	2043.3889755487	1821.5748753055\\
	2039.56713076747	1810.94233317308\\
	2036.90324416214	1810.3369374643\\
	2034.11554005541	1799.17398866861\\
	2009.93915808612	1787.6244029102\\
	1983.96006289044	1784.74753826884\\
	1963.23942467763	1776.0146325041\\
	1955.75426742451	1755.52968229937\\
	1948.56923718981	1737.91985198131\\
	1934.9800607606	1732.9678043698\\
	1943.59473887882	1712.48789632837\\
	1955.71342420744	1696.788274581\\
	1956.51422625273	1677.62376861267\\
	1971.27389685653	1672.20625746313\\
	1979.68743008474	1661.68321347959\\
	1989.82184856008	1650.12891032922\\
	1986.65934159378	1640.93106588765\\
	1995.59552413971	1628.17244377623\\
	2002.01929672407	1617.17600638387\\
	1996.8951332315	1603.66074308018\\
	1991.35245443164	1589.72765661907\\
	1981.35731446683	1579.55873360235\\
	1968.82642683513	1574.03075828364\\
	1966.55511601089	1561.33859688864\\
	1947.05329861589	1549.49585354442\\
	1928.8220655372	1535.58832846727\\
	1920.45742396594	1517.61156360269\\
	1906.09587977129	1517.72274569787\\
	1887.51113871047	1502.80947986609\\
	1883.93199698045	1496.83263380865\\
	1884.04701945751	1483.2333611041\\
	1903.22703513714	1473.66571151304\\
	1905.443275304	1459.10526821105\\
	1914.58339047734	1454.36441463718\\
	1916.3083747881	1447.93893874596\\
	1916.51099018265	1433.30053834239\\
	1926.52933209584	1426.29555440219\\
	1943.64139110768	1411.80238292598\\
	1934.55154546905	1393.74333983359\\
	1920.73300062716	1378.99473474524\\
	1908.54926983377	1361.48777900842\\
	1899.97630880236	1345.09549308754\\
	1886.0723524303	1329.89449434262\\
	1866.07140193685	1316.5298637435\\
	1847.64916246989	1304.34975379454\\
	1837.71012867609	1293.75633171383\\
	1824.85933599853	1291.62493674996\\
	1812.71775244364	1285.99351050868\\
	1801.61788265559	1283.65730085548\\
	1785.62918802728	1283.01267173917\\
	1778.96773280125	1267.37188328256\\
	1766.71305657597	1252.00833062424\\
	1760.11924083337	1238.59165706464\\
	1763.1893497362	1222.03696076281\\
	1759.58806524119	1209.50989157572\\
	1770.70859050417	1197.08228092109\\
	1757.35903157671	1186.71226651575\\
	1744.48709228265	1168.35134785893\\
	1749.29278527786	1148.09866472484\\
	1741.03475461646	1122.94417930474\\
	1719.16302378263	1106.38074824465\\
	1690.12224196768	1101.93087565922\\
	1672.81352235199	1078.69462931303\\
	1661.61426429301	1078.69625133348\\
	1637.10898999825	1078.893434967\\
	1623.36897223003	1072.80480933718\\
	1606.25768274382	1060.6388797527\\
	1587.76734010217	1055.26888475284\\
	1566.37586929415	1050.39231301788\\
	1549.19196464061	1033.13774131073\\
	1527.01253674772	1032.01419404662\\
	1513.87889435563	1018.81904808713\\
	1518.19178745285	976.324598088151\\
	1536.74977857805	940.225563188043\\
	1558.25673696577	921.027863638922\\
	1537.9114259395	909.331371275931\\
	1524.00627952256	869.981104107333\\
	1498.30441480752	861.892204925033\\
	1487.06006301562	843.669822783695\\
	1479.820959142	848.538987070691\\
	1469.10589238634	831.940691515519\\
	1455.03665211481	827.040860475658\\
	1416.922329557	794.88674619297\\
	1393.37449604174	830.63318395175\\
	1370.24085193011	845.636670997711\\
	1351.91395924476	842.45387945654\\
};
\addlegendentry{GP}
\end{axis}

\end{tikzpicture}

%% file: figures/gct_pred_avg_test.tikz
\begin{tikzpicture}

\definecolor{darkgray176}{RGB}{176,176,176}
\definecolor{green}{RGB}{0,128,0}
\definecolor{lightgray204}{RGB}{204,204,204}
\definecolor{purple}{RGB}{128,0,128}

\begin{axis}[
	width=\linewidth,
	height=0.8\linewidth,
	legend cell align={left},
	legend style={fill opacity=0.8, draw opacity=1, text opacity=1, draw=lightgray204},
	tick align=outside,
	tick pos=left,
	x grid style={darkgray176},
	xlabel={\footnotesize Time [s]},
	xmin=4, xmax=96,
	xtick style={color=black},
	y grid style={darkgray176},
	ylabel={\footnotesize RMSE [\si{\metre}]},
	ymin=7, ymax=33,
	ytick style={color=black},
	legend style={at={(0.97,0.97)},anchor=north east,legend cell align=left,draw=white!15!black,fill opacity=0.5,draw opacity=1,text opacity=1,font=\scriptsize, legend columns=2}
]
\path [fill=blue, fill opacity=0.1]
(axis cs:0,24.8263778686523)
--(axis cs:0,1.54150867462158)
--(axis cs:1,2.87966346740723)
--(axis cs:2,6.86192083358765)
--(axis cs:3,7.18405342102051)
--(axis cs:4,6.96474409103394)
--(axis cs:5,5.49107074737549)
--(axis cs:6,6.99497318267822)
--(axis cs:7,7.21903276443481)
--(axis cs:8,7.21485567092896)
--(axis cs:9,7.076340675354)
--(axis cs:10,7.90448665618896)
--(axis cs:11,7.49451351165771)
--(axis cs:12,6.36766338348389)
--(axis cs:13,5.79619407653809)
--(axis cs:14,6.71567249298096)
--(axis cs:15,5.59122943878174)
--(axis cs:16,4.79743671417236)
--(axis cs:17,5.41940689086914)
--(axis cs:18,4.95572185516357)
--(axis cs:19,5.98672914505005)
--(axis cs:20,5.05959987640381)
--(axis cs:21,3.7208890914917)
--(axis cs:22,3.74263286590576)
--(axis cs:23,6.80845069885254)
--(axis cs:24,5.22230434417725)
--(axis cs:25,7.91304969787598)
--(axis cs:26,7.34815549850464)
--(axis cs:27,7.82327556610107)
--(axis cs:28,9.19692707061768)
--(axis cs:29,8.19757080078125)
--(axis cs:30,7.14500665664673)
--(axis cs:31,5.97526454925537)
--(axis cs:32,8.92637538909912)
--(axis cs:33,7.43939113616943)
--(axis cs:34,5.87082290649414)
--(axis cs:35,7.28752660751343)
--(axis cs:36,5.74210119247437)
--(axis cs:37,5.98181581497192)
--(axis cs:38,7.25180244445801)
--(axis cs:39,6.85039377212524)
--(axis cs:40,7.14924573898315)
--(axis cs:41,5.25967597961426)
--(axis cs:42,5.7064733505249)
--(axis cs:43,6.35767650604248)
--(axis cs:44,7.38467884063721)
--(axis cs:45,7.31336069107056)
--(axis cs:46,7.78161811828613)
--(axis cs:47,8.20873737335205)
--(axis cs:48,6.08550548553467)
--(axis cs:49,5.49606275558472)
--(axis cs:50,5.66130542755127)
--(axis cs:51,5.65446138381958)
--(axis cs:52,7.53083419799805)
--(axis cs:53,6.85466814041138)
--(axis cs:54,8.33259868621826)
--(axis cs:55,6.40188455581665)
--(axis cs:56,7.20987224578857)
--(axis cs:57,8.88690757751465)
--(axis cs:58,8.75825595855713)
--(axis cs:59,7.59925651550293)
--(axis cs:60,8.25151920318604)
--(axis cs:61,8.41644668579102)
--(axis cs:62,6.5678186416626)
--(axis cs:63,8.05373001098633)
--(axis cs:64,8.23099803924561)
--(axis cs:65,6.84106826782227)
--(axis cs:66,5.55190944671631)
--(axis cs:67,7.00871181488037)
--(axis cs:68,5.2454137802124)
--(axis cs:69,6.9468092918396)
--(axis cs:70,7.10971736907959)
--(axis cs:71,7.35693359375)
--(axis cs:72,6.39690589904785)
--(axis cs:73,8.30521297454834)
--(axis cs:74,6.83198738098145)
--(axis cs:75,6.23256874084473)
--(axis cs:76,8.70876693725586)
--(axis cs:77,9.40045833587646)
--(axis cs:78,7.73372936248779)
--(axis cs:79,5.47037124633789)
--(axis cs:80,6.06292915344238)
--(axis cs:81,8.07365798950195)
--(axis cs:82,8.37150096893311)
--(axis cs:83,7.67401790618896)
--(axis cs:84,7.44198799133301)
--(axis cs:85,6.29481410980225)
--(axis cs:86,8.66633987426758)
--(axis cs:87,6.43224334716797)
--(axis cs:88,7.4678430557251)
--(axis cs:89,7.66540241241455)
--(axis cs:90,7.84202337265015)
--(axis cs:91,6.91495037078857)
--(axis cs:92,6.9057731628418)
--(axis cs:93,5.3596887588501)
--(axis cs:94,6.33808612823486)
--(axis cs:95,7.76704216003418)
--(axis cs:96,8.43212509155273)
--(axis cs:96,21.8254146575928)
--(axis cs:96,21.8254146575928)
--(axis cs:95,24.5322856903076)
--(axis cs:94,27.4630355834961)
--(axis cs:93,30.2244071960449)
--(axis cs:92,25.9829635620117)
--(axis cs:91,23.4519844055176)
--(axis cs:90,23.2978172302246)
--(axis cs:89,22.9552879333496)
--(axis cs:88,24.2194480895996)
--(axis cs:87,25.3314743041992)
--(axis cs:86,26.1816902160645)
--(axis cs:85,25.9646987915039)
--(axis cs:84,23.9423065185547)
--(axis cs:83,24.4853477478027)
--(axis cs:82,24.6358413696289)
--(axis cs:81,25.024471282959)
--(axis cs:80,25.2964344024658)
--(axis cs:79,29.4022979736328)
--(axis cs:78,26.6157417297363)
--(axis cs:77,27.0074081420898)
--(axis cs:76,24.306583404541)
--(axis cs:75,25.9285182952881)
--(axis cs:74,23.9222259521484)
--(axis cs:73,21.8493041992188)
--(axis cs:72,22.3987197875977)
--(axis cs:71,21.9280853271484)
--(axis cs:70,23.349422454834)
--(axis cs:69,21.3541011810303)
--(axis cs:68,21.6795349121094)
--(axis cs:67,20.2597351074219)
--(axis cs:66,22.392749786377)
--(axis cs:65,23.9019546508789)
--(axis cs:64,24.2012977600098)
--(axis cs:63,21.9472541809082)
--(axis cs:62,22.0747375488281)
--(axis cs:61,22.5926856994629)
--(axis cs:60,24.1209526062012)
--(axis cs:59,22.197114944458)
--(axis cs:58,23.8495483398438)
--(axis cs:57,23.4757518768311)
--(axis cs:56,24.2401428222656)
--(axis cs:55,21.7388076782227)
--(axis cs:54,23.4486846923828)
--(axis cs:53,20.2974033355713)
--(axis cs:52,20.724178314209)
--(axis cs:51,21.3319225311279)
--(axis cs:50,19.0572090148926)
--(axis cs:49,21.1419544219971)
--(axis cs:48,22.27001953125)
--(axis cs:47,23.6929550170898)
--(axis cs:46,24.2748279571533)
--(axis cs:45,20.6823787689209)
--(axis cs:44,24.041015625)
--(axis cs:43,24.5686569213867)
--(axis cs:42,25.1852836608887)
--(axis cs:41,24.5715770721436)
--(axis cs:40,22.3516006469727)
--(axis cs:39,21.5387134552002)
--(axis cs:38,20.4101963043213)
--(axis cs:37,21.3134536743164)
--(axis cs:36,18.8187656402588)
--(axis cs:35,20.6807098388672)
--(axis cs:34,21.2538948059082)
--(axis cs:33,21.0122985839844)
--(axis cs:32,20.122974395752)
--(axis cs:31,20.5563163757324)
--(axis cs:30,20.4907855987549)
--(axis cs:29,23.3574066162109)
--(axis cs:28,23.9831504821777)
--(axis cs:27,21.9284439086914)
--(axis cs:26,22.9224033355713)
--(axis cs:25,24.2601299285889)
--(axis cs:24,24.4206504821777)
--(axis cs:23,22.8560771942139)
--(axis cs:22,23.0661697387695)
--(axis cs:21,22.1530609130859)
--(axis cs:20,20.6739158630371)
--(axis cs:19,21.9065093994141)
--(axis cs:18,22.828800201416)
--(axis cs:17,22.4183330535889)
--(axis cs:16,22.209846496582)
--(axis cs:15,23.6555633544922)
--(axis cs:14,22.5038375854492)
--(axis cs:13,24.565860748291)
--(axis cs:12,22.9303817749023)
--(axis cs:11,21.557689666748)
--(axis cs:10,24.5347595214844)
--(axis cs:9,24.388484954834)
--(axis cs:8,22.4206314086914)
--(axis cs:7,22.3364219665527)
--(axis cs:6,19.3731231689453)
--(axis cs:5,19.6359443664551)
--(axis cs:4,20.84739112854)
--(axis cs:3,21.8002433776855)
--(axis cs:2,21.7335014343262)
--(axis cs:1,20.9194526672363)
--(axis cs:0,24.8263778686523)
--cycle;

\path [fill=green, fill opacity=0.1]
(axis cs:0,24.8896293640137)
--(axis cs:0,1.69249153137207)
--(axis cs:1,3.17431926727295)
--(axis cs:2,6.41425704956055)
--(axis cs:3,8.00248336791992)
--(axis cs:4,8.64943981170654)
--(axis cs:5,7.71674346923828)
--(axis cs:6,7.93945217132568)
--(axis cs:7,6.92886543273926)
--(axis cs:8,6.58117914199829)
--(axis cs:9,4.91750240325928)
--(axis cs:10,4.70568370819092)
--(axis cs:11,5.33300590515137)
--(axis cs:12,5.87457132339478)
--(axis cs:13,5.53173160552979)
--(axis cs:14,5.82688283920288)
--(axis cs:15,7.71999740600586)
--(axis cs:16,7.03338623046875)
--(axis cs:17,6.86618995666504)
--(axis cs:18,7.33235311508179)
--(axis cs:19,8.0637378692627)
--(axis cs:20,7.03243637084961)
--(axis cs:21,7.78982496261597)
--(axis cs:22,7.03995513916016)
--(axis cs:23,8.34623146057129)
--(axis cs:24,8.32987117767334)
--(axis cs:25,7.53711843490601)
--(axis cs:26,7.16815233230591)
--(axis cs:27,8.18232917785645)
--(axis cs:28,8.82693672180176)
--(axis cs:29,7.67749500274658)
--(axis cs:30,7.15529298782349)
--(axis cs:31,6.41695976257324)
--(axis cs:32,6.48359775543213)
--(axis cs:33,5.31591367721558)
--(axis cs:34,6.07204532623291)
--(axis cs:35,6.49577760696411)
--(axis cs:36,6.31224298477173)
--(axis cs:37,5.9906587600708)
--(axis cs:38,7.06605768203735)
--(axis cs:39,6.65041351318359)
--(axis cs:40,7.19128274917603)
--(axis cs:41,6.70650863647461)
--(axis cs:42,7.83679580688477)
--(axis cs:43,8.68842124938965)
--(axis cs:44,9.03618049621582)
--(axis cs:45,7.84906816482544)
--(axis cs:46,7.37735366821289)
--(axis cs:47,7.76407480239868)
--(axis cs:48,7.70157432556152)
--(axis cs:49,6.75446224212646)
--(axis cs:50,5.41815423965454)
--(axis cs:51,5.33692502975464)
--(axis cs:52,6.49486637115479)
--(axis cs:53,7.78209781646729)
--(axis cs:54,8.29882621765137)
--(axis cs:55,7.79362010955811)
--(axis cs:56,6.38142013549805)
--(axis cs:57,6.77075147628784)
--(axis cs:58,7.40492963790894)
--(axis cs:59,6.49972343444824)
--(axis cs:60,6.12051296234131)
--(axis cs:61,7.35258865356445)
--(axis cs:62,7.56563091278076)
--(axis cs:63,8.21882438659668)
--(axis cs:64,7.99412202835083)
--(axis cs:65,6.91143465042114)
--(axis cs:66,6.04421234130859)
--(axis cs:67,5.50186443328857)
--(axis cs:68,4.48895740509033)
--(axis cs:69,4.60325527191162)
--(axis cs:70,5.96212959289551)
--(axis cs:71,6.41599750518799)
--(axis cs:72,6.42307662963867)
--(axis cs:73,6.6512393951416)
--(axis cs:74,6.78277492523193)
--(axis cs:75,7.40376901626587)
--(axis cs:76,6.96563625335693)
--(axis cs:77,8.73249816894531)
--(axis cs:78,8.71712493896484)
--(axis cs:79,7.49790620803833)
--(axis cs:80,7.01327610015869)
--(axis cs:81,6.5898904800415)
--(axis cs:82,8.04177665710449)
--(axis cs:83,6.74784898757935)
--(axis cs:84,6.12733602523804)
--(axis cs:85,5.81457996368408)
--(axis cs:86,7.85832500457764)
--(axis cs:87,8.41287803649902)
--(axis cs:88,8.52383995056152)
--(axis cs:89,8.72008323669434)
--(axis cs:90,8.11962985992432)
--(axis cs:91,7.33227586746216)
--(axis cs:92,8.19137287139893)
--(axis cs:93,9.3114595413208)
--(axis cs:94,8.2537727355957)
--(axis cs:95,7.68732833862305)
--(axis cs:96,6.34932327270508)
--(axis cs:96,21.113094329834)
--(axis cs:96,21.113094329834)
--(axis cs:95,20.3382987976074)
--(axis cs:94,19.973424911499)
--(axis cs:93,20.2737655639648)
--(axis cs:92,20.0622444152832)
--(axis cs:91,20.1740036010742)
--(axis cs:90,20.4731864929199)
--(axis cs:89,20.7286281585693)
--(axis cs:88,22.1100006103516)
--(axis cs:87,20.6619262695312)
--(axis cs:86,20.1959037780762)
--(axis cs:85,19.8409957885742)
--(axis cs:84,20.3881855010986)
--(axis cs:83,20.0167560577393)
--(axis cs:82,19.2624683380127)
--(axis cs:81,19.805793762207)
--(axis cs:80,20.6624984741211)
--(axis cs:79,21.0109062194824)
--(axis cs:78,20.7184562683105)
--(axis cs:77,21.7687892913818)
--(axis cs:76,21.3002319335938)
--(axis cs:75,20.6086292266846)
--(axis cs:74,19.9405632019043)
--(axis cs:73,18.8985595703125)
--(axis cs:72,19.8891258239746)
--(axis cs:71,20.7906761169434)
--(axis cs:70,21.7119922637939)
--(axis cs:69,21.6895027160645)
--(axis cs:68,21.6845855712891)
--(axis cs:67,21.2237777709961)
--(axis cs:66,22.105863571167)
--(axis cs:65,22.621654510498)
--(axis cs:64,21.3195819854736)
--(axis cs:63,18.7263813018799)
--(axis cs:62,18.5604248046875)
--(axis cs:61,19.4594459533691)
--(axis cs:60,20.6110725402832)
--(axis cs:59,19.5813579559326)
--(axis cs:58,19.3445377349854)
--(axis cs:57,19.4929008483887)
--(axis cs:56,19.8939151763916)
--(axis cs:55,19.7854309082031)
--(axis cs:54,20.0242080688477)
--(axis cs:53,17.5842781066895)
--(axis cs:52,19.5659103393555)
--(axis cs:51,21.1019840240479)
--(axis cs:50,20.5297317504883)
--(axis cs:49,21.4945335388184)
--(axis cs:48,21.1351585388184)
--(axis cs:47,20.7011680603027)
--(axis cs:46,21.3668251037598)
--(axis cs:45,20.9396533966064)
--(axis cs:44,22.3011531829834)
--(axis cs:43,22.5567398071289)
--(axis cs:42,21.34397315979)
--(axis cs:41,21.3632488250732)
--(axis cs:40,20.4356861114502)
--(axis cs:39,21.0167388916016)
--(axis cs:38,20.3594970703125)
--(axis cs:37,19.6677551269531)
--(axis cs:36,17.8840732574463)
--(axis cs:35,19.2135066986084)
--(axis cs:34,19.0848083496094)
--(axis cs:33,19.1491851806641)
--(axis cs:32,19.8339462280273)
--(axis cs:31,19.1543464660645)
--(axis cs:30,18.5238132476807)
--(axis cs:29,19.0776405334473)
--(axis cs:28,19.1177272796631)
--(axis cs:27,19.2504482269287)
--(axis cs:26,20.5056781768799)
--(axis cs:25,21.564136505127)
--(axis cs:24,20.7358169555664)
--(axis cs:23,21.1403179168701)
--(axis cs:22,21.1218776702881)
--(axis cs:21,20.5529747009277)
--(axis cs:20,20.7230682373047)
--(axis cs:19,22.298957824707)
--(axis cs:18,23.0333576202393)
--(axis cs:17,21.7767162322998)
--(axis cs:16,20.3406219482422)
--(axis cs:15,20.2889671325684)
--(axis cs:14,19.056999206543)
--(axis cs:13,19.3090667724609)
--(axis cs:12,19.6548042297363)
--(axis cs:11,20.016393661499)
--(axis cs:10,22.6198997497559)
--(axis cs:9,21.1999549865723)
--(axis cs:8,20.8282909393311)
--(axis cs:7,21.1420402526855)
--(axis cs:6,21.7116203308105)
--(axis cs:5,23.1182765960693)
--(axis cs:4,24.4544258117676)
--(axis cs:3,23.2854423522949)
--(axis cs:2,21.2551250457764)
--(axis cs:1,20.5686874389648)
--(axis cs:0,24.8896293640137)
--cycle;

\path [fill=purple, fill opacity=0.1]
(axis cs:0,2.38749718666077)
--(axis cs:0,0.733117043972015)
--(axis cs:1,1.02621614933014)
--(axis cs:2,1.35463380813599)
--(axis cs:3,5.6408519744873)
--(axis cs:4,5.19582843780518)
--(axis cs:5,6.76295757293701)
--(axis cs:6,7.61734485626221)
--(axis cs:7,7.99481391906738)
--(axis cs:8,7.78724670410156)
--(axis cs:9,9.22162342071533)
--(axis cs:10,11.9968023300171)
--(axis cs:11,10.0593957901001)
--(axis cs:12,5.65886116027832)
--(axis cs:13,5.86268901824951)
--(axis cs:14,9.00158977508545)
--(axis cs:15,5.84768295288086)
--(axis cs:16,6.9818754196167)
--(axis cs:17,7.16990947723389)
--(axis cs:18,7.41440773010254)
--(axis cs:19,5.43074989318848)
--(axis cs:20,6.53870582580566)
--(axis cs:21,6.62424468994141)
--(axis cs:22,7.58370971679688)
--(axis cs:23,8.49090957641602)
--(axis cs:24,7.83946228027344)
--(axis cs:25,9.01579856872559)
--(axis cs:26,7.5393123626709)
--(axis cs:27,6.8069372177124)
--(axis cs:28,7.07397079467773)
--(axis cs:29,6.63718700408936)
--(axis cs:30,7.551513671875)
--(axis cs:31,8.22105121612549)
--(axis cs:32,8.14347457885742)
--(axis cs:33,6.90937232971191)
--(axis cs:34,6.01234340667725)
--(axis cs:35,7.18950748443604)
--(axis cs:36,7.74250602722168)
--(axis cs:37,1.6679573059082)
--(axis cs:38,7.6245813369751)
--(axis cs:39,8.11915874481201)
--(axis cs:40,10.5725975036621)
--(axis cs:41,9.76601028442383)
--(axis cs:42,9.59442234039307)
--(axis cs:43,6.76971530914307)
--(axis cs:44,4.79375743865967)
--(axis cs:45,5.39518260955811)
--(axis cs:46,5.34773921966553)
--(axis cs:47,5.93732452392578)
--(axis cs:48,6.52038192749023)
--(axis cs:49,8.04445648193359)
--(axis cs:50,8.5263671875)
--(axis cs:51,6.59734058380127)
--(axis cs:52,8.20012760162354)
--(axis cs:53,9.02327919006348)
--(axis cs:54,7.07889461517334)
--(axis cs:55,8.88260173797607)
--(axis cs:56,9.01115608215332)
--(axis cs:57,7.91311073303223)
--(axis cs:58,8.35500717163086)
--(axis cs:59,10.2463617324829)
--(axis cs:60,9.16429996490479)
--(axis cs:61,7.35320854187012)
--(axis cs:62,10.2118043899536)
--(axis cs:63,12.0499086380005)
--(axis cs:64,6.26028251647949)
--(axis cs:65,9.55041885375977)
--(axis cs:66,9.88023853302002)
--(axis cs:67,9.41187858581543)
--(axis cs:68,10.5269603729248)
--(axis cs:69,9.18580150604248)
--(axis cs:70,8.88197612762451)
--(axis cs:71,8.14023113250732)
--(axis cs:72,8.99397373199463)
--(axis cs:73,12.8743715286255)
--(axis cs:74,6.99787712097168)
--(axis cs:75,7.52673530578613)
--(axis cs:76,7.81089401245117)
--(axis cs:77,7.44574546813965)
--(axis cs:78,6.40349769592285)
--(axis cs:79,8.57744979858398)
--(axis cs:80,10.1004276275635)
--(axis cs:81,8.05494499206543)
--(axis cs:82,10.4776859283447)
--(axis cs:83,8.49537754058838)
--(axis cs:84,10.6466417312622)
--(axis cs:85,11.6500930786133)
--(axis cs:86,9.74882793426514)
--(axis cs:87,10.5025444030762)
--(axis cs:88,9.64763450622559)
--(axis cs:89,7.28092384338379)
--(axis cs:90,9.7667818069458)
--(axis cs:91,8.16392993927002)
--(axis cs:92,10.8243436813354)
--(axis cs:93,9.13239765167236)
--(axis cs:94,8.62982845306396)
--(axis cs:95,9.48371601104736)
--(axis cs:96,9.90354633331299)
--(axis cs:96,32.1866722106934)
--(axis cs:96,32.1866722106934)
--(axis cs:95,30.7697067260742)
--(axis cs:94,34.043514251709)
--(axis cs:93,30.3890686035156)
--(axis cs:92,32.122859954834)
--(axis cs:91,34.2376251220703)
--(axis cs:90,30.5842170715332)
--(axis cs:89,28.7322635650635)
--(axis cs:88,27.5578556060791)
--(axis cs:87,30.4957962036133)
--(axis cs:86,38.029125213623)
--(axis cs:85,34.7510299682617)
--(axis cs:84,34.2353477478027)
--(axis cs:83,35.9384841918945)
--(axis cs:82,34.8633499145508)
--(axis cs:81,40.4463043212891)
--(axis cs:80,36.7877578735352)
--(axis cs:79,36.9626655578613)
--(axis cs:78,36.3304214477539)
--(axis cs:77,36.1699600219727)
--(axis cs:76,34.7767944335938)
--(axis cs:75,32.1082000732422)
--(axis cs:74,35.3924713134766)
--(axis cs:73,29.114803314209)
--(axis cs:72,31.4748153686523)
--(axis cs:71,34.7641448974609)
--(axis cs:70,29.2462730407715)
--(axis cs:69,31.2107543945312)
--(axis cs:68,31.7223987579346)
--(axis cs:67,33.797492980957)
--(axis cs:66,32.6688461303711)
--(axis cs:65,31.8739852905273)
--(axis cs:64,35.6858749389648)
--(axis cs:63,32.3297004699707)
--(axis cs:62,31.566967010498)
--(axis cs:61,33.6926345825195)
--(axis cs:60,34.647029876709)
--(axis cs:59,33.0335311889648)
--(axis cs:58,31.5070877075195)
--(axis cs:57,29.9065647125244)
--(axis cs:56,33.6807708740234)
--(axis cs:55,32.6425552368164)
--(axis cs:54,32.7926712036133)
--(axis cs:53,30.0521755218506)
--(axis cs:52,32.6848068237305)
--(axis cs:51,35.1454124450684)
--(axis cs:50,35.2711219787598)
--(axis cs:49,31.4214553833008)
--(axis cs:48,39.5376663208008)
--(axis cs:47,40.216911315918)
--(axis cs:46,32.8021240234375)
--(axis cs:45,32.6556930541992)
--(axis cs:44,31.4318809509277)
--(axis cs:43,28.6763038635254)
--(axis cs:42,29.7624778747559)
--(axis cs:41,29.232780456543)
--(axis cs:40,28.9651374816895)
--(axis cs:39,29.3567733764648)
--(axis cs:38,28.0931854248047)
--(axis cs:37,39.7268753051758)
--(axis cs:36,30.3243923187256)
--(axis cs:35,31.911304473877)
--(axis cs:34,25.7876930236816)
--(axis cs:33,25.7129039764404)
--(axis cs:32,31.7712211608887)
--(axis cs:31,30.5420837402344)
--(axis cs:30,31.3675880432129)
--(axis cs:29,34.8095474243164)
--(axis cs:28,30.5986976623535)
--(axis cs:27,31.2539443969727)
--(axis cs:26,29.4115314483643)
--(axis cs:25,28.9928150177002)
--(axis cs:24,28.5245056152344)
--(axis cs:23,32.6046752929688)
--(axis cs:22,29.895622253418)
--(axis cs:21,33.521125793457)
--(axis cs:20,31.7322673797607)
--(axis cs:19,29.1242275238037)
--(axis cs:18,30.7400455474854)
--(axis cs:17,33.5906066894531)
--(axis cs:16,32.2921676635742)
--(axis cs:15,30.084529876709)
--(axis cs:14,30.4722671508789)
--(axis cs:13,34.0954170227051)
--(axis cs:12,32.9468078613281)
--(axis cs:11,32.6246032714844)
--(axis cs:10,26.5142860412598)
--(axis cs:9,28.5015106201172)
--(axis cs:8,29.8572845458984)
--(axis cs:7,28.8068943023682)
--(axis cs:6,24.6208343505859)
--(axis cs:5,24.1141929626465)
--(axis cs:4,23.2671318054199)
--(axis cs:3,16.2584419250488)
--(axis cs:2,5.869215965271)
--(axis cs:1,4.45814323425293)
--(axis cs:0,2.38749718666077)
--cycle;

\path [fill=red, fill opacity=0.1]
(axis cs:0,24.8263702392578)
--(axis cs:0,1.54151630401611)
--(axis cs:1,2.9256706237793)
--(axis cs:2,6.83733367919922)
--(axis cs:3,5.74896240234375)
--(axis cs:4,7.46495962142944)
--(axis cs:5,7.13270711898804)
--(axis cs:6,7.05327796936035)
--(axis cs:7,7.06783103942871)
--(axis cs:8,8.10237503051758)
--(axis cs:9,6.40572071075439)
--(axis cs:10,8.3616361618042)
--(axis cs:11,7.58741426467896)
--(axis cs:12,7.69613742828369)
--(axis cs:13,5.67314338684082)
--(axis cs:14,6.84592533111572)
--(axis cs:15,5.73347282409668)
--(axis cs:16,5.1025505065918)
--(axis cs:17,5.82545948028564)
--(axis cs:18,4.55907821655273)
--(axis cs:19,6.23553085327148)
--(axis cs:20,5.05865287780762)
--(axis cs:21,3.36568069458008)
--(axis cs:22,5.08084106445312)
--(axis cs:23,8.0064582824707)
--(axis cs:24,6.01008033752441)
--(axis cs:25,9.34685707092285)
--(axis cs:26,6.38553237915039)
--(axis cs:27,6.77414321899414)
--(axis cs:28,8.387038230896)
--(axis cs:29,7.49566841125488)
--(axis cs:30,7.75841045379639)
--(axis cs:31,6.88885450363159)
--(axis cs:32,8.81764793395996)
--(axis cs:33,7.25757884979248)
--(axis cs:34,5.76841259002686)
--(axis cs:35,7.56208419799805)
--(axis cs:36,5.7597131729126)
--(axis cs:37,5.87843418121338)
--(axis cs:38,6.33715343475342)
--(axis cs:39,6.66381454467773)
--(axis cs:40,8.05002975463867)
--(axis cs:41,6.32151412963867)
--(axis cs:42,6.74908542633057)
--(axis cs:43,7.34720993041992)
--(axis cs:44,8.83250331878662)
--(axis cs:45,7.76843070983887)
--(axis cs:46,7.66365337371826)
--(axis cs:47,7.66905212402344)
--(axis cs:48,5.95624160766602)
--(axis cs:49,4.58195304870605)
--(axis cs:50,5.83402824401855)
--(axis cs:51,6.00506019592285)
--(axis cs:52,7.92079544067383)
--(axis cs:53,7.56756114959717)
--(axis cs:54,9.39053153991699)
--(axis cs:55,6.87938737869263)
--(axis cs:56,8.22433090209961)
--(axis cs:57,8.99559593200684)
--(axis cs:58,8.38162803649902)
--(axis cs:59,8.18204307556152)
--(axis cs:60,8.42301940917969)
--(axis cs:61,8.42700576782227)
--(axis cs:62,6.08104515075684)
--(axis cs:63,8.79242324829102)
--(axis cs:64,10.0999975204468)
--(axis cs:65,7.41543674468994)
--(axis cs:66,5.26625919342041)
--(axis cs:67,6.56230211257935)
--(axis cs:68,6.16181755065918)
--(axis cs:69,8.16462802886963)
--(axis cs:70,7.67440223693848)
--(axis cs:71,7.52901697158813)
--(axis cs:72,5.72290420532227)
--(axis cs:73,8.28251647949219)
--(axis cs:74,7.61944103240967)
--(axis cs:75,7.03934574127197)
--(axis cs:76,9.05318260192871)
--(axis cs:77,9.43513488769531)
--(axis cs:78,8.22940158843994)
--(axis cs:79,5.21457386016846)
--(axis cs:80,7.23278999328613)
--(axis cs:81,8.56697082519531)
--(axis cs:82,8.18489646911621)
--(axis cs:83,8.15569114685059)
--(axis cs:84,7.50386428833008)
--(axis cs:85,7.45695209503174)
--(axis cs:86,8.86858367919922)
--(axis cs:87,5.65998554229736)
--(axis cs:88,7.71631717681885)
--(axis cs:89,7.42528295516968)
--(axis cs:90,7.34743881225586)
--(axis cs:91,7.58266544342041)
--(axis cs:92,7.26715087890625)
--(axis cs:93,5.44184494018555)
--(axis cs:94,6.20370769500732)
--(axis cs:95,7.50959777832031)
--(axis cs:96,8.12942504882812)
--(axis cs:96,22.7319221496582)
--(axis cs:96,22.7319221496582)
--(axis cs:95,24.6631698608398)
--(axis cs:94,27.3382110595703)
--(axis cs:93,30.9827194213867)
--(axis cs:92,26.5342788696289)
--(axis cs:91,24.0091323852539)
--(axis cs:90,23.9225425720215)
--(axis cs:89,23.1467361450195)
--(axis cs:88,25.3492012023926)
--(axis cs:87,26.5365333557129)
--(axis cs:86,27.4260063171387)
--(axis cs:85,27.2630500793457)
--(axis cs:84,25.1346549987793)
--(axis cs:83,24.8513050079346)
--(axis cs:82,24.9694499969482)
--(axis cs:81,25.1698684692383)
--(axis cs:80,25.5448131561279)
--(axis cs:79,29.8127403259277)
--(axis cs:78,27.626033782959)
--(axis cs:77,28.4018630981445)
--(axis cs:76,24.5901470184326)
--(axis cs:75,27.0352020263672)
--(axis cs:74,25.3828048706055)
--(axis cs:73,23.5376224517822)
--(axis cs:72,23.2919044494629)
--(axis cs:71,22.8528480529785)
--(axis cs:70,25.2360210418701)
--(axis cs:69,22.4236831665039)
--(axis cs:68,23.0407810211182)
--(axis cs:67,21.6816711425781)
--(axis cs:66,22.6053047180176)
--(axis cs:65,25.6017990112305)
--(axis cs:64,26.1393089294434)
--(axis cs:63,23.3873672485352)
--(axis cs:62,22.5894241333008)
--(axis cs:61,24.0719947814941)
--(axis cs:60,25.5044784545898)
--(axis cs:59,22.5228748321533)
--(axis cs:58,24.7741718292236)
--(axis cs:57,24.2164287567139)
--(axis cs:56,25.3867645263672)
--(axis cs:55,22.5552825927734)
--(axis cs:54,25.5416049957275)
--(axis cs:53,21.6412887573242)
--(axis cs:52,22.3437328338623)
--(axis cs:51,22.4229297637939)
--(axis cs:50,19.6466846466064)
--(axis cs:49,22.5848808288574)
--(axis cs:48,23.2304153442383)
--(axis cs:47,25.0398864746094)
--(axis cs:46,25.9498252868652)
--(axis cs:45,20.7767448425293)
--(axis cs:44,24.3979911804199)
--(axis cs:43,25.286735534668)
--(axis cs:42,26.374683380127)
--(axis cs:41,26.5608520507812)
--(axis cs:40,24.6456413269043)
--(axis cs:39,23.8916778564453)
--(axis cs:38,22.6483421325684)
--(axis cs:37,23.8518295288086)
--(axis cs:36,19.8069305419922)
--(axis cs:35,21.3512268066406)
--(axis cs:34,21.872802734375)
--(axis cs:33,22.2191352844238)
--(axis cs:32,21.0946941375732)
--(axis cs:31,21.5492134094238)
--(axis cs:30,21.2369232177734)
--(axis cs:29,25.9433422088623)
--(axis cs:28,24.7235488891602)
--(axis cs:27,22.691104888916)
--(axis cs:26,24.0471134185791)
--(axis cs:25,25.0642147064209)
--(axis cs:24,26.357816696167)
--(axis cs:23,25.0509300231934)
--(axis cs:22,25.0566806793213)
--(axis cs:21,24.5503940582275)
--(axis cs:20,22.4124164581299)
--(axis cs:19,23.1471862792969)
--(axis cs:18,23.9904727935791)
--(axis cs:17,23.6748352050781)
--(axis cs:16,23.2954540252686)
--(axis cs:15,25.4457263946533)
--(axis cs:14,24.4016418457031)
--(axis cs:13,26.7523441314697)
--(axis cs:12,23.5908317565918)
--(axis cs:11,21.8618831634521)
--(axis cs:10,25.5003967285156)
--(axis cs:9,25.8379440307617)
--(axis cs:8,24.2726135253906)
--(axis cs:7,25.2313289642334)
--(axis cs:6,22.7102203369141)
--(axis cs:5,22.2068138122559)
--(axis cs:4,20.8540420532227)
--(axis cs:3,19.6031055450439)
--(axis cs:2,20.4713363647461)
--(axis cs:1,20.5740909576416)
--(axis cs:0,24.8263702392578)
--cycle;

\addplot [semithick, blue, opacity=0.9]
table {%
0 13.1839427947998
1 11.8995580673218
2 14.2977113723755
3 14.492148399353
4 13.9060678482056
5 12.5635080337524
6 13.1840476989746
7 14.7777271270752
8 14.8177433013916
9 15.7324132919312
10 16.2196235656738
11 14.5261011123657
12 14.6490230560303
13 15.1810274124146
14 14.6097555160522
15 14.6233968734741
16 13.50364112854
17 13.918869972229
18 13.8922605514526
19 13.9466190338135
20 12.8667573928833
21 12.936975479126
22 13.4044017791748
23 14.8322639465332
24 14.8214778900146
25 16.0865898132324
26 15.1352796554565
27 14.8758592605591
28 16.5900382995605
29 15.7774887084961
30 13.8178958892822
31 13.2657899856567
32 14.5246753692627
33 14.2258443832397
34 13.5623588562012
35 13.9841184616089
36 12.2804336547852
37 13.6476345062256
38 13.8309993743896
39 14.1945533752441
40 14.7504234313965
41 14.9156265258789
42 15.4458780288696
43 15.4631662368774
44 15.7128477096558
45 13.9978694915771
46 16.0282230377197
47 15.9508457183838
48 14.1777620315552
49 13.3190088272095
50 12.3592567443848
51 13.4931917190552
52 14.1275062561035
53 13.5760354995728
54 15.8906421661377
55 14.0703458786011
56 15.7250070571899
57 16.1813297271729
58 16.3039016723633
59 14.8981857299805
60 16.1862354278564
61 15.504566192627
62 14.3212776184082
63 15.0004920959473
64 16.2161483764648
65 15.3715114593506
66 13.9723291397095
67 13.6342239379883
68 13.462474822998
69 14.1504554748535
70 15.2295694351196
71 14.6425094604492
72 14.3978128433228
73 15.0772590637207
74 15.3771066665649
75 16.0805435180664
76 16.5076751708984
77 18.2039337158203
78 17.1747360229492
79 17.4363346099854
80 15.6796817779541
81 16.5490646362305
82 16.5036716461182
83 16.079683303833
84 15.6921472549438
85 16.1297569274902
86 17.424015045166
87 15.8818588256836
88 15.8436450958252
89 15.3103456497192
90 15.569920539856
91 15.1834669113159
92 16.4443683624268
93 17.7920475006104
94 16.9005603790283
95 16.1496639251709
96 15.1287698745728
};
\addlegendentry{Opt. IMM}
\addplot [semithick, blue, opacity=0.9, dashed, forget plot]
table {%
0 14.9194183349609
1 14.9194183349609
2 14.9194183349609
3 14.9194183349609
4 14.9194183349609
5 14.9194183349609
6 14.9194183349609
7 14.9194183349609
8 14.9194183349609
9 14.9194183349609
10 14.9194183349609
11 14.9194183349609
12 14.9194183349609
13 14.9194183349609
14 14.9194183349609
15 14.9194183349609
16 14.9194183349609
17 14.9194183349609
18 14.9194183349609
19 14.9194183349609
20 14.9194183349609
21 14.9194183349609
22 14.9194183349609
23 14.9194183349609
24 14.9194183349609
25 14.9194183349609
26 14.9194183349609
27 14.9194183349609
28 14.9194183349609
29 14.9194183349609
30 14.9194183349609
31 14.9194183349609
32 14.9194183349609
33 14.9194183349609
34 14.9194183349609
35 14.9194183349609
36 14.9194183349609
37 14.9194183349609
38 14.9194183349609
39 14.9194183349609
40 14.9194183349609
41 14.9194183349609
42 14.9194183349609
43 14.9194183349609
44 14.9194183349609
45 14.9194183349609
46 14.9194183349609
47 14.9194183349609
48 14.9194183349609
49 14.9194183349609
50 14.9194183349609
51 14.9194183349609
52 14.9194183349609
53 14.9194183349609
54 14.9194183349609
55 14.9194183349609
56 14.9194183349609
57 14.9194183349609
58 14.9194183349609
59 14.9194183349609
60 14.9194183349609
61 14.9194183349609
62 14.9194183349609
63 14.9194183349609
64 14.9194183349609
65 14.9194183349609
66 14.9194183349609
67 14.9194183349609
68 14.9194183349609
69 14.9194183349609
70 14.9194183349609
71 14.9194183349609
72 14.9194183349609
73 14.9194183349609
74 14.9194183349609
75 14.9194183349609
76 14.9194183349609
77 14.9194183349609
78 14.9194183349609
79 14.9194183349609
80 14.9194183349609
81 14.9194183349609
82 14.9194183349609
83 14.9194183349609
84 14.9194183349609
85 14.9194183349609
86 14.9194183349609
87 14.9194183349609
88 14.9194183349609
89 14.9194183349609
90 14.9194183349609
91 14.9194183349609
92 14.9194183349609
93 14.9194183349609
94 14.9194183349609
95 14.9194183349609
96 14.9194183349609
};
\addplot [semithick, green, opacity=0.9]
table {%
0 13.2910604476929
1 11.8715028762817
2 13.8346910476685
3 15.6439628601074
4 16.5519332885742
5 15.4175100326538
6 14.8255367279053
7 14.0354528427124
8 13.7047348022461
9 13.0587282180786
10 13.6627922058105
11 12.6746997833252
12 12.764687538147
13 12.4203987121582
14 12.4419412612915
15 14.0044822692871
16 13.6870040893555
17 14.3214530944824
18 15.1828556060791
19 15.1813478469849
20 13.8777523040771
21 14.1714000701904
22 14.0809164047241
23 14.7432746887207
24 14.532844543457
25 14.5506277084351
26 13.8369150161743
27 13.7163887023926
28 13.9723320007324
29 13.3775682449341
30 12.8395528793335
31 12.7856531143188
32 13.1587715148926
33 12.2325496673584
34 12.578426361084
35 12.8546419143677
36 12.0981578826904
37 12.8292064666748
38 13.7127771377563
39 13.8335762023926
40 13.8134841918945
41 14.0348787307739
42 14.5903844833374
43 15.6225805282593
44 15.6686668395996
45 14.3943605422974
46 14.3720893859863
47 14.2326211929321
48 14.4183664321899
49 14.1244974136353
50 12.9739427566528
51 13.2194547653198
52 13.030387878418
53 12.6831874847412
54 14.1615171432495
55 13.7895250320435
56 13.1376676559448
57 13.1318264007568
58 13.3747339248657
59 13.0405406951904
60 13.3657922744751
61 13.4060173034668
62 13.0630283355713
63 13.4726028442383
64 14.6568517684937
65 14.766544342041
66 14.0750379562378
67 13.3628206253052
68 13.0867710113525
69 13.1463794708252
70 13.8370609283447
71 13.6033363342285
72 13.1561012268066
73 12.7748994827271
74 13.361668586731
75 14.0061988830566
76 14.1329336166382
77 15.2506437301636
78 14.7177906036377
79 14.2544059753418
80 13.8378868103027
81 13.1978416442871
82 13.6521224975586
83 13.3823022842407
84 13.2577610015869
85 12.8277883529663
86 14.0271139144897
87 14.5374021530151
88 15.3169202804565
89 14.7243556976318
90 14.2964086532593
91 13.7531394958496
92 14.1268091201782
93 14.7926120758057
94 14.1135988235474
95 14.0128135681152
96 13.7312088012695
};
\addlegendentry{MKF}
\addplot [semithick, green, opacity=0.9, dashed, forget plot]
table {%
0 13.8068017959595
1 13.8068017959595
2 13.8068017959595
3 13.8068017959595
4 13.8068017959595
5 13.8068017959595
6 13.8068017959595
7 13.8068017959595
8 13.8068017959595
9 13.8068017959595
10 13.8068017959595
11 13.8068017959595
12 13.8068017959595
13 13.8068017959595
14 13.8068017959595
15 13.8068017959595
16 13.8068017959595
17 13.8068017959595
18 13.8068017959595
19 13.8068017959595
20 13.8068017959595
21 13.8068017959595
22 13.8068017959595
23 13.8068017959595
24 13.8068017959595
25 13.8068017959595
26 13.8068017959595
27 13.8068017959595
28 13.8068017959595
29 13.8068017959595
30 13.8068017959595
31 13.8068017959595
32 13.8068017959595
33 13.8068017959595
34 13.8068017959595
35 13.8068017959595
36 13.8068017959595
37 13.8068017959595
38 13.8068017959595
39 13.8068017959595
40 13.8068017959595
41 13.8068017959595
42 13.8068017959595
43 13.8068017959595
44 13.8068017959595
45 13.8068017959595
46 13.8068017959595
47 13.8068017959595
48 13.8068017959595
49 13.8068017959595
50 13.8068017959595
51 13.8068017959595
52 13.8068017959595
53 13.8068017959595
54 13.8068017959595
55 13.8068017959595
56 13.8068017959595
57 13.8068017959595
58 13.8068017959595
59 13.8068017959595
60 13.8068017959595
61 13.8068017959595
62 13.8068017959595
63 13.8068017959595
64 13.8068017959595
65 13.8068017959595
66 13.8068017959595
67 13.8068017959595
68 13.8068017959595
69 13.8068017959595
70 13.8068017959595
71 13.8068017959595
72 13.8068017959595
73 13.8068017959595
74 13.8068017959595
75 13.8068017959595
76 13.8068017959595
77 13.8068017959595
78 13.8068017959595
79 13.8068017959595
80 13.8068017959595
81 13.8068017959595
82 13.8068017959595
83 13.8068017959595
84 13.8068017959595
85 13.8068017959595
86 13.8068017959595
87 13.8068017959595
88 13.8068017959595
89 13.8068017959595
90 13.8068017959595
91 13.8068017959595
92 13.8068017959595
93 13.8068017959595
94 13.8068017959595
95 13.8068017959595
96 13.8068017959595
};
\addplot [semithick, purple, opacity=0.9]
table {%
	0 1.56030714511871
	1 2.74217963218689
	2 3.61192488670349
	3 10.9496469497681
	4 14.2314796447754
	5 15.4385757446289
	6 16.1190891265869
	7 18.4008541107178
	8 18.822265625
	9 18.8615665435791
	10 19.2555446624756
	11 21.3419990539551
	12 19.3028335571289
	13 19.9790534973145
	14 19.7369289398193
	15 17.9661064147949
	16 19.6370220184326
	17 20.3802585601807
	18 19.0772266387939
	19 17.2774887084961
	20 19.1354866027832
	21 20.0726852416992
	22 18.7396659851074
	23 20.5477924346924
	24 18.1819839477539
	25 19.0043067932129
	26 18.4754219055176
	27 19.0304412841797
	28 18.8363342285156
	29 20.72336769104
	30 19.4595508575439
	31 19.3815670013428
	32 19.957347869873
	33 16.3111381530762
	34 15.9000186920166
	35 19.5504055023193
	36 19.0334491729736
	37 20.697416305542
	38 17.8588829040527
	39 18.7379665374756
	40 19.7688674926758
	41 19.4993953704834
	42 19.6784496307373
	43 17.7230091094971
	44 18.1128196716309
	45 19.0254383087158
	46 19.0749320983887
	47 23.0771179199219
	48 23.0290241241455
	49 19.7329559326172
	50 21.8987445831299
	51 20.8713760375977
	52 20.4424667358398
	53 19.537727355957
	54 19.9357833862305
	55 20.7625789642334
	56 21.3459625244141
	57 18.9098377227783
	58 19.9310474395752
	59 21.6399459838867
	60 21.9056644439697
	61 20.5229225158691
	62 20.8893852233887
	63 22.1898040771484
	64 20.9730796813965
	65 20.7122020721436
	66 21.2745418548584
	67 21.6046867370605
	68 21.1246795654297
	69 20.198278427124
	70 19.0641250610352
	71 21.4521884918213
	72 20.2343940734863
	73 20.9945869445801
	74 21.1951751708984
	75 19.8174686431885
	76 21.2938442230225
	77 21.8078517913818
	78 21.3669586181641
	79 22.7700576782227
	80 23.444091796875
	81 24.2506256103516
	82 22.6705169677734
	83 22.2169303894043
	84 22.4409942626953
	85 23.2005615234375
	86 23.8889770507812
	87 20.4991703033447
	88 18.6027450561523
	89 18.0065937042236
	90 20.1754989624023
	91 21.200777053833
	92 21.4736022949219
	93 19.7607326507568
	94 21.3366718292236
	95 20.1267108917236
	96 21.045108795166
};
\addlegendentry{GP}
\addplot [semithick, purple, opacity=0.9, dashed, forget plot]
table {%
0 19.4747409820557
1 19.4747409820557
2 19.4747409820557
3 19.4747409820557
4 19.4747409820557
5 19.4747409820557
6 19.4747409820557
7 19.4747409820557
8 19.4747409820557
9 19.4747409820557
10 19.4747409820557
11 19.4747409820557
12 19.4747409820557
13 19.4747409820557
14 19.4747409820557
15 19.4747409820557
16 19.4747409820557
17 19.4747409820557
18 19.4747409820557
19 19.4747409820557
20 19.4747409820557
21 19.4747409820557
22 19.4747409820557
23 19.4747409820557
24 19.4747409820557
25 19.4747409820557
26 19.4747409820557
27 19.4747409820557
28 19.4747409820557
29 19.4747409820557
30 19.4747409820557
31 19.4747409820557
32 19.4747409820557
33 19.4747409820557
34 19.4747409820557
35 19.4747409820557
36 19.4747409820557
37 19.4747409820557
38 19.4747409820557
39 19.4747409820557
40 19.4747409820557
41 19.4747409820557
42 19.4747409820557
43 19.4747409820557
44 19.4747409820557
45 19.4747409820557
46 19.4747409820557
47 19.4747409820557
48 19.4747409820557
49 19.4747409820557
50 19.4747409820557
51 19.4747409820557
52 19.4747409820557
53 19.4747409820557
54 19.4747409820557
55 19.4747409820557
56 19.4747409820557
57 19.4747409820557
58 19.4747409820557
59 19.4747409820557
60 19.4747409820557
61 19.4747409820557
62 19.4747409820557
63 19.4747409820557
64 19.4747409820557
65 19.4747409820557
66 19.4747409820557
67 19.4747409820557
68 19.4747409820557
69 19.4747409820557
70 19.4747409820557
71 19.4747409820557
72 19.4747409820557
73 19.4747409820557
74 19.4747409820557
75 19.4747409820557
76 19.4747409820557
77 19.4747409820557
78 19.4747409820557
79 19.4747409820557
80 19.4747409820557
81 19.4747409820557
82 19.4747409820557
83 19.4747409820557
84 19.4747409820557
85 19.4747409820557
86 19.4747409820557
87 19.4747409820557
88 19.4747409820557
89 19.4747409820557
90 19.4747409820557
91 19.4747409820557
92 19.4747409820557
93 19.4747409820557
94 19.4747409820557
95 19.4747409820557
96 19.4747409820557
};
\addplot [semithick, red, opacity=0.9]
table {%
0 13.1839437484741
1 11.7498807907104
2 13.6543350219727
3 12.6760339736938
4 14.1595010757446
5 14.6697607040405
6 14.8817491531372
7 16.1495800018311
8 16.1874942779541
9 16.1218318939209
10 16.9310169219971
11 14.724648475647
12 15.6434850692749
13 16.2127437591553
14 15.6237840652466
15 15.589599609375
16 14.1990022659302
17 14.750147819519
18 14.2747755050659
19 14.6913585662842
20 13.7355346679688
21 13.9580373764038
22 15.0687608718872
23 16.528694152832
24 16.1839485168457
25 17.2055358886719
26 15.2163228988647
27 14.7326240539551
28 16.5552940368652
29 16.7195053100586
30 14.4976673126221
31 14.2190341949463
32 14.9561710357666
33 14.738356590271
34 13.8206081390381
35 14.4566555023193
36 12.7833223342896
37 14.8651313781738
38 14.492748260498
39 15.2777462005615
40 16.3478355407715
41 16.44118309021
42 16.5618839263916
43 16.3169727325439
44 16.6152477264404
45 14.2725877761841
46 16.8067398071289
47 16.3544692993164
48 14.5933284759521
49 13.5834169387817
50 12.7403564453125
51 14.2139949798584
52 15.1322641372681
53 14.6044254302979
54 17.4660682678223
55 14.7173347473145
56 16.8055477142334
57 16.6060123443604
58 16.5778999328613
59 15.3524589538574
60 16.9637489318848
61 16.2495002746582
62 14.3352346420288
63 16.0898952484131
64 18.1196537017822
65 16.508617401123
66 13.9357824325562
67 14.1219863891602
68 14.6012992858887
69 15.2941551208496
70 16.4552116394043
71 15.1909322738647
72 14.5074043273926
73 15.9100694656372
74 16.5011234283447
75 17.0372734069824
76 16.8216648101807
77 18.9184989929199
78 17.9277172088623
79 17.5136566162109
80 16.388801574707
81 16.8684196472168
82 16.5771732330322
83 16.5034980773926
84 16.3192596435547
85 17.3600006103516
86 18.1472949981689
87 16.098258972168
88 16.5327587127686
89 15.2860097885132
90 15.6349906921387
91 15.7958993911743
92 16.9007148742676
93 18.2122821807861
94 16.770959854126
95 16.0863838195801
96 15.4306735992432
};
\addlegendentry{EKF}
\addplot [semithick, red, opacity=0.9, dashed, forget plot]
table {%
0 15.5887956619263
1 15.5887956619263
2 15.5887956619263
3 15.5887956619263
4 15.5887956619263
5 15.5887956619263
6 15.5887956619263
7 15.5887956619263
8 15.5887956619263
9 15.5887956619263
10 15.5887956619263
11 15.5887956619263
12 15.5887956619263
13 15.5887956619263
14 15.5887956619263
15 15.5887956619263
16 15.5887956619263
17 15.5887956619263
18 15.5887956619263
19 15.5887956619263
20 15.5887956619263
21 15.5887956619263
22 15.5887956619263
23 15.5887956619263
24 15.5887956619263
25 15.5887956619263
26 15.5887956619263
27 15.5887956619263
28 15.5887956619263
29 15.5887956619263
30 15.5887956619263
31 15.5887956619263
32 15.5887956619263
33 15.5887956619263
34 15.5887956619263
35 15.5887956619263
36 15.5887956619263
37 15.5887956619263
38 15.5887956619263
39 15.5887956619263
40 15.5887956619263
41 15.5887956619263
42 15.5887956619263
43 15.5887956619263
44 15.5887956619263
45 15.5887956619263
46 15.5887956619263
47 15.5887956619263
48 15.5887956619263
49 15.5887956619263
50 15.5887956619263
51 15.5887956619263
52 15.5887956619263
53 15.5887956619263
54 15.5887956619263
55 15.5887956619263
56 15.5887956619263
57 15.5887956619263
58 15.5887956619263
59 15.5887956619263
60 15.5887956619263
61 15.5887956619263
62 15.5887956619263
63 15.5887956619263
64 15.5887956619263
65 15.5887956619263
66 15.5887956619263
67 15.5887956619263
68 15.5887956619263
69 15.5887956619263
70 15.5887956619263
71 15.5887956619263
72 15.5887956619263
73 15.5887956619263
74 15.5887956619263
75 15.5887956619263
76 15.5887956619263
77 15.5887956619263
78 15.5887956619263
79 15.5887956619263
80 15.5887956619263
81 15.5887956619263
82 15.5887956619263
83 15.5887956619263
84 15.5887956619263
85 15.5887956619263
86 15.5887956619263
87 15.5887956619263
88 15.5887956619263
89 15.5887956619263
90 15.5887956619263
91 15.5887956619263
92 15.5887956619263
93 15.5887956619263
94 15.5887956619263
95 15.5887956619263
96 15.5887956619263
};
\addplot [semithick, black, opacity=0.9, mark=x, mark size=3, mark options={solid}]
table {%
0 13.1839437484741
1 15.9896545410156
2 13.867880821228
3 14.7092027664185
4 18.3605365753174
5 16.7674674987793
6 14.9182348251343
7 15.6318788528442
8 14.7217264175415
9 13.9438953399658
10 15.9824094772339
11 13.6165084838867
12 14.4943180084229
13 16.8203601837158
14 14.6307325363159
15 15.4902696609497
16 11.9295539855957
17 13.7931203842163
18 13.823016166687
19 13.9921407699585
20 12.3312997817993
21 15.8135433197021
22 15.0151720046997
23 14.5668210983276
24 14.5276746749878
25 14.4882068634033
26 13.9100856781006
27 13.2583751678467
28 14.9425535202026
29 16.2055206298828
30 11.6449089050293
31 12.092755317688
32 12.5512266159058
33 15.8596935272217
34 10.9728813171387
35 14.5666246414185
36 12.2668590545654
37 14.4842624664307
38 13.5005302429199
39 16.1150665283203
40 17.1802043914795
41 17.0659122467041
42 12.205753326416
43 13.8765392303467
44 14.4287757873535
45 12.3787508010864
46 16.1644973754883
47 13.9942684173584
48 14.7310085296631
49 12.2735872268677
50 12.9925403594971
51 13.5361766815186
52 13.6683130264282
53 16.1655807495117
54 18.0310897827148
55 13.4524793624878
56 16.8783874511719
57 15.0577869415283
58 14.792896270752
59 12.2660074234009
60 16.1155128479004
61 13.4383249282837
62 13.0288925170898
63 13.8526487350464
64 16.8131504058838
65 15.3843946456909
66 14.6525182723999
67 14.1252355575562
68 13.8502502441406
69 13.0684471130371
70 14.0781650543213
71 11.9355783462524
72 16.9747409820557
73 14.6308813095093
74 14.2783803939819
75 13.5612173080444
76 13.3538999557495
77 15.2429838180542
78 14.3507823944092
79 12.8196907043457
80 15.6162195205688
81 13.438943862915
82 12.432258605957
83 17.1009273529053
84 13.4039220809937
85 15.3619413375854
86 14.9507102966309
87 14.3290462493896
88 13.2657222747803
89 13.8780870437622
90 14.7275772094727
91 15.1292171478271
92 15.9924869537354
93 15.9047574996948
94 14.1250352859497
95 12.7294731140137
96 15.6693391799927
};
\addlegendentry{Measurements}
\addplot [semithick, black, opacity=0.3, dashed, forget plot]
table {%
0 14.435133934021
1 14.435133934021
2 14.435133934021
3 14.435133934021
4 14.435133934021
5 14.435133934021
6 14.435133934021
7 14.435133934021
8 14.435133934021
9 14.435133934021
10 14.435133934021
11 14.435133934021
12 14.435133934021
13 14.435133934021
14 14.435133934021
15 14.435133934021
16 14.435133934021
17 14.435133934021
18 14.435133934021
19 14.435133934021
20 14.435133934021
21 14.435133934021
22 14.435133934021
23 14.435133934021
24 14.435133934021
25 14.435133934021
26 14.435133934021
27 14.435133934021
28 14.435133934021
29 14.435133934021
30 14.435133934021
31 14.435133934021
32 14.435133934021
33 14.435133934021
34 14.435133934021
35 14.435133934021
36 14.435133934021
37 14.435133934021
38 14.435133934021
39 14.435133934021
40 14.435133934021
41 14.435133934021
42 14.435133934021
43 14.435133934021
44 14.435133934021
45 14.435133934021
46 14.435133934021
47 14.435133934021
48 14.435133934021
49 14.435133934021
50 14.435133934021
51 14.435133934021
52 14.435133934021
53 14.435133934021
54 14.435133934021
55 14.435133934021
56 14.435133934021
57 14.435133934021
58 14.435133934021
59 14.435133934021
60 14.435133934021
61 14.435133934021
62 14.435133934021
63 14.435133934021
64 14.435133934021
65 14.435133934021
66 14.435133934021
67 14.435133934021
68 14.435133934021
69 14.435133934021
70 14.435133934021
71 14.435133934021
72 14.435133934021
73 14.435133934021
74 14.435133934021
75 14.435133934021
76 14.435133934021
77 14.435133934021
78 14.435133934021
79 14.435133934021
80 14.435133934021
81 14.435133934021
82 14.435133934021
83 14.435133934021
84 14.435133934021
85 14.435133934021
86 14.435133934021
87 14.435133934021
88 14.435133934021
89 14.435133934021
90 14.435133934021
91 14.435133934021
92 14.435133934021
93 14.435133934021
94 14.435133934021
95 14.435133934021
96 14.435133934021
};
\end{axis}

\end{tikzpicture}

%% file: figures/gct_post_avg_test.tikz
\begin{tikzpicture}

\definecolor{darkgray176}{RGB}{176,176,176}
\definecolor{green}{RGB}{0,128,0}
\definecolor{lightgray204}{RGB}{204,204,204}
\definecolor{purple}{RGB}{128,0,128}

\begin{axis}[
	width=\linewidth,
	height=0.8\linewidth,
	legend cell align={left},
	legend style={fill opacity=0.8, draw opacity=1, text opacity=1, draw=lightgray204},
	tick align=outside,
	tick pos=left,
	x grid style={darkgray176},
	xlabel={\footnotesize Time [s]},
	xmin=4, xmax=96,
	xtick style={color=black},
	y grid style={darkgray176},
	ylabel={\footnotesize RMSE [\si{\metre}]},
	ymin=5, ymax=27,
	ytick style={color=black},
	legend style={at={(0.97,0.97)},anchor=north east,legend cell align=left,draw=white!15!black,fill opacity=0.5,draw opacity=1,text opacity=1,font=\scriptsize, legend columns=2}
]
\path [fill=blue, fill opacity=0.1]
(axis cs:0,21.7148323059082)
--(axis cs:0,2.08285331726074)
--(axis cs:1,3.09840869903564)
--(axis cs:2,4.9767632484436)
--(axis cs:3,6.49474239349365)
--(axis cs:4,6.18614959716797)
--(axis cs:5,6.99565172195435)
--(axis cs:6,6.63976335525513)
--(axis cs:7,6.61631965637207)
--(axis cs:8,6.87242412567139)
--(axis cs:9,7.27603816986084)
--(axis cs:10,7.02590703964233)
--(axis cs:11,6.69410610198975)
--(axis cs:12,5.14774751663208)
--(axis cs:13,6.26605463027954)
--(axis cs:14,5.89288902282715)
--(axis cs:15,5.29903697967529)
--(axis cs:16,5.40767383575439)
--(axis cs:17,4.09589147567749)
--(axis cs:18,5.11192035675049)
--(axis cs:19,5.127610206604)
--(axis cs:20,4.21951484680176)
--(axis cs:21,4.15783882141113)
--(axis cs:22,7.08782005310059)
--(axis cs:23,5.48874235153198)
--(axis cs:24,7.63939905166626)
--(axis cs:25,7.57658529281616)
--(axis cs:26,7.3891716003418)
--(axis cs:27,8.40401840209961)
--(axis cs:28,7.41043090820312)
--(axis cs:29,7.0058741569519)
--(axis cs:30,6.34590768814087)
--(axis cs:31,8.72825050354004)
--(axis cs:32,7.22288990020752)
--(axis cs:33,5.9196891784668)
--(axis cs:34,6.61820411682129)
--(axis cs:35,5.75169801712036)
--(axis cs:36,5.72763013839722)
--(axis cs:37,5.90805387496948)
--(axis cs:38,6.06753587722778)
--(axis cs:39,6.92602968215942)
--(axis cs:40,5.30482244491577)
--(axis cs:41,6.35527467727661)
--(axis cs:42,7.63269424438477)
--(axis cs:43,6.6952338218689)
--(axis cs:44,6.22498178482056)
--(axis cs:45,6.48292350769043)
--(axis cs:46,6.31476640701294)
--(axis cs:47,5.65568923950195)
--(axis cs:48,6.08829021453857)
--(axis cs:49,6.0941162109375)
--(axis cs:50,5.64777898788452)
--(axis cs:51,6.29626560211182)
--(axis cs:52,6.56927490234375)
--(axis cs:53,7.91044902801514)
--(axis cs:54,6.63949871063232)
--(axis cs:55,6.24608850479126)
--(axis cs:56,7.39777660369873)
--(axis cs:57,7.4129958152771)
--(axis cs:58,7.29315137863159)
--(axis cs:59,8.0329761505127)
--(axis cs:60,7.01664972305298)
--(axis cs:61,6.42779350280762)
--(axis cs:62,7.68297863006592)
--(axis cs:63,7.82605648040771)
--(axis cs:64,6.91067028045654)
--(axis cs:65,5.05465698242188)
--(axis cs:66,5.61871910095215)
--(axis cs:67,5.37725830078125)
--(axis cs:68,7.26583766937256)
--(axis cs:69,6.85723876953125)
--(axis cs:70,7.34565114974976)
--(axis cs:71,6.18773889541626)
--(axis cs:72,7.37746286392212)
--(axis cs:73,6.58628177642822)
--(axis cs:74,6.5668306350708)
--(axis cs:75,7.74629878997803)
--(axis cs:76,7.87313985824585)
--(axis cs:77,6.79268264770508)
--(axis cs:78,5.6533031463623)
--(axis cs:79,6.49949550628662)
--(axis cs:80,7.49001741409302)
--(axis cs:81,7.36588191986084)
--(axis cs:82,6.61560249328613)
--(axis cs:83,6.6098051071167)
--(axis cs:84,5.85328435897827)
--(axis cs:85,7.87733030319214)
--(axis cs:86,5.50000143051147)
--(axis cs:87,6.81538772583008)
--(axis cs:88,6.6154613494873)
--(axis cs:89,7.18513202667236)
--(axis cs:90,6.42636251449585)
--(axis cs:91,5.80150318145752)
--(axis cs:92,5.23655128479004)
--(axis cs:93,5.2292366027832)
--(axis cs:94,6.79866313934326)
--(axis cs:95,7.52877950668335)
--(axis cs:96,8.36234092712402)
--(axis cs:96,20.1887702941895)
--(axis cs:96,20.1887702941895)
--(axis cs:95,18.7108669281006)
--(axis cs:94,21.0587615966797)
--(axis cs:93,22.6730079650879)
--(axis cs:92,24.4468936920166)
--(axis cs:91,21.9422454833984)
--(axis cs:90,19.5348854064941)
--(axis cs:89,19.7458190917969)
--(axis cs:88,19.497220993042)
--(axis cs:87,20.6898899078369)
--(axis cs:86,21.3239059448242)
--(axis cs:85,21.274471282959)
--(axis cs:84,21.6582336425781)
--(axis cs:83,20.2669830322266)
--(axis cs:82,20.7691459655762)
--(axis cs:81,20.5114212036133)
--(axis cs:80,21.1561908721924)
--(axis cs:79,21.4534530639648)
--(axis cs:78,24.306037902832)
--(axis cs:77,22.4971237182617)
--(axis cs:76,22.1831188201904)
--(axis cs:75,20.2168159484863)
--(axis cs:74,21.388111114502)
--(axis cs:73,20.2174987792969)
--(axis cs:72,18.8752326965332)
--(axis cs:71,19.1839714050293)
--(axis cs:70,18.7337398529053)
--(axis cs:69,19.2792453765869)
--(axis cs:68,17.7932434082031)
--(axis cs:67,18.181734085083)
--(axis cs:66,17.6182994842529)
--(axis cs:65,18.2218914031982)
--(axis cs:64,19.5131912231445)
--(axis cs:63,20.3476486206055)
--(axis cs:62,18.6388664245605)
--(axis cs:61,19.0826206207275)
--(axis cs:60,19.4133949279785)
--(axis cs:59,20.161750793457)
--(axis cs:58,18.6030941009521)
--(axis cs:57,20.0494613647461)
--(axis cs:56,19.6666221618652)
--(axis cs:55,20.1836185455322)
--(axis cs:54,18.4152145385742)
--(axis cs:53,19.6649169921875)
--(axis cs:52,17.0193576812744)
--(axis cs:51,18.2214889526367)
--(axis cs:50,18.870885848999)
--(axis cs:49,16.6483402252197)
--(axis cs:48,18.6337242126465)
--(axis cs:47,20.1663150787354)
--(axis cs:46,20.6931629180908)
--(axis cs:45,21.7114734649658)
--(axis cs:44,18.8373527526855)
--(axis cs:43,20.9587059020996)
--(axis cs:42,21.350191116333)
--(axis cs:41,21.0850486755371)
--(axis cs:40,20.9535579681396)
--(axis cs:39,18.9633483886719)
--(axis cs:38,18.2937641143799)
--(axis cs:37,17.1328220367432)
--(axis cs:36,18.1620864868164)
--(axis cs:35,16.1043758392334)
--(axis cs:34,17.832010269165)
--(axis cs:33,17.8841686248779)
--(axis cs:32,17.6521949768066)
--(axis cs:31,16.967134475708)
--(axis cs:30,17.5528011322021)
--(axis cs:29,18.0063819885254)
--(axis cs:28,20.1933441162109)
--(axis cs:27,19.7301273345947)
--(axis cs:26,18.2023735046387)
--(axis cs:25,19.179141998291)
--(axis cs:24,20.0382518768311)
--(axis cs:23,20.0317192077637)
--(axis cs:22,19.2846755981445)
--(axis cs:21,19.7250175476074)
--(axis cs:20,19.7424850463867)
--(axis cs:19,18.4222030639648)
--(axis cs:18,19.5535316467285)
--(axis cs:17,19.9520702362061)
--(axis cs:16,19.1053161621094)
--(axis cs:15,18.5986862182617)
--(axis cs:14,18.9522724151611)
--(axis cs:13,18.2924003601074)
--(axis cs:12,20.381649017334)
--(axis cs:11,18.8214988708496)
--(axis cs:10,17.5015411376953)
--(axis cs:9,19.056568145752)
--(axis cs:8,18.3677940368652)
--(axis cs:7,17.4856090545654)
--(axis cs:6,17.3549251556396)
--(axis cs:5,15.3490028381348)
--(axis cs:4,15.6388092041016)
--(axis cs:3,16.5588226318359)
--(axis cs:2,16.7728595733643)
--(axis cs:1,17.4040145874023)
--(axis cs:0,21.7148323059082)
--cycle;

\path [fill=green, fill opacity=0.1]
(axis cs:0,21.8722991943359)
--(axis cs:0,2.13152313232422)
--(axis cs:1,3.0949010848999)
--(axis cs:2,5.40106630325317)
--(axis cs:3,7.10391235351562)
--(axis cs:4,8.14694976806641)
--(axis cs:5,8.79400062561035)
--(axis cs:6,8.28070640563965)
--(axis cs:7,7.3240327835083)
--(axis cs:8,5.70418405532837)
--(axis cs:9,5.03078079223633)
--(axis cs:10,4.83746862411499)
--(axis cs:11,5.73351049423218)
--(axis cs:12,5.85992527008057)
--(axis cs:13,5.45054006576538)
--(axis cs:14,6.11131286621094)
--(axis cs:15,6.90406656265259)
--(axis cs:16,6.96854829788208)
--(axis cs:17,6.13187122344971)
--(axis cs:18,7.11064672470093)
--(axis cs:19,7.50123834609985)
--(axis cs:20,6.83753871917725)
--(axis cs:21,7.61385726928711)
--(axis cs:22,7.62072372436523)
--(axis cs:23,8.51078224182129)
--(axis cs:24,8.22103881835938)
--(axis cs:25,7.49120664596558)
--(axis cs:26,7.53709316253662)
--(axis cs:27,8.35990142822266)
--(axis cs:28,8.18264198303223)
--(axis cs:29,7.20124816894531)
--(axis cs:30,6.84923982620239)
--(axis cs:31,6.1332368850708)
--(axis cs:32,6.14283514022827)
--(axis cs:33,6.08379745483398)
--(axis cs:34,6.49643325805664)
--(axis cs:35,6.06857585906982)
--(axis cs:36,5.95967721939087)
--(axis cs:37,6.16246509552002)
--(axis cs:38,6.43867015838623)
--(axis cs:39,6.89074230194092)
--(axis cs:40,7.21438074111938)
--(axis cs:41,7.60894727706909)
--(axis cs:42,8.19928169250488)
--(axis cs:43,9.13819313049316)
--(axis cs:44,8.60297203063965)
--(axis cs:45,8.03077697753906)
--(axis cs:46,6.91845703125)
--(axis cs:47,6.95362710952759)
--(axis cs:48,7.05288887023926)
--(axis cs:49,6.23453617095947)
--(axis cs:50,5.99534702301025)
--(axis cs:51,6.05602407455444)
--(axis cs:52,6.98738241195679)
--(axis cs:53,8.04423904418945)
--(axis cs:54,7.92569160461426)
--(axis cs:55,6.7370285987854)
--(axis cs:56,6.42216730117798)
--(axis cs:57,6.8628625869751)
--(axis cs:58,6.95552110671997)
--(axis cs:59,6.40751934051514)
--(axis cs:60,6.86496067047119)
--(axis cs:61,7.46617078781128)
--(axis cs:62,7.99006366729736)
--(axis cs:63,8.05658531188965)
--(axis cs:64,7.23481845855713)
--(axis cs:65,6.38844537734985)
--(axis cs:66,6.08774042129517)
--(axis cs:67,5.4615626335144)
--(axis cs:68,4.75959825515747)
--(axis cs:69,5.02541255950928)
--(axis cs:70,5.81409358978271)
--(axis cs:71,6.22601938247681)
--(axis cs:72,6.51272106170654)
--(axis cs:73,7.06150197982788)
--(axis cs:74,7.03816318511963)
--(axis cs:75,6.7659764289856)
--(axis cs:76,8.05552291870117)
--(axis cs:77,8.3675479888916)
--(axis cs:78,8.26142024993896)
--(axis cs:79,7.69278907775879)
--(axis cs:80,6.81988573074341)
--(axis cs:81,7.10413837432861)
--(axis cs:82,7.47308492660522)
--(axis cs:83,6.46972560882568)
--(axis cs:84,6.09537649154663)
--(axis cs:85,6.39034509658813)
--(axis cs:86,7.40923166275024)
--(axis cs:87,8.02074909210205)
--(axis cs:88,8.12761688232422)
--(axis cs:89,8.47712326049805)
--(axis cs:90,8.15485382080078)
--(axis cs:91,7.51487588882446)
--(axis cs:92,8.20792102813721)
--(axis cs:93,8.32221412658691)
--(axis cs:94,8.30779075622559)
--(axis cs:95,6.77713489532471)
--(axis cs:96,6.35081243515015)
--(axis cs:96,20.0295257568359)
--(axis cs:96,20.0295257568359)
--(axis cs:95,19.2103424072266)
--(axis cs:94,18.5989322662354)
--(axis cs:93,18.8740825653076)
--(axis cs:92,19.3956871032715)
--(axis cs:91,19.6748943328857)
--(axis cs:90,19.6918239593506)
--(axis cs:89,19.8712482452393)
--(axis cs:88,20.2362194061279)
--(axis cs:87,20.2973823547363)
--(axis cs:86,19.1997013092041)
--(axis cs:85,18.7131004333496)
--(axis cs:84,18.8735733032227)
--(axis cs:83,18.7918395996094)
--(axis cs:82,18.2267589569092)
--(axis cs:81,18.7888565063477)
--(axis cs:80,19.3461933135986)
--(axis cs:79,19.5639781951904)
--(axis cs:78,19.8753852844238)
--(axis cs:77,20.0067710876465)
--(axis cs:76,20.395393371582)
--(axis cs:75,19.5718879699707)
--(axis cs:74,19.2677803039551)
--(axis cs:73,18.6886444091797)
--(axis cs:72,18.4729537963867)
--(axis cs:71,19.4606018066406)
--(axis cs:70,20.3984336853027)
--(axis cs:69,20.3750038146973)
--(axis cs:68,19.9075412750244)
--(axis cs:67,19.6822090148926)
--(axis cs:66,19.8638343811035)
--(axis cs:65,20.4193096160889)
--(axis cs:64,19.8740463256836)
--(axis cs:63,18.3687953948975)
--(axis cs:62,17.5795707702637)
--(axis cs:61,18.4883193969727)
--(axis cs:60,19.065372467041)
--(axis cs:59,18.81982421875)
--(axis cs:58,18.2704601287842)
--(axis cs:57,18.5395965576172)
--(axis cs:56,18.9134750366211)
--(axis cs:55,19.1553611755371)
--(axis cs:54,18.7535076141357)
--(axis cs:53,17.9024696350098)
--(axis cs:52,18.0378684997559)
--(axis cs:51,19.6799182891846)
--(axis cs:50,19.7356262207031)
--(axis cs:49,19.7337493896484)
--(axis cs:48,19.6412220001221)
--(axis cs:47,19.5182800292969)
--(axis cs:46,19.8116836547852)
--(axis cs:45,20.5594921112061)
--(axis cs:44,20.9962482452393)
--(axis cs:43,21.2993927001953)
--(axis cs:42,20.697078704834)
--(axis cs:41,19.827522277832)
--(axis cs:40,19.4584522247314)
--(axis cs:39,19.3400917053223)
--(axis cs:38,19.1971244812012)
--(axis cs:37,18.4617767333984)
--(axis cs:36,17.8273906707764)
--(axis cs:35,17.4027633666992)
--(axis cs:34,17.7564506530762)
--(axis cs:33,18.2263145446777)
--(axis cs:32,18.1872615814209)
--(axis cs:31,18.1019172668457)
--(axis cs:30,17.5518569946289)
--(axis cs:29,17.6477279663086)
--(axis cs:28,17.5335788726807)
--(axis cs:27,18.0147819519043)
--(axis cs:26,19.25634765625)
--(axis cs:25,19.7365608215332)
--(axis cs:24,19.4601402282715)
--(axis cs:23,19.5355625152588)
--(axis cs:22,19.8998565673828)
--(axis cs:21,19.4602222442627)
--(axis cs:20,19.5763397216797)
--(axis cs:19,20.4176616668701)
--(axis cs:18,21.2346878051758)
--(axis cs:17,20.8926582336426)
--(axis cs:16,19.3178844451904)
--(axis cs:15,18.5134468078613)
--(axis cs:14,17.529670715332)
--(axis cs:13,17.4878425598145)
--(axis cs:12,18.1108093261719)
--(axis cs:11,18.4021072387695)
--(axis cs:10,19.5007438659668)
--(axis cs:9,19.2831687927246)
--(axis cs:8,18.5558624267578)
--(axis cs:7,18.6969299316406)
--(axis cs:6,19.4515743255615)
--(axis cs:5,20.1108837127686)
--(axis cs:4,20.6656036376953)
--(axis cs:3,20.1799716949463)
--(axis cs:2,18.1595497131348)
--(axis cs:1,17.8094749450684)
--(axis cs:0,21.8722991943359)
--cycle;

\path [fill=purple, fill opacity=0.1]
(axis cs:0,10.1704654693604)
--(axis cs:0,3.30607938766479)
--(axis cs:1,3.20776963233948)
--(axis cs:2,2.63629865646362)
--(axis cs:3,3.05047082901001)
--(axis cs:4,2.85902214050293)
--(axis cs:5,3.48690938949585)
--(axis cs:6,4.42828750610352)
--(axis cs:7,3.78808689117432)
--(axis cs:8,6.05818319320679)
--(axis cs:9,6.49007558822632)
--(axis cs:10,7.20859909057617)
--(axis cs:11,5.69744777679443)
--(axis cs:12,2.29622554779053)
--(axis cs:13,5.84415054321289)
--(axis cs:14,4.00105571746826)
--(axis cs:15,3.95192241668701)
--(axis cs:16,4.32051086425781)
--(axis cs:17,4.36221599578857)
--(axis cs:18,2.63394546508789)
--(axis cs:19,2.04404449462891)
--(axis cs:20,3.16490173339844)
--(axis cs:21,3.10076904296875)
--(axis cs:22,3.93887901306152)
--(axis cs:23,4.22132682800293)
--(axis cs:24,4.9404821395874)
--(axis cs:25,5.02299213409424)
--(axis cs:26,4.23128223419189)
--(axis cs:27,4.99328804016113)
--(axis cs:28,3.62486457824707)
--(axis cs:29,4.07669258117676)
--(axis cs:30,5.49896430969238)
--(axis cs:31,5.67351722717285)
--(axis cs:32,4.50606107711792)
--(axis cs:33,3.73430585861206)
--(axis cs:34,3.58558750152588)
--(axis cs:35,3.42502880096436)
--(axis cs:36,1.66903114318848)
--(axis cs:37,5.13279056549072)
--(axis cs:38,5.62481498718262)
--(axis cs:39,7.57330131530762)
--(axis cs:40,7.36032009124756)
--(axis cs:41,6.08763885498047)
--(axis cs:42,5.8002462387085)
--(axis cs:43,3.42833995819092)
--(axis cs:44,3.59620380401611)
--(axis cs:45,4.15628910064697)
--(axis cs:46,5.48808765411377)
--(axis cs:47,4.783034324646)
--(axis cs:48,5.67276191711426)
--(axis cs:49,5.16354751586914)
--(axis cs:50,4.05662631988525)
--(axis cs:51,6.03568744659424)
--(axis cs:52,6.32121753692627)
--(axis cs:53,4.3416109085083)
--(axis cs:54,5.37652206420898)
--(axis cs:55,5.33694267272949)
--(axis cs:56,4.68190288543701)
--(axis cs:57,5.94252395629883)
--(axis cs:58,6.73019599914551)
--(axis cs:59,6.23934745788574)
--(axis cs:60,5.15752983093262)
--(axis cs:61,7.31526756286621)
--(axis cs:62,7.50124359130859)
--(axis cs:63,4.81335926055908)
--(axis cs:64,6.1517391204834)
--(axis cs:65,5.68857860565186)
--(axis cs:66,6.77075672149658)
--(axis cs:67,5.72012519836426)
--(axis cs:68,6.30141544342041)
--(axis cs:69,5.4810962677002)
--(axis cs:70,5.95918846130371)
--(axis cs:71,5.43866348266602)
--(axis cs:72,6.88849258422852)
--(axis cs:73,6.14068126678467)
--(axis cs:74,4.82559299468994)
--(axis cs:75,4.94007301330566)
--(axis cs:76,4.76076030731201)
--(axis cs:77,3.26760578155518)
--(axis cs:78,5.65968322753906)
--(axis cs:79,5.73438453674316)
--(axis cs:80,6.82522773742676)
--(axis cs:81,6.94861030578613)
--(axis cs:82,5.06446647644043)
--(axis cs:83,6.40379619598389)
--(axis cs:84,7.17978382110596)
--(axis cs:85,7.97115707397461)
--(axis cs:86,8.75151634216309)
--(axis cs:87,7.25137901306152)
--(axis cs:88,5.4180006980896)
--(axis cs:89,6.08282327651978)
--(axis cs:90,5.79253196716309)
--(axis cs:91,5.97683906555176)
--(axis cs:92,6.43727493286133)
--(axis cs:93,5.61734771728516)
--(axis cs:94,6.24769496917725)
--(axis cs:95,5.07527923583984)
--(axis cs:96,5.96713447570801)
--(axis cs:96,24.5221118927002)
--(axis cs:96,24.5221118927002)
--(axis cs:95,24.308895111084)
--(axis cs:94,23.8996772766113)
--(axis cs:93,25.8631687164307)
--(axis cs:92,23.785888671875)
--(axis cs:91,23.5445213317871)
--(axis cs:90,22.9774055480957)
--(axis cs:89,20.3900852203369)
--(axis cs:88,20.7329063415527)
--(axis cs:87,20.9339256286621)
--(axis cs:86,23.8231639862061)
--(axis cs:85,26.9215698242188)
--(axis cs:84,25.9876365661621)
--(axis cs:83,26.5421981811523)
--(axis cs:82,27.8639011383057)
--(axis cs:81,27.6841220855713)
--(axis cs:80,30.1687488555908)
--(axis cs:79,27.1290683746338)
--(axis cs:78,26.8539695739746)
--(axis cs:77,27.7563247680664)
--(axis cs:76,27.1653633117676)
--(axis cs:75,25.8333473205566)
--(axis cs:74,24.625617980957)
--(axis cs:73,25.5549392700195)
--(axis cs:72,23.0915546417236)
--(axis cs:71,25.0297660827637)
--(axis cs:70,25.2841339111328)
--(axis cs:69,22.6231651306152)
--(axis cs:68,23.8176460266113)
--(axis cs:67,24.3101196289062)
--(axis cs:66,25.8746452331543)
--(axis cs:65,25.0398368835449)
--(axis cs:64,25.1151256561279)
--(axis cs:63,27.7277450561523)
--(axis cs:62,25.3047485351562)
--(axis cs:61,25.2767238616943)
--(axis cs:60,24.1559524536133)
--(axis cs:59,25.971342086792)
--(axis cs:58,25.5184726715088)
--(axis cs:57,22.9923191070557)
--(axis cs:56,22.5438385009766)
--(axis cs:55,24.0896053314209)
--(axis cs:54,24.062894821167)
--(axis cs:53,26.1553382873535)
--(axis cs:52,22.581672668457)
--(axis cs:51,22.2678260803223)
--(axis cs:50,25.5582237243652)
--(axis cs:49,26.3028373718262)
--(axis cs:48,23.2682132720947)
--(axis cs:47,28.9325141906738)
--(axis cs:46,25.7271995544434)
--(axis cs:45,22.6543121337891)
--(axis cs:44,22.5793228149414)
--(axis cs:43,23.9422988891602)
--(axis cs:42,23.236083984375)
--(axis cs:41,23.6731491088867)
--(axis cs:40,22.5730476379395)
--(axis cs:39,20.7965793609619)
--(axis cs:38,22.4376735687256)
--(axis cs:37,22.2500152587891)
--(axis cs:36,29.1567687988281)
--(axis cs:35,20.9956512451172)
--(axis cs:34,23.7585372924805)
--(axis cs:33,19.1296863555908)
--(axis cs:32,19.3966083526611)
--(axis cs:31,25.7173385620117)
--(axis cs:30,23.8648529052734)
--(axis cs:29,24.3083934783936)
--(axis cs:28,25.7658023834229)
--(axis cs:27,21.866870880127)
--(axis cs:26,23.6359100341797)
--(axis cs:25,22.2424812316895)
--(axis cs:24,22.4045066833496)
--(axis cs:23,22.5239696502686)
--(axis cs:22,26.2210121154785)
--(axis cs:21,25.0888805389404)
--(axis cs:20,26.0731678009033)
--(axis cs:19,23.8198127746582)
--(axis cs:18,23.8448581695557)
--(axis cs:17,26.3380279541016)
--(axis cs:16,26.5606555938721)
--(axis cs:15,24.4535751342773)
--(axis cs:14,24.5040512084961)
--(axis cs:13,23.807559967041)
--(axis cs:12,27.9548377990723)
--(axis cs:11,23.9276237487793)
--(axis cs:10,23.7069759368896)
--(axis cs:9,20.6720371246338)
--(axis cs:8,21.9201335906982)
--(axis cs:7,22.9642715454102)
--(axis cs:6,21.2724189758301)
--(axis cs:5,17.3399925231934)
--(axis cs:4,16.7183132171631)
--(axis cs:3,15.0019493103027)
--(axis cs:2,10.2633094787598)
--(axis cs:1,10.5407752990723)
--(axis cs:0,10.1704654693604)
--cycle;

\path [fill=red, fill opacity=0.1]
(axis cs:0,21.6238594055176)
--(axis cs:0,2.08793926239014)
--(axis cs:1,3.3388500213623)
--(axis cs:2,4.56069278717041)
--(axis cs:3,6.58764982223511)
--(axis cs:4,6.65646076202393)
--(axis cs:5,6.82531547546387)
--(axis cs:6,6.72487735748291)
--(axis cs:7,7.38351106643677)
--(axis cs:8,6.91515350341797)
--(axis cs:9,7.64673185348511)
--(axis cs:10,7.03953838348389)
--(axis cs:11,6.93037223815918)
--(axis cs:12,5.04776382446289)
--(axis cs:13,6.57592916488647)
--(axis cs:14,5.74393224716187)
--(axis cs:15,5.6700701713562)
--(axis cs:16,5.67949867248535)
--(axis cs:17,3.72122573852539)
--(axis cs:18,5.35400056838989)
--(axis cs:19,5.12941598892212)
--(axis cs:20,4.16556167602539)
--(axis cs:21,4.80872249603271)
--(axis cs:22,7.43748760223389)
--(axis cs:23,6.01868629455566)
--(axis cs:24,8.65562343597412)
--(axis cs:25,7.13594102859497)
--(axis cs:26,6.7163257598877)
--(axis cs:27,8.06342697143555)
--(axis cs:28,7.45751762390137)
--(axis cs:29,7.71041250228882)
--(axis cs:30,6.84082698822021)
--(axis cs:31,8.61453056335449)
--(axis cs:32,7.11735868453979)
--(axis cs:33,5.84089612960815)
--(axis cs:34,7.03143167495728)
--(axis cs:35,6.0151572227478)
--(axis cs:36,5.60219240188599)
--(axis cs:37,5.71896171569824)
--(axis cs:38,6.22115802764893)
--(axis cs:39,7.30694484710693)
--(axis cs:40,5.94673538208008)
--(axis cs:41,6.88591861724854)
--(axis cs:42,8.48215675354004)
--(axis cs:43,7.30057621002197)
--(axis cs:44,6.66402530670166)
--(axis cs:45,6.36834812164307)
--(axis cs:46,6.31915426254272)
--(axis cs:47,6.04048109054565)
--(axis cs:48,6.13689470291138)
--(axis cs:49,6.37648963928223)
--(axis cs:50,5.6817307472229)
--(axis cs:51,6.97388935089111)
--(axis cs:52,7.24478149414062)
--(axis cs:53,8.68753242492676)
--(axis cs:54,7.16192579269409)
--(axis cs:55,7.04764413833618)
--(axis cs:56,7.6578426361084)
--(axis cs:57,7.03102874755859)
--(axis cs:58,7.49496603012085)
--(axis cs:59,8.18131351470947)
--(axis cs:60,6.91994047164917)
--(axis cs:61,6.27514934539795)
--(axis cs:62,8.28626823425293)
--(axis cs:63,9.31236457824707)
--(axis cs:64,7.22611618041992)
--(axis cs:65,5.25372123718262)
--(axis cs:66,5.8782958984375)
--(axis cs:67,6.19598913192749)
--(axis cs:68,7.96712493896484)
--(axis cs:69,7.09225702285767)
--(axis cs:70,7.91197776794434)
--(axis cs:71,5.89356994628906)
--(axis cs:72,7.28721761703491)
--(axis cs:73,7.16733169555664)
--(axis cs:74,7.03250646591187)
--(axis cs:75,7.7280478477478)
--(axis cs:76,8.12278175354004)
--(axis cs:77,7.31443119049072)
--(axis cs:78,5.61539268493652)
--(axis cs:79,7.12610244750977)
--(axis cs:80,8.00386810302734)
--(axis cs:81,7.11173677444458)
--(axis cs:82,7.35514259338379)
--(axis cs:83,7.12872838973999)
--(axis cs:84,6.54291248321533)
--(axis cs:85,7.90720415115356)
--(axis cs:86,4.80933952331543)
--(axis cs:87,6.88901567459106)
--(axis cs:88,6.43995904922485)
--(axis cs:89,6.77461528778076)
--(axis cs:90,6.46625947952271)
--(axis cs:91,6.13637256622314)
--(axis cs:92,5.47322177886963)
--(axis cs:93,5.41723728179932)
--(axis cs:94,6.80942726135254)
--(axis cs:95,7.9572057723999)
--(axis cs:96,9.0858850479126)
--(axis cs:96,21.4252014160156)
--(axis cs:96,21.4252014160156)
--(axis cs:95,19.2942237854004)
--(axis cs:94,21.4975776672363)
--(axis cs:93,22.8937110900879)
--(axis cs:92,25.0483016967773)
--(axis cs:91,22.3565330505371)
--(axis cs:90,20.1511821746826)
--(axis cs:89,20.3709297180176)
--(axis cs:88,19.8181095123291)
--(axis cs:87,21.824291229248)
--(axis cs:86,22.5386848449707)
--(axis cs:85,22.6132335662842)
--(axis cs:84,22.8867835998535)
--(axis cs:83,21.2399406433105)
--(axis cs:82,21.1040191650391)
--(axis cs:81,20.8735294342041)
--(axis cs:80,21.2571239471436)
--(axis cs:79,21.7285842895508)
--(axis cs:78,24.7657203674316)
--(axis cs:77,23.4427680969238)
--(axis cs:76,23.4105587005615)
--(axis cs:75,20.9902305603027)
--(axis cs:74,22.5128402709961)
--(axis cs:73,21.3881053924561)
--(axis cs:72,20.1841564178467)
--(axis cs:71,20.0632266998291)
--(axis cs:70,19.5815982818604)
--(axis cs:69,20.8423519134521)
--(axis cs:68,18.9958438873291)
--(axis cs:67,19.4890270233154)
--(axis cs:66,18.996021270752)
--(axis cs:65,18.7953453063965)
--(axis cs:64,21.0352993011475)
--(axis cs:63,21.6874599456787)
--(axis cs:62,19.571117401123)
--(axis cs:61,19.4365386962891)
--(axis cs:60,20.4421215057373)
--(axis cs:59,21.1731185913086)
--(axis cs:58,19.1291103363037)
--(axis cs:57,21.0213317871094)
--(axis cs:56,20.4023017883301)
--(axis cs:55,21.1064529418945)
--(axis cs:54,19.1033706665039)
--(axis cs:53,21.0238265991211)
--(axis cs:52,17.8584651947021)
--(axis cs:51,19.1951942443848)
--(axis cs:50,19.7521572113037)
--(axis cs:49,17.2248954772949)
--(axis cs:48,19.7371788024902)
--(axis cs:47,21.0880737304688)
--(axis cs:46,21.7777137756348)
--(axis cs:45,23.0294647216797)
--(axis cs:44,19.1279487609863)
--(axis cs:43,21.4322204589844)
--(axis cs:42,21.9218368530273)
--(axis cs:41,22.0518264770508)
--(axis cs:40,22.4855670928955)
--(axis cs:39,20.9185180664062)
--(axis cs:38,20.113353729248)
--(axis cs:37,18.7747974395752)
--(axis cs:36,19.9510555267334)
--(axis cs:35,16.6725482940674)
--(axis cs:34,18.327995300293)
--(axis cs:33,18.3441696166992)
--(axis cs:32,18.4916343688965)
--(axis cs:31,17.864351272583)
--(axis cs:30,18.4189987182617)
--(axis cs:29,18.620719909668)
--(axis cs:28,21.9212741851807)
--(axis cs:27,20.3828964233398)
--(axis cs:26,19.1328220367432)
--(axis cs:25,20.2483253479004)
--(axis cs:24,20.7675704956055)
--(axis cs:23,21.3814430236816)
--(axis cs:22,20.8779487609863)
--(axis cs:21,21.0331115722656)
--(axis cs:20,21.3233413696289)
--(axis cs:19,19.7630825042725)
--(axis cs:18,20.5299625396729)
--(axis cs:17,20.7841510772705)
--(axis cs:16,19.900598526001)
--(axis cs:15,19.1517734527588)
--(axis cs:14,19.9990653991699)
--(axis cs:13,19.3770637512207)
--(axis cs:12,21.7922649383545)
--(axis cs:11,19.532320022583)
--(axis cs:10,17.9728126525879)
--(axis cs:9,19.9691982269287)
--(axis cs:8,19.7379283905029)
--(axis cs:7,19.0587768554688)
--(axis cs:6,19.2755126953125)
--(axis cs:5,17.1616382598877)
--(axis cs:4,16.8449935913086)
--(axis cs:3,16.1178207397461)
--(axis cs:2,15.4594411849976)
--(axis cs:1,16.7094135284424)
--(axis cs:0,21.6238594055176)
--cycle;

\addplot [semithick, blue, opacity=0.9]
table {%
0 11.8988428115845
1 10.2512111663818
2 10.8748111724854
3 11.5267820358276
4 10.9124794006348
5 11.172327041626
6 11.9973440170288
7 12.0509643554688
8 12.6201095581055
9 13.1663036346436
10 12.2637243270874
11 12.7578029632568
12 12.7646980285645
13 12.2792272567749
14 12.4225807189941
15 11.9488620758057
16 12.256495475769
17 12.0239810943604
18 12.3327255249023
19 11.7749061584473
20 11.9809999465942
21 11.9414281845093
22 13.1862478256226
23 12.7602310180664
24 13.8388252258301
25 13.3778638839722
26 12.7957725524902
27 14.0670728683472
28 13.801887512207
29 12.5061283111572
30 11.9493541717529
31 12.847692489624
32 12.4375429153442
33 11.9019289016724
34 12.2251071929932
35 10.9280366897583
36 11.9448585510254
37 11.5204381942749
38 12.1806497573853
39 12.9446887969971
40 13.1291904449463
41 13.7201614379883
42 14.4914426803589
43 13.8269701004028
44 12.5311670303345
45 14.0971984863281
46 13.5039644241333
47 12.9110021591187
48 12.3610067367554
49 11.3712282180786
50 12.2593326568604
51 12.2588777542114
52 11.7943162918091
53 13.7876834869385
54 12.5273571014404
55 13.2148532867432
56 13.5321998596191
57 13.7312288284302
58 12.9481229782104
59 14.0973634719849
60 13.2150220870972
61 12.7552070617676
62 13.1609220504761
63 14.0868530273438
64 13.2119312286377
65 11.6382741928101
66 11.6185092926025
67 11.7794961929321
68 12.529541015625
69 13.0682420730591
70 13.0396957397461
71 12.6858549118042
72 13.1263475418091
73 13.4018898010254
74 13.9774703979492
75 13.9815578460693
76 15.0281295776367
77 14.6449031829834
78 14.9796705245972
79 13.9764738082886
80 14.3231039047241
81 13.9386510848999
82 13.6923742294312
83 13.4383935928345
84 13.7557592391968
85 14.5759010314941
86 13.4119539260864
87 13.7526388168335
88 13.0563411712646
89 13.4654750823975
90 12.9806241989136
91 13.8718748092651
92 14.8417224884033
93 13.9511222839355
94 13.9287118911743
95 13.1198234558105
96 14.2755556106567
};
\addlegendentry{Opt. IMM}
\addplot [semithick, blue, opacity=0.9, dashed, forget plot]
table {%
0 12.9156036376953
1 12.9156036376953
2 12.9156036376953
3 12.9156036376953
4 12.9156036376953
5 12.9156036376953
6 12.9156036376953
7 12.9156036376953
8 12.9156036376953
9 12.9156036376953
10 12.9156036376953
11 12.9156036376953
12 12.9156036376953
13 12.9156036376953
14 12.9156036376953
15 12.9156036376953
16 12.9156036376953
17 12.9156036376953
18 12.9156036376953
19 12.9156036376953
20 12.9156036376953
21 12.9156036376953
22 12.9156036376953
23 12.9156036376953
24 12.9156036376953
25 12.9156036376953
26 12.9156036376953
27 12.9156036376953
28 12.9156036376953
29 12.9156036376953
30 12.9156036376953
31 12.9156036376953
32 12.9156036376953
33 12.9156036376953
34 12.9156036376953
35 12.9156036376953
36 12.9156036376953
37 12.9156036376953
38 12.9156036376953
39 12.9156036376953
40 12.9156036376953
41 12.9156036376953
42 12.9156036376953
43 12.9156036376953
44 12.9156036376953
45 12.9156036376953
46 12.9156036376953
47 12.9156036376953
48 12.9156036376953
49 12.9156036376953
50 12.9156036376953
51 12.9156036376953
52 12.9156036376953
53 12.9156036376953
54 12.9156036376953
55 12.9156036376953
56 12.9156036376953
57 12.9156036376953
58 12.9156036376953
59 12.9156036376953
60 12.9156036376953
61 12.9156036376953
62 12.9156036376953
63 12.9156036376953
64 12.9156036376953
65 12.9156036376953
66 12.9156036376953
67 12.9156036376953
68 12.9156036376953
69 12.9156036376953
70 12.9156036376953
71 12.9156036376953
72 12.9156036376953
73 12.9156036376953
74 12.9156036376953
75 12.9156036376953
76 12.9156036376953
77 12.9156036376953
78 12.9156036376953
79 12.9156036376953
80 12.9156036376953
81 12.9156036376953
82 12.9156036376953
83 12.9156036376953
84 12.9156036376953
85 12.9156036376953
86 12.9156036376953
87 12.9156036376953
88 12.9156036376953
89 12.9156036376953
90 12.9156036376953
91 12.9156036376953
92 12.9156036376953
93 12.9156036376953
94 12.9156036376953
95 12.9156036376953
96 12.9156036376953
};
\addplot [semithick, green, opacity=0.9]
table {%
0 12.0019111633301
1 10.4521884918213
2 11.7803077697754
3 13.641942024231
4 14.4062767028809
5 14.4524421691895
6 13.8661403656006
7 13.0104808807373
8 12.1300230026245
9 12.1569747924805
10 12.1691064834595
11 12.0678091049194
12 11.9853677749634
13 11.4691915512085
14 11.8204917907715
15 12.7087564468384
16 13.1432161331177
17 13.5122652053833
18 14.1726675033569
19 13.9594497680664
20 13.2069387435913
21 13.5370397567749
22 13.760290145874
23 14.02317237854
24 13.8405895233154
25 13.613883972168
26 13.3967199325562
27 13.1873416900635
28 12.8581104278564
29 12.424488067627
30 12.2005481719971
31 12.1175765991211
32 12.1650485992432
33 12.1550559997559
34 12.1264419555664
35 11.7356691360474
36 11.893533706665
37 12.3121204376221
38 12.8178968429565
39 13.1154174804688
40 13.3364162445068
41 13.7182350158691
42 14.4481801986694
43 15.2187929153442
44 14.7996101379395
45 14.2951345443726
46 13.3650703430176
47 13.2359533309937
48 13.3470554351807
49 12.9841423034668
50 12.8654870986938
51 12.8679714202881
52 12.5126256942749
53 12.9733543395996
54 13.339599609375
55 12.9461946487427
56 12.667820930481
57 12.7012300491333
58 12.6129903793335
59 12.6136722564697
60 12.9651660919189
61 12.9772453308105
62 12.7848176956177
63 13.2126903533936
64 13.5544319152832
65 13.4038772583008
66 12.9757871627808
67 12.5718860626221
68 12.3335695266724
69 12.7002077102661
70 13.1062631607056
71 12.8433103561401
72 12.4928369522095
73 12.8750734329224
74 13.1529712677002
75 13.1689319610596
76 14.2254581451416
77 14.187159538269
78 14.0684032440186
79 13.6283836364746
80 13.0830392837524
81 12.9464979171753
82 12.8499221801758
83 12.6307821273804
84 12.4844751358032
85 12.5517225265503
86 13.3044662475586
87 14.1590662002563
88 14.1819181442261
89 14.1741857528687
90 13.9233388900757
91 13.5948848724365
92 13.8018045425415
93 13.5981483459473
94 13.4533615112305
95 12.9937381744385
96 13.1901693344116
};
\addlegendentry{MKF}
\addplot [semithick, green, opacity=0.9, dashed, forget plot]
table {%
0 13.0762319564819
1 13.0762319564819
2 13.0762319564819
3 13.0762319564819
4 13.0762319564819
5 13.0762319564819
6 13.0762319564819
7 13.0762319564819
8 13.0762319564819
9 13.0762319564819
10 13.0762319564819
11 13.0762319564819
12 13.0762319564819
13 13.0762319564819
14 13.0762319564819
15 13.0762319564819
16 13.0762319564819
17 13.0762319564819
18 13.0762319564819
19 13.0762319564819
20 13.0762319564819
21 13.0762319564819
22 13.0762319564819
23 13.0762319564819
24 13.0762319564819
25 13.0762319564819
26 13.0762319564819
27 13.0762319564819
28 13.0762319564819
29 13.0762319564819
30 13.0762319564819
31 13.0762319564819
32 13.0762319564819
33 13.0762319564819
34 13.0762319564819
35 13.0762319564819
36 13.0762319564819
37 13.0762319564819
38 13.0762319564819
39 13.0762319564819
40 13.0762319564819
41 13.0762319564819
42 13.0762319564819
43 13.0762319564819
44 13.0762319564819
45 13.0762319564819
46 13.0762319564819
47 13.0762319564819
48 13.0762319564819
49 13.0762319564819
50 13.0762319564819
51 13.0762319564819
52 13.0762319564819
53 13.0762319564819
54 13.0762319564819
55 13.0762319564819
56 13.0762319564819
57 13.0762319564819
58 13.0762319564819
59 13.0762319564819
60 13.0762319564819
61 13.0762319564819
62 13.0762319564819
63 13.0762319564819
64 13.0762319564819
65 13.0762319564819
66 13.0762319564819
67 13.0762319564819
68 13.0762319564819
69 13.0762319564819
70 13.0762319564819
71 13.0762319564819
72 13.0762319564819
73 13.0762319564819
74 13.0762319564819
75 13.0762319564819
76 13.0762319564819
77 13.0762319564819
78 13.0762319564819
79 13.0762319564819
80 13.0762319564819
81 13.0762319564819
82 13.0762319564819
83 13.0762319564819
84 13.0762319564819
85 13.0762319564819
86 13.0762319564819
87 13.0762319564819
88 13.0762319564819
89 13.0762319564819
90 13.0762319564819
91 13.0762319564819
92 13.0762319564819
93 13.0762319564819
94 13.0762319564819
95 13.0762319564819
96 13.0762319564819
};
\addplot [semithick, purple, opacity=0.9]
table {%
0 6.73827266693115
1 6.87427234649658
2 6.44980382919312
3 9.02620983123779
4 9.78866767883301
5 10.4134511947632
6 12.8503532409668
7 13.3761787414551
8 13.9891586303711
9 13.5810565948486
10 15.4577875137329
11 14.8125352859497
12 15.1255321502686
13 14.825855255127
14 14.2525539398193
15 14.2027492523193
16 15.4405832290649
17 15.3501224517822
18 13.2394018173218
19 12.9319286346436
20 14.6190347671509
21 14.0948247909546
22 15.07994556427
23 13.3726482391357
24 13.6724948883057
25 13.6327362060547
26 13.9335966110229
27 13.430079460144
28 14.695333480835
29 14.1925430297852
30 14.6819086074829
31 15.6954278945923
32 11.9513349533081
33 11.43199634552
34 13.6720628738403
35 12.2103395462036
36 15.4128999710083
37 13.6914024353027
38 14.0312442779541
39 14.1849403381348
40 14.9666843414307
41 14.8803939819336
42 14.5181655883789
43 13.6853199005127
44 13.0877637863159
45 13.4053010940552
46 15.6076431274414
47 16.8577747344971
48 14.4704875946045
49 15.7331924438477
50 14.8074245452881
51 14.1517562866211
52 14.4514446258545
53 15.2484750747681
54 14.719708442688
55 14.7132740020752
56 13.6128711700439
57 14.4674215316772
58 16.1243343353271
59 16.1053447723389
60 14.6567411422729
61 16.2959957122803
62 16.4029960632324
63 16.2705516815186
64 15.6334323883057
65 15.3642072677612
66 16.3227005004883
67 15.0151224136353
68 15.0595302581787
69 14.0521306991577
70 15.6216611862183
71 15.2342147827148
72 14.9900236129761
73 15.8478097915649
74 14.7256050109863
75 15.3867101669312
76 15.963062286377
77 15.5119657516479
78 16.2568264007568
79 16.4317264556885
80 18.4969882965088
81 17.3163661956787
82 16.464183807373
83 16.4729976654053
84 16.5837097167969
85 17.4463634490967
86 16.2873401641846
87 14.0926523208618
88 13.0754537582397
89 13.2364540100098
90 14.3849687576294
91 14.7606801986694
92 15.1115818023682
93 15.7402582168579
94 15.0736865997314
95 14.6920871734619
96 15.2446231842041
};
\addlegendentry{GP}
\addplot [semithick, purple, opacity=0.9, dashed, forget plot]
table {%
0 14.4377660751343
1 14.4377660751343
2 14.4377660751343
3 14.4377660751343
4 14.4377660751343
5 14.4377660751343
6 14.4377660751343
7 14.4377660751343
8 14.4377660751343
9 14.4377660751343
10 14.4377660751343
11 14.4377660751343
12 14.4377660751343
13 14.4377660751343
14 14.4377660751343
15 14.4377660751343
16 14.4377660751343
17 14.4377660751343
18 14.4377660751343
19 14.4377660751343
20 14.4377660751343
21 14.4377660751343
22 14.4377660751343
23 14.4377660751343
24 14.4377660751343
25 14.4377660751343
26 14.4377660751343
27 14.4377660751343
28 14.4377660751343
29 14.4377660751343
30 14.4377660751343
31 14.4377660751343
32 14.4377660751343
33 14.4377660751343
34 14.4377660751343
35 14.4377660751343
36 14.4377660751343
37 14.4377660751343
38 14.4377660751343
39 14.4377660751343
40 14.4377660751343
41 14.4377660751343
42 14.4377660751343
43 14.4377660751343
44 14.4377660751343
45 14.4377660751343
46 14.4377660751343
47 14.4377660751343
48 14.4377660751343
49 14.4377660751343
50 14.4377660751343
51 14.4377660751343
52 14.4377660751343
53 14.4377660751343
54 14.4377660751343
55 14.4377660751343
56 14.4377660751343
57 14.4377660751343
58 14.4377660751343
59 14.4377660751343
60 14.4377660751343
61 14.4377660751343
62 14.4377660751343
63 14.4377660751343
64 14.4377660751343
65 14.4377660751343
66 14.4377660751343
67 14.4377660751343
68 14.4377660751343
69 14.4377660751343
70 14.4377660751343
71 14.4377660751343
72 14.4377660751343
73 14.4377660751343
74 14.4377660751343
75 14.4377660751343
76 14.4377660751343
77 14.4377660751343
78 14.4377660751343
79 14.4377660751343
80 14.4377660751343
81 14.4377660751343
82 14.4377660751343
83 14.4377660751343
84 14.4377660751343
85 14.4377660751343
86 14.4377660751343
87 14.4377660751343
88 14.4377660751343
89 14.4377660751343
90 14.4377660751343
91 14.4377660751343
92 14.4377660751343
93 14.4377660751343
94 14.4377660751343
95 14.4377660751343
96 14.4377660751343
};
\addplot [semithick, red, opacity=0.9]
table {%
0 11.855899810791
1 10.0241317749023
2 10.010066986084
3 11.3527355194092
4 11.7507266998291
5 11.9934768676758
6 13.0001945495605
7 13.2211437225342
8 13.3265409469604
9 13.8079652786255
10 12.506175994873
11 13.2313461303711
12 13.4200143814087
13 12.9764966964722
14 12.8714990615845
15 12.4109220504761
16 12.7900485992432
17 12.2526884078979
18 12.9419813156128
19 12.4462490081787
20 12.7444515228271
21 12.9209175109863
22 14.1577177047729
23 13.7000646591187
24 14.711597442627
25 13.6921329498291
26 12.9245738983154
27 14.2231616973877
28 14.689395904541
29 13.165566444397
30 12.6299133300781
31 13.2394409179688
32 12.8044967651367
33 12.0925331115723
34 12.6797132492065
35 11.3438529968262
36 12.7766237258911
37 12.2468795776367
38 13.1672554016113
39 14.1127319335938
40 14.2161512374878
41 14.4688730239868
42 15.2019968032837
43 14.366397857666
44 12.8959875106812
45 14.6989059448242
46 14.0484342575073
47 13.5642776489258
48 12.9370365142822
49 11.8006925582886
50 12.7169437408447
51 13.0845413208008
52 12.5516233444214
53 14.8556795120239
54 13.1326484680176
55 14.0770483016968
56 14.0300722122192
57 14.026180267334
58 13.3120384216309
59 14.6772155761719
60 13.6810312271118
61 12.8558435440063
62 13.928692817688
63 15.4999122619629
64 14.1307077407837
65 12.0245332717896
66 12.4371585845947
67 12.84250831604
68 13.481484413147
69 13.9673042297363
70 13.7467880249023
71 12.9783983230591
72 13.7356872558594
73 14.2777185440063
74 14.7726736068726
75 14.3591394424438
76 15.7666702270508
77 15.3786001205444
78 15.1905565261841
79 14.4273433685303
80 14.6304960250854
81 13.9926328659058
82 14.2295808792114
83 14.1843347549438
84 14.7148485183716
85 15.2602186203003
86 13.6740121841431
87 14.356653213501
88 13.1290340423584
89 13.5727729797363
90 13.3087205886841
91 14.246452331543
92 15.2607612609863
93 14.1554737091064
94 14.1535024642944
95 13.625714302063
96 15.255542755127
};
\addlegendentry{EKF}
\addplot [semithick, red, opacity=0.9, dashed, forget plot]
table {%
0 13.4856653213501
1 13.4856653213501
2 13.4856653213501
3 13.4856653213501
4 13.4856653213501
5 13.4856653213501
6 13.4856653213501
7 13.4856653213501
8 13.4856653213501
9 13.4856653213501
10 13.4856653213501
11 13.4856653213501
12 13.4856653213501
13 13.4856653213501
14 13.4856653213501
15 13.4856653213501
16 13.4856653213501
17 13.4856653213501
18 13.4856653213501
19 13.4856653213501
20 13.4856653213501
21 13.4856653213501
22 13.4856653213501
23 13.4856653213501
24 13.4856653213501
25 13.4856653213501
26 13.4856653213501
27 13.4856653213501
28 13.4856653213501
29 13.4856653213501
30 13.4856653213501
31 13.4856653213501
32 13.4856653213501
33 13.4856653213501
34 13.4856653213501
35 13.4856653213501
36 13.4856653213501
37 13.4856653213501
38 13.4856653213501
39 13.4856653213501
40 13.4856653213501
41 13.4856653213501
42 13.4856653213501
43 13.4856653213501
44 13.4856653213501
45 13.4856653213501
46 13.4856653213501
47 13.4856653213501
48 13.4856653213501
49 13.4856653213501
50 13.4856653213501
51 13.4856653213501
52 13.4856653213501
53 13.4856653213501
54 13.4856653213501
55 13.4856653213501
56 13.4856653213501
57 13.4856653213501
58 13.4856653213501
59 13.4856653213501
60 13.4856653213501
61 13.4856653213501
62 13.4856653213501
63 13.4856653213501
64 13.4856653213501
65 13.4856653213501
66 13.4856653213501
67 13.4856653213501
68 13.4856653213501
69 13.4856653213501
70 13.4856653213501
71 13.4856653213501
72 13.4856653213501
73 13.4856653213501
74 13.4856653213501
75 13.4856653213501
76 13.4856653213501
77 13.4856653213501
78 13.4856653213501
79 13.4856653213501
80 13.4856653213501
81 13.4856653213501
82 13.4856653213501
83 13.4856653213501
84 13.4856653213501
85 13.4856653213501
86 13.4856653213501
87 13.4856653213501
88 13.4856653213501
89 13.4856653213501
90 13.4856653213501
91 13.4856653213501
92 13.4856653213501
93 13.4856653213501
94 13.4856653213501
95 13.4856653213501
96 13.4856653213501
};
\addplot [semithick, black, opacity=0.7, mark=x, mark size=3, mark options={solid}]
table {%
0 13.1839437484741
1 15.9896545410156
2 13.867880821228
3 14.7092027664185
4 18.3605365753174
5 16.7674674987793
6 14.9182348251343
7 15.6318788528442
8 14.7217264175415
9 13.9438953399658
10 15.9824094772339
11 13.6165084838867
12 14.4943180084229
13 16.8203601837158
14 14.6307325363159
15 15.4902696609497
16 11.9295539855957
17 13.7931203842163
18 13.823016166687
19 13.9921407699585
20 12.3312997817993
21 15.8135433197021
22 15.0151720046997
23 14.5668210983276
24 14.5276746749878
25 14.4882068634033
26 13.9100856781006
27 13.2583751678467
28 14.9425535202026
29 16.2055206298828
30 11.6449089050293
31 12.092755317688
32 12.5512266159058
33 15.8596935272217
34 10.9728813171387
35 14.5666246414185
36 12.2668590545654
37 14.4842624664307
38 13.5005302429199
39 16.1150665283203
40 17.1802043914795
41 17.0659122467041
42 12.205753326416
43 13.8765392303467
44 14.4287757873535
45 12.3787508010864
46 16.1644973754883
47 13.9942684173584
48 14.7310085296631
49 12.2735872268677
50 12.9925403594971
51 13.5361766815186
52 13.6683130264282
53 16.1655807495117
54 18.0310897827148
55 13.4524793624878
56 16.8783874511719
57 15.0577869415283
58 14.792896270752
59 12.2660074234009
60 16.1155128479004
61 13.4383249282837
62 13.0288925170898
63 13.8526487350464
64 16.8131504058838
65 15.3843946456909
66 14.6525182723999
67 14.1252355575562
68 13.8502502441406
69 13.0684471130371
70 14.0781650543213
71 11.9355783462524
72 16.9747409820557
73 14.6308813095093
74 14.2783803939819
75 13.5612173080444
76 13.3538999557495
77 15.2429838180542
78 14.3507823944092
79 12.8196907043457
80 15.6162195205688
81 13.438943862915
82 12.432258605957
83 17.1009273529053
84 13.4039220809937
85 15.3619413375854
86 14.9507102966309
87 14.3290462493896
88 13.2657222747803
89 13.8780870437622
90 14.7275772094727
91 15.1292171478271
92 15.9924869537354
93 15.9047574996948
94 14.1250352859497
95 12.7294731140137
96 15.6693391799927
};
\addlegendentry{Measurements}
\addplot [semithick, black, opacity=0.7, dashed, forget plot]
table {%
0 14.435133934021
1 14.435133934021
2 14.435133934021
3 14.435133934021
4 14.435133934021
5 14.435133934021
6 14.435133934021
7 14.435133934021
8 14.435133934021
9 14.435133934021
10 14.435133934021
11 14.435133934021
12 14.435133934021
13 14.435133934021
14 14.435133934021
15 14.435133934021
16 14.435133934021
17 14.435133934021
18 14.435133934021
19 14.435133934021
20 14.435133934021
21 14.435133934021
22 14.435133934021
23 14.435133934021
24 14.435133934021
25 14.435133934021
26 14.435133934021
27 14.435133934021
28 14.435133934021
29 14.435133934021
30 14.435133934021
31 14.435133934021
32 14.435133934021
33 14.435133934021
34 14.435133934021
35 14.435133934021
36 14.435133934021
37 14.435133934021
38 14.435133934021
39 14.435133934021
40 14.435133934021
41 14.435133934021
42 14.435133934021
43 14.435133934021
44 14.435133934021
45 14.435133934021
46 14.435133934021
47 14.435133934021
48 14.435133934021
49 14.435133934021
50 14.435133934021
51 14.435133934021
52 14.435133934021
53 14.435133934021
54 14.435133934021
55 14.435133934021
56 14.435133934021
57 14.435133934021
58 14.435133934021
59 14.435133934021
60 14.435133934021
61 14.435133934021
62 14.435133934021
63 14.435133934021
64 14.435133934021
65 14.435133934021
66 14.435133934021
67 14.435133934021
68 14.435133934021
69 14.435133934021
70 14.435133934021
71 14.435133934021
72 14.435133934021
73 14.435133934021
74 14.435133934021
75 14.435133934021
76 14.435133934021
77 14.435133934021
78 14.435133934021
79 14.435133934021
80 14.435133934021
81 14.435133934021
82 14.435133934021
83 14.435133934021
84 14.435133934021
85 14.435133934021
86 14.435133934021
87 14.435133934021
88 14.435133934021
89 14.435133934021
90 14.435133934021
91 14.435133934021
92 14.435133934021
93 14.435133934021
94 14.435133934021
95 14.435133934021
96 14.435133934021
};
\end{axis}

\end{tikzpicture}

%% file: figures/drone_trajectory_sample5_test.tikz
\begin{tikzpicture}

\definecolor{darkgray176}{RGB}{176,176,176}
\definecolor{green}{RGB}{0,128,0}
\definecolor{lightgray204}{RGB}{204,204,204}
\definecolor{purple}{RGB}{128,0,128}

\begin{axis}[
	width=0.9\linewidth,
	height=0.8\linewidth,
	legend cell align={left},
	tick align=outside,
	tick pos=left,
	x grid style={darkgray176},
	xlabel={\footnotesize Rel. East Position [km]},
	xmin=-76.4565835952759, xmax=-53.6275342941284,
	xtick style={color=black},
	y grid style={darkgray176},
	ylabel={\footnotesize Rel. North Position [km]},
	ymin=-46.9177104949951, ymax=14.8000423431396,
	ytick style={color=black},
	legend style={at={(0.03,0.97)},anchor=north west,legend cell align=left,draw=white!15!black,fill opacity=0.5,draw opacity=1,text opacity=1,font=\scriptsize}
]
\addplot [draw=green, fill=green, mark=x, only marks, opacity=0.5]
table{%
x  y
-72.7561416625977 -14.1470031738281
-73.0429077148438 -15.088306427002
-74.6092987060547 -22.4831485748291
-74.296630859375 -19.3793716430664
-75.4188995361328 -24.4189319610596
-73.7339630126953 -20.4089775085449
-74.5842971801758 -25.1789207458496
-73.2356262207031 -25.5031795501709
-72.9073715209961 -26.0135917663574
-72.875358581543 -26.8740139007568
-72.2694854736328 -26.7636470794678
-70.6387710571289 -29.0469551086426
-72.0517120361328 -33.5064926147461
-71.4840698242188 -31.2054042816162
-69.9866104125977 -30.2672863006592
-68.3456192016602 -29.2030048370361
-68.5008697509766 -35.8480224609375
-67.4082489013672 -35.2069702148438
-67.9065856933594 -36.3354148864746
-67.294807434082 -35.2045478820801
-65.9630279541016 -38.403938293457
-65.740234375 -36.4710693359375
-65.0595245361328 -36.1580085754395
-63.6734924316406 -37.7019424438477
-61.9761581420898 -38.389835357666
-62.7665710449219 -42.5408401489258
-62.5640335083008 -38.2113914489746
-60.5050735473633 -37.7403411865234
-60.554931640625 -39.722225189209
-60.772216796875 -44.1123580932617
-61.4237594604492 -43.0018920898438
-58.6251754760742 -39.4049758911133
-59.5200805664062 -40.6113357543945
-58.0342636108398 -38.004638671875
-58.2473602294922 -38.0185508728027
-56.2810592651367 -37.2261695861816
-57.9272842407227 -36.4763145446777
-56.7131423950195 -35.664363861084
-58.4343338012695 -39.3670196533203
-56.4401168823242 -36.9466934204102
-56.5543060302734 -35.5349655151367
-56.1527404785156 -36.9189224243164
-56.3438568115234 -35.576774597168
-55.6691741943359 -39.1451187133789
-56.5774230957031 -36.1809768676758
-54.9209671020508 -30.6217212677002
-54.9214324951172 -32.4474487304688
-56.1057739257812 -33.0882148742676
-56.8143539428711 -37.9417343139648
-56.8524856567383 -31.6661796569824
-55.8102798461914 -30.7088851928711
-56.8504791259766 -32.8315811157227
-55.1495132446289 -28.5167236328125
-55.5400390625 -29.3972434997559
-55.2903900146484 -29.2383193969727
-57.0254440307617 -31.5030632019043
-56.6632385253906 -29.7824535369873
-56.7900619506836 -30.0300559997559
-58.8639144897461 -30.5748291015625
-58.8190002441406 -27.375
-59.1912612915039 -29.7122554779053
-58.4747009277344 -28.635082244873
-57.1801910400391 -25.7595367431641
-59.1691665649414 -27.362964630127
-60.8012237548828 -27.1829414367676
-59.8067474365234 -26.2017688751221
-60.8912887573242 -25.5029907226562
-60.3856811523438 -23.416015625
-59.4256286621094 -20.7714691162109
-60.3983306884766 -24.3038425445557
-62.1292343139648 -23.9797134399414
-59.4300079345703 -18.5848236083984
-60.6740188598633 -17.9452323913574
-61.8202133178711 -19.9109687805176
-60.4882888793945 -15.781177520752
-61.3146286010742 -13.0037651062012
-60.2061462402344 -15.6238975524902
-59.7993621826172 -12.3333358764648
-62.3864288330078 -14.5932502746582
-60.5336608886719 -9.82938766479492
-63.0410537719727 -12.1091384887695
-62.6350708007812 -12.9919242858887
-62.7525100708008 -7.42779541015625
-61.9749450683594 -9.23740386962891
-61.8883361816406 -5.57217025756836
-63.7571868896484 -6.05472183227539
-63.6167831420898 -5.25815200805664
-61.9436416625977 -0.748516082763672
-61.8400650024414 -0.958751678466797
-60.6391754150391 0.658515930175781
-63.4032287597656 -1.97574615478516
-61.0917587280273 2.37361526489258
-61.0761489868164 4.65365219116211
-59.9574813842773 4.01910400390625
-58.7867889404297 6.58717346191406
-60.3125 2.01164627075195
-59.4106826782227 7.52653121948242
-57.6988983154297 9.41825485229492
-56.5693206787109 11.9946899414062
};
\addlegendentry{Measurements}
\addplot [semithick, black, opacity=0.9]
table {%
-73.6245727539062 -15.7358198165894
-73.8234329223633 -16.9946670532227
-73.9086151123047 -18.2597980499268
-73.9593276977539 -19.5405807495117
-73.9066772460938 -20.7958698272705
-73.8131942749023 -22.0659828186035
-73.6211547851562 -23.2835826873779
-73.3943862915039 -24.5158615112305
-73.0834732055664 -25.6465301513672
-72.7270126342773 -26.7834568023682
-72.3045654296875 -27.8320331573486
-71.8430099487305 -28.9274215698242
-71.3392944335938 -29.9237270355225
-70.7897033691406 -30.937442779541
-70.2181167602539 -31.8365936279297
-69.5964965820312 -32.7463264465332
-68.9738998413086 -33.5466384887695
-68.2942276000977 -34.3501739501953
-67.6338653564453 -35.0341262817383
-66.9170761108398 -35.7117538452148
-66.2527389526367 -36.2598571777344
-65.5521087646484 -36.7792587280273
-64.8894653320312 -37.2118453979492
-64.1366500854492 -37.6663780212402
-63.4449996948242 -38.0355110168457
-62.7141456604004 -38.3626022338867
-62.0617790222168 -38.6033706665039
-61.3711929321289 -38.7926254272461
-60.7621421813965 -38.9022064208984
-60.1144981384277 -38.9508934020996
-59.5831298828125 -38.9133415222168
-59.0365905761719 -38.8118286132812
-58.6276245117188 -38.6449699401855
-58.2023429870605 -38.425350189209
-57.9033432006836 -38.1718444824219
-57.5697898864746 -37.8762512207031
-57.3449172973633 -37.5737075805664
-57.086784362793 -37.2275581359863
-56.9194145202637 -36.9070701599121
-56.7260665893555 -36.5385971069336
-56.6020736694336 -36.2144775390625
-56.4666175842285 -35.839241027832
-56.3830451965332 -35.5201797485352
-56.2961959838867 -35.139331817627
-56.2523155212402 -34.8222923278809
-56.2195510864258 -34.4424858093262
-56.2172737121582 -34.1208992004395
-56.2343635559082 -33.7343559265137
-56.2766876220703 -33.4011497497559
-56.3424491882324 -33.0098838806152
-56.4247970581055 -32.662971496582
-56.5319137573242 -32.2582130432129
-56.6496887207031 -31.8811874389648
-56.7949714660645 -31.4525089263916
-56.9475440979004 -31.0405750274658
-57.1313514709473 -30.5844612121582
-57.3183288574219 -30.1444053649902
-57.5319404602051 -29.6720581054688
-57.7436294555664 -29.2019195556641
-57.9821586608887 -28.6953716278076
-58.2165718078613 -28.1785297393799
-58.479118347168 -27.6287460327148
-58.7348747253418 -27.0533218383789
-59.0126838684082 -26.4475574493408
-59.2768249511719 -25.8151073455811
-59.5554122924805 -25.1564273834229
-59.8085136413574 -24.4702758789062
-60.0669784545898 -23.7609920501709
-60.298828125 -23.0010643005371
-60.5392036437988 -22.20973777771
-60.7472686767578 -21.3819808959961
-60.9584808349609 -20.5330390930176
-61.1431198120117 -19.6267337799072
-61.3348999023438 -18.6977787017822
-61.4964904785156 -17.7371864318848
-61.6609077453613 -16.7631950378418
-61.8058395385742 -15.7494783401489
-61.9556503295898 -14.7172145843506
-62.0930519104004 -13.6292896270752
-62.226734161377 -12.5231151580811
-62.33740234375 -11.3806810379028
-62.4341583251953 -10.2315187454224
-62.5016326904297 -9.06919097900391
-62.5467910766602 -7.89506721496582
-62.5462265014648 -6.68537378311157
-62.5130386352539 -5.44865703582764
-62.4063835144043 -4.19541311264038
-62.2516746520996 -2.93096137046814
-62.0082740783691 -1.66943955421448
-61.7258071899414 -0.407344818115234
-61.3527374267578 0.843922913074493
-60.9439544677734 2.09688258171082
-60.4492988586426 3.32713389396667
-59.9231986999512 4.54996585845947
-59.307502746582 5.75236749649048
-58.6604385375977 6.94279670715332
-57.942195892334 8.0724983215332
-57.1939239501953 9.17639636993408
-56.3700942993164 10.2385950088501
};
\addlegendentry{True Trajectory}
\addplot [semithick, blue, opacity=0.9]
table {%
-73.0317459106445 -15.0437450408936
-74.2574234008789 -20.5804290771484
-74.5893936157227 -20.8277606964111
-75.3064346313477 -23.7886791229248
-74.6681518554688 -22.7583084106445
-74.6979217529297 -24.5843658447266
-74.0260314941406 -25.6646480560303
-73.4510345458984 -26.5166168212891
-73.0775680541992 -27.3772869110107
-72.6093444824219 -27.865894317627
-71.6164245605469 -28.9014892578125
-71.5017318725586 -31.1006832122803
-71.2793884277344 -31.9937038421631
-70.6042022705078 -32.25341796875
-69.5470962524414 -32.0066909790039
-68.769401550293 -33.739315032959
-67.8375778198242 -34.8284950256348
-67.4477386474609 -35.9946899414062
-67.0419006347656 -36.4993133544922
-66.2668380737305 -37.7071113586426
-65.7174911499023 -38.0427627563477
-65.1440582275391 -38.1068344116211
-64.2360305786133 -38.5053062438965
-62.9699974060059 -38.9433670043945
-62.399845123291 -40.4759330749512
-62.0510559082031 -40.4089431762695
-61.1058082580566 -40.0705032348633
-60.4707412719727 -40.303165435791
-60.1622314453125 -41.7394638061523
-60.282772064209 -42.5754432678223
-59.4070205688477 -42.0708389282227
-59.1419486999512 -41.9745559692383
-58.4919090270996 -41.040958404541
-58.1263656616211 -40.2059059143066
-57.1822204589844 -39.2446632385254
-57.1292839050293 -38.2071647644043
-56.7268409729004 -37.1496391296387
-57.0667495727539 -37.4197158813477
-56.6680526733398 -37.0505828857422
-56.4628410339355 -36.3576927185059
-56.1623229980469 -36.247673034668
-56.0559272766113 -35.8000831604004
-55.7326736450195 -36.5123634338379
-55.8695030212402 -36.2711143493652
-55.5547828674316 -34.5105514526367
-55.2403831481934 -33.5868988037109
-55.4053382873535 -32.9971160888672
-55.7814140319824 -33.9936866760254
-56.1893348693848 -33.1008148193359
-56.1367568969727 -32.1264190673828
-56.4058723449707 -31.9636058807373
-56.0826988220215 -30.6702461242676
-55.9060478210449 -29.9012775421143
-55.6648139953613 -29.2877464294434
-56.0581016540527 -29.444507598877
-56.2730560302734 -29.181999206543
-56.4766540527344 -29.0869998931885
-57.3614311218262 -29.1916275024414
-58.1000328063965 -28.4970722198486
-58.710376739502 -28.5363731384277
-58.8702545166016 -28.3286552429199
-58.5392379760742 -27.4370193481445
-58.8948974609375 -27.1278419494629
-59.7405624389648 -26.8329544067383
-60.0132179260254 -26.4024868011475
-60.5788803100586 -25.8780784606934
-60.8407096862793 -24.9542694091797
-60.7318077087402 -23.5469341278076
-60.7683601379395 -23.3276042938232
-61.3661308288574 -23.0672836303711
-61.0335235595703 -21.5386714935303
-61.1359710693359 -20.1040935516357
-61.4716606140137 -19.4649295806885
-61.3872337341309 -17.9466152191162
-61.6251449584961 -15.9487323760986
-61.2891807556152 -15.2414865493774
-60.9253768920898 -13.804461479187
-61.3530464172363 -13.1951541900635
-61.2371101379395 -11.6002788543701
-61.8157844543457 -10.8561544418335
-62.0823822021484 -10.6509799957275
-62.5193061828613 -9.10502147674561
-62.4501113891602 -8.45467948913574
-62.4623718261719 -7.01826095581055
-62.9715003967285 -5.93416452407837
-63.311393737793 -4.96364974975586
-63.2013969421387 -3.15313482284546
-62.9618377685547 -1.82594847679138
-62.3794212341309 -0.453357815742493
-62.5775489807129 0.0357479453086853
-62.1586265563965 1.38316607475281
-61.8769493103027 3.03159809112549
-61.2083396911621 4.04513740539551
-60.3802871704102 5.44971513748169
-59.9463119506836 5.30671310424805
-59.5844459533691 6.61230325698853
-58.8451805114746 8.04296112060547
};
\addlegendentry{Opt. IMM}
\addplot [semithick, green, opacity=0.9]
table {%
-73.0155715942383 -14.9805307388306
-73.9785308837891 -19.6946392059326
-74.0005187988281 -18.527250289917
-75.0032043457031 -22.5200023651123
-73.8629302978516 -20.9408836364746
-74.3596878051758 -23.7936992645264
-73.3086242675781 -25.1203956604004
-72.7510986328125 -25.2793979644775
-73.167236328125 -27.4617958068848
-72.3741226196289 -26.5075054168701
-70.7837448120117 -29.0249767303467
-71.8032989501953 -31.6051979064941
-71.9794464111328 -32.532039642334
-70.9703140258789 -32.5088233947754
-69.2296752929688 -31.1739883422852
-68.7417526245117 -35.8987503051758
-67.2499160766602 -34.2239112854004
-67.5811004638672 -34.9733657836914
-67.1795806884766 -35.0251770019531
-65.7330932617188 -36.5429229736328
-65.589729309082 -35.8998603820801
-65.0936660766602 -36.6635398864746
-63.9373474121094 -38.5731163024902
-62.2928810119629 -38.7766723632812
-62.7181015014648 -41.6808471679688
-62.520938873291 -38.6354866027832
-60.5096664428711 -37.7531318664551
-60.3313255310059 -39.1279258728027
-60.8688926696777 -43.9100379943848
-61.2259712219238 -42.7623634338379
-59.0430183410645 -40.6929168701172
-59.5604057312012 -40.616283416748
-58.0675659179688 -38.3467102050781
-58.1483955383301 -38.0782814025879
-56.4561576843262 -37.260425567627
-57.6767768859863 -35.9093933105469
-56.699291229248 -35.4906272888184
-58.2504005432129 -37.6342506408691
-56.8133697509766 -37.4827537536621
-56.2106552124023 -34.6604194641113
-56.0920333862305 -36.5408935546875
-56.6044540405273 -35.9634017944336
-55.656810760498 -38.7509956359863
-56.7842445373535 -37.0421943664551
-55.2076072692871 -31.6869354248047
-54.6652183532715 -32.1043090820312
-55.8427619934082 -30.8113231658936
-56.847354888916 -36.8762397766113
-56.7258987426758 -32.1780319213867
-55.9789085388184 -31.3674144744873
-56.6725120544434 -31.6539154052734
-55.4047927856445 -28.6054630279541
-55.6401290893555 -29.3722133636475
-55.3093338012695 -29.2073745727539
-57.0745735168457 -31.3743152618408
-56.4254264831543 -29.3907051086426
-56.6432189941406 -29.8381481170654
-58.8016166687012 -30.7059745788574
-58.9569473266602 -28.1119785308838
-59.0891265869141 -29.5388126373291
-58.5658721923828 -28.9212512969971
-57.2180252075195 -25.8810539245605
-59.114990234375 -27.0418014526367
-60.4667320251465 -26.2896919250488
-59.8692779541016 -26.2263927459717
-60.5832710266113 -25.1559963226318
-60.178840637207 -23.4563465118408
-59.7541198730469 -21.5336017608643
-60.9803009033203 -25.2215995788574
-62.0772171020508 -23.7406883239746
-60.5282287597656 -20.8341827392578
-61.6382064819336 -20.1537742614746
-61.7973670959473 -19.8330364227295
-60.8991661071777 -16.6079902648926
-62.1647758483887 -14.6189794540405
-60.9690971374512 -16.7881507873535
-60.1895408630371 -13.0601329803467
-61.8472480773926 -13.3339929580688
-60.386646270752 -9.62549781799316
-62.4272117614746 -10.6398506164551
-62.5640182495117 -12.3353939056396
-62.9982643127441 -7.88075494766235
-61.9032096862793 -8.96553993225098
-61.820987701416 -5.48167943954468
-63.5810661315918 -5.8921046257019
-63.8176231384277 -5.48355531692505
-62.725025177002 -2.07144570350647
-62.453742980957 -1.97663044929504
-61.0394439697266 0.030032753944397
-63.1806640625 -1.6854008436203
-61.5009918212891 1.79150092601776
-61.2813377380371 4.37644481658936
-60.0629806518555 4.06301784515381
-58.7587089538574 6.79885959625244
-59.7541923522949 3.12502932548523
-60.3635482788086 6.14040088653564
-57.8831253051758 9.29314231872559
};
\addlegendentry{MKF}
\addplot [semithick, red, opacity=0.9]
table {%
-73.0327987670898 -15.0466690063477
-74.6254730224609 -21.7167282104492
-74.8794250488281 -20.3797912597656
-75.3967208862305 -23.9220390319824
-74.1432876586914 -21.1762084960938
-74.5368347167969 -24.4039001464844
-73.3230743408203 -25.6826877593994
-72.9113998413086 -26.1402816772461
-72.865608215332 -26.8450489044189
-72.3053283691406 -26.8601303100586
-70.6486206054688 -28.8528385162354
-71.9481811523438 -33.0589904785156
-71.7691116333008 -31.8281326293945
-70.0539855957031 -30.387321472168
-68.3364868164062 -29.1841354370117
-68.6127853393555 -34.8655891418457
-67.5956954956055 -35.6926002502441
-67.8804016113281 -36.3876914978027
-67.366943359375 -35.415828704834
-66.0074691772461 -38.0415687561035
-65.817985534668 -36.9359130859375
-65.0845489501953 -36.1619873046875
-63.6867446899414 -37.4819374084473
-61.9831199645996 -38.4171257019043
-62.7265472412109 -42.1197242736816
-62.9201965332031 -38.9538841247559
-60.5730400085449 -37.6658592224121
-60.5057373046875 -39.2610778808594
-60.8155403137207 -43.7409973144531
-61.5608558654785 -43.5560340881348
-58.8603401184082 -39.7156677246094
-59.4675979614258 -39.9412651062012
-58.1392211914062 -38.1589469909668
-58.1932716369629 -37.8264389038086
-56.3426475524902 -37.2528953552246
-57.824348449707 -36.4449310302734
-56.7996520996094 -35.6811866760254
-58.4022445678711 -38.778564453125
-56.6995086669922 -37.2500190734863
-56.5055770874023 -35.5564193725586
-56.1777534484863 -36.650074005127
-56.3720970153809 -35.7844772338867
-55.7472496032715 -38.8128814697266
-56.7462730407715 -36.8126754760742
-55.1959533691406 -30.9723739624023
-55.0211448669434 -31.7067699432373
-56.0618362426758 -32.9541397094727
-56.8597450256348 -37.5904998779297
-57.5715370178223 -32.7335205078125
-55.8857803344727 -30.6094379425049
-56.7923851013184 -32.2339782714844
-55.4466934204102 -28.9203472137451
-55.490650177002 -29.0004901885986
-55.294319152832 -29.2072792053223
-56.9550285339355 -31.227840423584
-56.8370666503906 -30.0763473510742
-56.7724647521973 -29.9425945281982
-58.7878227233887 -30.4303131103516
-59.0124320983887 -27.6991596221924
-59.2049331665039 -29.2891654968262
-58.5509643554688 -28.8306751251221
-57.2989501953125 -26.0292167663574
-59.063404083252 -26.7898139953613
-60.8100395202637 -27.1892681121826
-59.9288139343262 -26.3511257171631
-60.825122833252 -25.4697704315186
-60.4713325500488 -23.5461368560791
-59.4740295410156 -20.8797740936279
-60.3701820373535 -23.6039505004883
-62.1593170166016 -24.1819763183594
-59.9483909606934 -19.0979881286621
-60.5041389465332 -17.4185180664062
-61.7749099731445 -19.4316444396973
-60.8674507141113 -16.3256988525391
-61.2876739501953 -13.0266408920288
-60.2510032653809 -15.2036991119385
-60.0138778686523 -12.854621887207
-62.2515907287598 -14.0677471160889
-60.9883460998535 -10.330135345459
-62.8498611450195 -11.3921966552734
-62.6785545349121 -12.9391679763794
-63.218090057373 -8.24220275878906
-62.014965057373 -8.91477394104004
-62.013542175293 -5.94947576522827
-63.625659942627 -5.67598724365234
-63.675968170166 -5.29599523544312
-62.2829246520996 -1.21808242797852
-61.7202987670898 -0.605659246444702
-60.6689834594727 0.609089970588684
-63.1288642883301 -1.34251952171326
-61.6551208496094 1.82666158676147
-61.0490188598633 4.6610689163208
-59.8878326416016 4.29281187057495
-58.9032554626465 6.35118865966797
-60.1193656921387 2.72215747833252
-60.1496429443359 6.64427423477173
-57.8654556274414 9.34117412567139
};
\addlegendentry{EKF}
\addplot [semithick, purple, opacity=0.9]
table {%
-73.2590787077534 -15.4996578286922
-72.9638596756174 -15.9694692695795
-73.7524670345725 -20.100397483492
-74.2203909902519 -19.3180158366817
-75.1675338610188 -23.4387965502716
-73.958138111897 -20.8979524612371
-73.914567293668 -23.192768240495
-73.1176757948413 -24.8510559014701
-72.8232366668237 -25.8348085716351
-72.864461992902 -27.0500813628886
-72.1733028806586 -26.5187162781794
-70.5120748370692 -28.0275715571188
-72.0280863594157 -32.9987223764454
-71.5754133592519 -32.0072104883082
-70.0582026986014 -30.8204719876618
-68.7006086724736 -29.7241644938377
-68.11177573381 -35.0636713329688
-67.6073603798093 -35.5173326409747
-68.1437833950763 -36.3908707660423
-67.392982354533 -35.3591689778419
-65.6011879980115 -37.0921985738724
-65.9680974860406 -36.6932039126093
-65.1277060233697 -36.1333075191214
-63.7515432364042 -36.8122355589761
-62.8893529116767 -38.7288743580511
-62.604384232043 -40.7321961661716
-62.2441517766176 -38.8915869938972
-60.4640920565873 -37.8317577610065
-60.5262423019312 -39.4239384253514
-60.8897649362641 -43.9404488613417
-61.3810970941822 -40.5434068284846
-58.7639519528365 -39.2959703103181
-59.4182719181227 -40.4502214471725
-57.9526237452252 -37.465014947534
-58.1247646155974 -37.7031028187427
-56.7871105819691 -37.4307232812514
-58.0003771970313 -36.5347610773836
-56.8179596245067 -36.1123507650134
-58.1058608066735 -38.730890581415
-56.5991733436078 -36.9416832814532
-56.9176357086593 -36.0808833471679
-55.5601780515819 -35.3607267515885
-56.0310256732832 -35.2606156967122
-55.5772706641842 -38.2527283287864
-56.5510141128025 -36.0793924242268
-54.8239561893556 -31.7867268839951
-54.9403551137657 -32.8553587623383
-56.020716313899 -33.386105277049
-56.6391501747646 -37.0418958318078
-56.6612614001993 -31.349440682472
-55.9071721255406 -30.7447990836163
-56.6960740541874 -33.7256786785667
-55.3773412769869 -30.936815529661
-55.7562692194559 -30.3167918021651
-55.2807465836071 -29.7315196155553
-56.9848342180847 -31.1141124499727
-56.4288053686158 -29.7699203157334
-56.8733887855265 -29.9680777277559
-58.2922700901842 -28.8069846127159
-58.7741568350691 -27.8222383466475
-58.8840007634613 -28.9721941778043
-58.5669712504632 -28.8318459964187
-57.3378298692367 -26.0549340665395
-58.9583319113251 -27.0514109870061
-60.720590155361 -27.3521042014851
-59.4892272404614 -26.1594120246814
-60.9089403885944 -25.3531630624759
-60.0861483458057 -23.2796430929145
-59.505825619735 -20.8604407127426
-59.9818965524765 -22.6426756358495
-62.0200854837527 -23.6142594080444
-60.1182830151804 -19.4644550450963
-59.8811090561664 -16.3237982096815
-61.9209893271838 -19.6757247122586
-60.9652854123681 -17.1932986014624
-61.5330081479137 -13.5782092791499
-60.1095017316932 -15.0196169270053
-60.6011787902673 -14.4331610076357
-62.1319972714812 -13.6652867069545
-60.5151937076457 -10.4067420481503
-62.3352936300947 -11.0314674121946
-62.3138352392993 -12.2196245502775
-62.6139881097334 -6.49722440537136
-61.6877314104437 -8.85325289740728
-61.8915235945144 -5.90488911276057
-63.218540603892 -5.11507041788713
-62.5486884980004 -4.03188197560702
-62.5915772548438 -1.73463693951968
-61.3079796825789 -0.495391716643258
-60.430670378379 0.708232268861206
-63.0270923405729 -1.14917375888326
-60.9556271782117 1.82465825260722
-61.4693656693617 3.43559455427539
-60.1390109140265 2.85902446892624
-58.7528153003791 6.47697921420705
-59.1672104996679 3.74139293429332
-59.0872452525467 8.04815185026322
-57.8280365955328 9.05094736469492
-57.0218570474696 11.3680422572647
-56.6811589016666 9.34973812744008
};
\addlegendentry{GP}
\end{axis}

\end{tikzpicture}

%% file: figures/rib_trajectory_sample72_test.tikz
\begin{tikzpicture}

\definecolor{darkgray176}{RGB}{176,176,176}
\definecolor{green}{RGB}{0,128,0}
\definecolor{lightgray204}{RGB}{204,204,204}
\definecolor{purple}{RGB}{128,0,128}

\begin{axis}[
	width=0.9\linewidth,
	height=0.8\linewidth,
	legend cell align={left},
	tick align=outside,
	tick pos=left,
	x grid style={darkgray176},
	xlabel={\footnotesize Rel. East Position [km]},
	xmin=8317.18896454963, xmax=9093.0623689755,
	xtick style={color=black},
	xtick={8200,8300,8400,8500,8600,8700,8800,8900,9000,9100,9200},
	xticklabels={8.2,8.3,8.4,8.5,8.6,8.7,8.8,8.9,9,9.1,9.2},
	y grid style={darkgray176},
	ylabel={\footnotesize Rel. North Position [km]},
	ymin=3705.97428119337, ymax=4322.4570468777,
	ytick style={color=black},
	ytick={3700,3800,3900,4000,4100,4200,4300,4400},
	yticklabels={3.7,3.8,3.9,4,4.1,4.2,4.3,4.4},
	legend style={at={(0.03,0.03)},anchor=south west,legend cell align=left,draw=white!15!black,fill opacity=0.5,draw opacity=1,text opacity=1,font=\scriptsize} 
]
\addplot [draw=green, fill=green, mark=x, only marks, opacity=0.5]
table{%
x  y
8498.5986328125 4206.6767578125
8498.94140625 4210.8369140625
8443.9462890625 4198.498046875
8446.697265625 4204.3056640625
8416.2109375 4198.77734375
8461.95703125 4221.30078125
8552.44921875 4258.6416015625
8465.7216796875 4229.537109375
8421.4658203125 4215.796875
8448.080078125 4225.56640625
8407.5078125 4210.0322265625
8516.3369140625 4246.080078125
8501.234375 4237.8330078125
8479.2919921875 4225.642578125
8512.0576171875 4232.859375
8534.005859375 4235.6279296875
8582.2353515625 4246.6279296875
8584.6259765625 4243.3544921875
8483.322265625 4203.529296875
8508.3603515625 4207.7978515625
8606.0634765625 4237.462890625
8609.2099609375 4235.5380859375
8554.8955078125 4214.9150390625
8683.1787109375 4257.6689453125
8569.0185546875 4216.3134765625
8679.0556640625 4251.953125
8627.6240234375 4234.0380859375
8790.8876953125 4284.8896484375
8753.2587890625 4269.4013671875
8752.080078125 4267.7431640625
8717.306640625 4254.6044921875
8691.6484375 4243.306640625
8690.2412109375 4239.923828125
8735.4326171875 4253.37109375
8821.537109375 4280.001953125
8711.0908203125 4243.4287109375
8805.2548828125 4271.076171875
8725.916015625 4245.34375
8819.1962890625 4274.7421875
8804.7607421875 4268.3623046875
8844.4169921875 4280.27734375
8792.4443359375 4263.560546875
8796.2705078125 4262.0859375
8764.1484375 4248.8447265625
8792.767578125 4252.33203125
8813.6640625 4253.3134765625
8847.5205078125 4259.4365234375
8806.5068359375 4239.2255859375
8831.1904296875 4237.830078125
8913.208984375 4256.7265625
8853.7822265625 4229.2109375
8822.4189453125 4211.5166015625
8879.744140625 4221.84375
8826.3916015625 4196.556640625
8822.474609375 4187.1640625
8971.2529296875 4224.0322265625
8933.2783203125 4204.375
8912.0224609375 4188.619140625
8912.1416015625 4180.0498046875
8891.4375 4162.7275390625
8856.087890625 4144.865234375
8824.1044921875 4123.9365234375
8965.97265625 4158.0576171875
8841.6484375 4110.5439453125
8889.80078125 4115.7421875
8943.23046875 4121.37109375
8934.5634765625 4109.3427734375
8824.8115234375 4063.82543945312
8934.671875 4087.63989257812
8965.6923828125 4087.19702148438
8985.19140625 4081.94750976562
8946.7060546875 4060.33959960938
8796.3603515625 4002.40161132812
8879.5087890625 4017.09936523438
9006.4462890625 4048.18090820312
8911.443359375 4006.85961914062
8883.8154296875 3987.65454101562
8971.607421875 4005.12329101562
8874.2333984375 3963.86938476562
8965.5625 3983.09204101562
8892.990234375 3948.99047851562
8950.3720703125 3956.25756835938
8963.701171875 3949.92358398438
8987.7626953125 3948.28149414062
8974.4931640625 3933.07006835938
8941.302734375 3911.88305664062
8956.818359375 3906.04028320312
8863.6005859375 3864.95434570312
8930.767578125 3876.60131835938
8986.5341796875 3884.66577148438
8938.4619140625 3857.54370117188
8930.876953125 3845.14135742188
9006.8671875 3858.98120117188
8977.443359375 3839.20922851562
8977.009765625 3828.51782226562
8978.4931640625 3818.70727539062
8983.0546875 3808.35424804688
8960.0166015625 3791.41674804688
9017.1455078125 3798.00024414062
};
\addlegendentry{Measurements}
\addplot [semithick, black, opacity=0.9]
table {%
8446.3583984375 4186.412109375
8440.2421875 4189.740234375
8436.181640625 4194.4091796875
8433.2412109375 4199.525390625
8433.9765625 4205.873046875
8435.8359375 4211.888671875
8438.62890625 4217.7958984375
8443.046875 4222.03515625
8449.0859375 4226.3896484375
8456.685546875 4229.298828125
8464.9736328125 4231.0966796875
8474.19921875 4231.44921875
8483.17578125 4231.466796875
8493.650390625 4231.1533203125
8503.9384765625 4229.8369140625
8514.3525390625 4228.6328125
8525.451171875 4227.318359375
8537.484375 4226.896484375
8549.578125 4226.9208984375
8561.4853515625 4226.6103515625
8573.953125 4226.96875
8585.9208984375 4227.8837890625
8598.1982421875 4229.578125
8610.412109375 4232.1630859375
8622.5634765625 4234.748046875
8634.52734375 4237.3330078125
8647.1162109375 4239.4736328125
8659.0185546875 4241.61328125
8671.171875 4243.3076171875
8683.1376953125 4245.11328125
8695.1044921875 4246.361328125
8706.509765625 4248.0546875
8717.9140625 4249.859375
8728.8193359375 4251.9970703125
8739.5380859375 4253.6884765625
8749.6337890625 4255.37890625
8759.1689453125 4257.068359375
8768.7021484375 4259.203125
8778.298828125 4261.44921875
8787.646484375 4263.138671875
8797.0546875 4265.2734375
8806.52734375 4266.962890625
8816.0634765625 4267.873046875
8824.9169921875 4267.4453125
8832.8994140625 4265.7919921875
8840.076171875 4262.3544921875
8846.62890625 4258.58203125
8852.1884765625 4253.0263671875
8856.56640625 4246.244140625
8859.44921875 4239.0126953125
8861.6455078125 4231.890625
8863.6552734375 4224.658203125
8865.7275390625 4217.4248046875
8868.0498046875 4209.7470703125
8870.0615234375 4201.734375
8871.6982421875 4193.609375
8873.7099609375 4185.5966796875
8875.7841796875 4177.4736328125
8877.9833984375 4169.015625
8880.1826171875 4160.55859375
8882.1953125 4151.9892578125
8884.58203125 4143.0869140625
8887.0322265625 4134.1845703125
8889.2958984375 4124.8359375
8891.7451171875 4115.93359375
8894.5693359375 4106.5869140625
8897.3935546875 4097.685546875
8900.40625 4087.89331054688
8903.35546875 4078.54663085938
8906.4921875 4068.86596679688
8909.69140625 4059.52026367188
8912.828125 4050.17358398438
8915.90234375 4040.38159179688
8919.0390625 4031.03540039062
8922.3623046875 4021.68920898438
8925.6240234375 4012.00952148438
8928.63671875 4002.66284179688
8931.6484375 3993.31616210938
8934.91015625 3983.52514648438
8938.171875 3974.17895507812
8941.3701171875 3964.83276367188
8944.4453125 3955.04077148438
8947.64453125 3945.69506835938
8950.78125 3936.01440429688
8954.2294921875 3926.66870117188
8957.6787109375 3917.32299804688
8961.1279296875 3907.53198242188
8964.2646484375 3898.18579101562
8967.6513671875 3888.39477539062
8970.5380859375 3879.15893554688
8973.67578125 3869.36791992188
8976.9384765625 3859.57592773438
8980.51171875 3849.78588867188
8984.0234375 3840.44018554688
8988.0322265625 3831.09545898438
8992.1044921875 3821.86254882812
8996.17578125 3812.51831054688
8999.8115234375 3803.17309570312
9003.447265625 3793.82788085938
};
\addlegendentry{True Trajectory}
\addplot [semithick, blue, opacity=0.9]
table {%
8497.3720703125 4210.24853515625
8492.61328125 4216.01904296875
8487.943359375 4219.81640625
8480.1455078125 4222.65771484375
8475.4970703125 4226.52294921875
8481.1396484375 4233.69482421875
8476.1875 4235.52099609375
8466.8681640625 4235.185546875
8465.86328125 4236.39501953125
8457.943359375 4233.986328125
8473.6953125 4238.1552734375
8468.2158203125 4233.47607421875
8469.2802734375 4229.96826171875
8475.732421875 4227.46337890625
8481.09765625 4223.79736328125
8492.4208984375 4221.521484375
8501.1103515625 4218.53466796875
8497.1240234375 4211.4560546875
8500.708984375 4207.18310546875
8516.7744140625 4207.48876953125
8531.45703125 4208.06640625
8540.9462890625 4207.9208984375
8565.9150390625 4214.265625
8578.529296875 4216.56103515625
8600.099609375 4222.23779296875
8616.3203125 4226.7412109375
8644.0361328125 4234.72802734375
8662.9951171875 4239.0458984375
8680.8359375 4243.279296875
8693.146484375 4245.68408203125
8700.6123046875 4246.12744140625
8707.05078125 4245.919921875
8717.01953125 4247.16015625
8733.697265625 4251.0146484375
8739.654296875 4251.6572265625
8753.1064453125 4254.265625
8758.7841796875 4254.96142578125
8771.5625 4258.24609375
8782.37890625 4260.6328125
8795.7568359375 4264.08935546875
8804.1318359375 4266.1826171875
8811.470703125 4267.2451171875
8815.544921875 4266.6259765625
8822.2890625 4265.318359375
8830.4287109375 4263.30419921875
8840.2001953125 4261.4140625
8845.3896484375 4256.83984375
8853.123046875 4251.47119140625
8866.2861328125 4247.7119140625
8872.28125 4240.724609375
8876.783203125 4233.1611328125
8885.982421875 4227.158203125
8890.0732421875 4219.38525390625
8893.86328125 4211.609375
8908.4013671875 4207.013671875
8918.45703125 4201.14306640625
8926.12890625 4194.29443359375
8933.4853515625 4187.44482421875
8938.6884765625 4179.1708984375
8941.1376953125 4171.03076171875
8941.169921875 4161.48974609375
8951.5478515625 4154.880859375
8951.6953125 4145.328125
8955.291015625 4136.62939453125
8962.494140625 4128.56884765625
8968.1767578125 4120.2490234375
8965.3544921875 4108.96337890625
8970.921875 4099.9775390625
8977.8876953125 4091.59204101562
8985.5986328125 4083.13037109375
8989.91015625 4073.86254882812
8982.8984375 4061.35668945312
8982.6640625 4050.212890625
8990.279296875 4042.48315429688
8990.9970703125 4031.912109375
8989.408203125 4020.72485351562
8993.927734375 4011.66259765625
8990.953125 4000.23486328125
8994.31640625 3991.13623046875
8992.2666015625 3979.99096679688
8994.4033203125 3970.01538085938
8997.185546875 3960.27954101562
9001.3095703125 3951.52416992188
9004.1904296875 3942.04296875
9004.296875 3931.53833007812
9005.4248046875 3921.27563476562
8999.470703125 3908.58837890625
8998.5068359375 3897.70556640625
9001.150390625 3888.38500976562
9000.5078125 3877.39135742188
8999.0185546875 3866.39965820312
9002.9619140625 3857.32470703125
9004.5576171875 3847.42065429688
9005.935546875 3837.40747070312
9007.2294921875 3827.43896484375
9008.779296875 3817.0419921875
9008.357421875 3806.41967773438
};
\addlegendentry{Opt. IMM}
\addplot [semithick, green, opacity=0.9]
table {%
8497.2822265625 4210.21484375
8492.646484375 4216.0478515625
8496.00390625 4222.2978515625
8490.4931640625 4225.8447265625
8489.34375 4231.1064453125
8494.70703125 4237.96435546875
8491.92578125 4239.08935546875
8484.1845703125 4238.595703125
8482.1435546875 4237.916015625
8479.0166015625 4236.5224609375
8481.154296875 4233.431640625
8484.181640625 4231.6689453125
8488.9697265625 4229.42822265625
8493.9990234375 4226.74658203125
8497.9140625 4223.03759765625
8503.1953125 4218.89501953125
8507.837890625 4216.3583984375
8506.7080078125 4212.3056640625
8508.875 4208.1748046875
8514.970703125 4205.5546875
8519.5419921875 4204.12890625
8520.57421875 4202.84423828125
8529.857421875 4204.63525390625
8537.2197265625 4205.353515625
8548.087890625 4207.08349609375
8559.19921875 4210.48291015625
8580.0185546875 4214.7734375
8601.013671875 4218.56787109375
8620.6298828125 4223.95166015625
8637.1787109375 4227.85107421875
8651.7060546875 4230.0703125
8665.228515625 4231.78857421875
8679.7919921875 4235.15625
8698.837890625 4240.53125
8710.8564453125 4243.5009765625
8724.7646484375 4245.36279296875
8733.8798828125 4248.11669921875
8743.9541015625 4250.81494140625
8754.7041015625 4252.5166015625
8765.3623046875 4255.4130859375
8773.3251953125 4257.57373046875
8779.8291015625 4257.060546875
8783.2783203125 4255.26123046875
8781.5986328125 4249.11474609375
8784.412109375 4244.44873046875
8788.064453125 4241.13330078125
8788.8662109375 4234.21923828125
8794.685546875 4226.90234375
8802.0595703125 4222.794921875
8806.173828125 4214.8935546875
8810.90625 4208.4462890625
8816.8291015625 4202.69287109375
8819.4501953125 4194.96435546875
8823.330078125 4187.97509765625
8833.3505859375 4182.48779296875
8841.0048828125 4176.50732421875
8847.8271484375 4169.322265625
8854.6611328125 4162.79541015625
8860.419921875 4153.75341796875
8863.6318359375 4147.72607421875
8864.4755859375 4137.39990234375
8872.515625 4130.05322265625
8872.3359375 4120.873046875
8876.0517578125 4112.03369140625
8882.6591796875 4103.41357421875
8888.29296875 4095.68627929688
8888.224609375 4084.71508789062
8894.1064453125 4075.72631835938
8901.1708984375 4068.07763671875
8909.427734375 4059.54248046875
8915.2197265625 4051.27221679688
8912.4677734375 4040.32299804688
8913.203125 4028.39697265625
8920.669921875 4022.6015625
8922.1962890625 4011.0009765625
8922.8291015625 4000.56860351562
8927.1669921875 3992.04541015625
8926.6796875 3981.212890625
8930.6806640625 3972.88061523438
8931.654296875 3961.97827148438
8935.4091796875 3952.22290039062
8939.94140625 3943.18627929688
8945.4345703125 3935.76904296875
8949.876953125 3926.15209960938
8953.29296875 3916.33911132812
8956.4111328125 3906.57373046875
8953.0625 3894.26440429688
8953.2138671875 3884.10620117188
8958.328125 3876.32690429688
8959.845703125 3864.86596679688
8960.5966796875 3854.8720703125
8965.55078125 3846.541015625
8969.0654296875 3836.99853515625
8973.8447265625 3827.91333007812
8978.0830078125 3818.94946289062
8982.708984375 3808.68676757812
8985.4375 3799.728515625
};
\addlegendentry{MKF}
\addplot [semithick, red, opacity=0.9]
table {%
8497.3759765625 4210.25146484375
8491.478515625 4215.88623046875
8483.6953125 4218.00732421875
8466.90625 4217.5791015625
8459.3271484375 4220.3427734375
8482.0126953125 4233.47900390625
8476.47265625 4233.541015625
8456.794921875 4228.9091796875
8451.283203125 4226.82177734375
8433.146484375 4219.63525390625
8458.44140625 4225.2109375
8473.861328125 4227.8701171875
8478.2783203125 4225.30615234375
8492.7763671875 4225.90478515625
8511.40234375 4227.54638671875
8542.3251953125 4232.6201171875
8566.2255859375 4236.86083984375
8546.9765625 4226.5625
8539.212890625 4218.9453125
8565.998046875 4223.48974609375
8587.5615234375 4227.9765625
8583.353515625 4224.93408203125
8623.6708984375 4237.56884765625
8613.908203125 4232.05322265625
8643.21875 4239.86376953125
8646.9755859375 4240.619140625
8705.8349609375 4257.52197265625
8736.9169921875 4264.103515625
8757.4033203125 4269.40625
8758.2138671875 4268.0244140625
8746.798828125 4261.58642578125
8734.6318359375 4254.6572265625
8739.060546875 4254.51806640625
8772.0048828125 4264.3125
8757.9990234375 4258.84423828125
8779.5830078125 4262.9638671875
8767.2568359375 4258.77001953125
8788.6767578125 4265.11474609375
8799.591796875 4266.75634765625
8821.017578125 4272.94091796875
8818.298828125 4271.7646484375
8816.66015625 4268.6142578125
8802.4248046875 4261.14208984375
8801.216796875 4255.09423828125
8807.0126953125 4251.19873046875
8823.0673828125 4251.7666015625
8820.951171875 4243.88134765625
8827.99609375 4236.86572265625
8861.4130859375 4240.8671875
8865.4921875 4232.9365234375
8856.123046875 4222.0908203125
8868.021484375 4218.2197265625
8857.6474609375 4206.4365234375
8847.12109375 4194.955078125
8891.474609375 4200.1640625
8912.771484375 4198.17236328125
8920.7119140625 4191.28564453125
8924.9599609375 4183.935546875
8919.732421875 4171.4931640625
8900.4912109375 4158.63623046875
8874.6064453125 4140.0966796875
8904.06640625 4139.32421875
8883.5859375 4123.79443359375
8884.662109375 4114.1650390625
8904.623046875 4109.59228515625
8917.4560546875 4104.06787109375
8887.7705078125 4084.00463867188
8903.0595703125 4077.89526367188
8926.5537109375 4075.28686523438
8951.6455078125 4071.8857421875
8956.5771484375 4063.31762695312
8904.658203125 4037.53979492188
8894.3671875 4021.86767578125
8931.3330078125 4025.41552734375
8925.2646484375 4011.310546875
8910.75 3996.13037109375
8931.1943359375 3992.71411132812
8912.3857421875 3976.05444335938
8929.791015625 3972.01928710938
8917.7763671875 3956.912109375
8928.97265625 3949.578125
8942.75390625 3943.439453125
8961.708984375 3940.27001953125
8971.283203125 3932.1416015625
8965.6953125 3919.47900390625
8965.7451171875 3908.80981445312
8931.2578125 3886.84594726562
8927.71875 3875.6025390625
8946.095703125 3872.1171875
8943.681640625 3859.26782226562
8938.6298828125 3847.54028320312
8962.568359375 3845.32299804688
8970.70703125 3837.13549804688
8976.3896484375 3828.31616210938
8980.5576171875 3819.34423828125
8984.962890625 3808.98828125
8978.853515625 3797.29467773438
};
\addlegendentry{EKF}
\addplot [semithick, purple, opacity=0.9]
table {%
8364.27592189224 4145.09703212395
8462.80166707664 4185.14917980974
8453.59207134421 4187.26032868742
8459.25837797 4196.73991613574
8324.27411929626 4152.16056567656
8393.07013043387 4183.26936283929
8475.39766204424 4218.0366896947
8395.00608368101 4191.15145718623
8400.94546336573 4195.20333936274
8475.88877639628 4224.29587515319
8497.7027214938 4231.56196102239
8536.58218413914 4240.02376393818
8412.61412116077 4193.35139290592
8465.87319451304 4207.88620685195
8552.11466477847 4234.70958109106
8468.76235463069 4199.98086501219
8494.39767675816 4203.35120186145
8532.10266303188 4212.0416120026
8549.34736031943 4214.33225695745
8496.23902043491 4189.99482585128
8575.08326281996 4213.64391301083
8593.35801439166 4216.9687275636
8438.71435235073 4160.41164806832
8644.27154876478 4230.10963156799
8548.51424358479 4196.74046619019
8510.08010047877 4179.69277181616
8636.15865285157 4222.91017219226
8565.21653349761 4196.55456862139
8533.3776534532 4183.26896386191
8655.73616351294 4221.95192826261
8596.48126573321 4199.28124167785
8605.33921720834 4199.99980740328
8722.92767251422 4237.87331371545
8653.55399514036 4214.06376746725
8670.55509666749 4217.53730943024
8701.77597480801 4226.1808011381
8667.81969759816 4212.83134502542
8744.61499246524 4237.08633295087
8756.44631736354 4240.29106502706
8778.02011550737 4245.17966539132
8735.55462703207 4231.00174783157
8770.23041184929 4241.65083033608
8817.54574154609 4252.79128916837
8840.80810448134 4257.36007267406
8808.58642097865 4243.1728477709
8847.54783065755 4250.15371769885
8836.21081930474 4241.4780432878
8833.95901488562 4233.03965617635
8782.39072529181 4208.20806745479
8885.87264815296 4231.97672954545
8833.62256237676 4209.01364735923
8932.18996214674 4230.22985846705
8768.17315919136 4171.97274908563
8833.72305280475 4185.10400917298
8856.88027624454 4181.08240090129
8799.98415774202 4155.92854363708
8778.12251699833 4141.07664997443
8743.01487568346 4121.46732765537
8775.53995523844 4123.09104606081
8812.38885109491 4124.82039699988
8891.92221341359 4140.98426750716
8898.53691388526 4131.93189939184
8845.48063387156 4106.86429288547
8914.55822210622 4118.14376591355
8871.44748605306 4095.36566872334
8790.74309760694 4059.99897821654
8789.6209977071 4047.9890960702
8818.89297949089 4047.57152470281
8868.48821612274 4052.31204152583
8800.95343633619 4021.49331629163
8922.72373165875 4048.1042973363
8863.19674724984 4020.38146001155
8883.09931846654 4015.51490428718
8919.56248190441 4016.63280263253
8856.8547004488 3986.14928717333
8807.86888175294 3959.41484356088
8881.82138698097 3971.79332502345
8955.59095089699 3986.16626775495
8830.45951433861 3936.79285602651
8916.88891033413 3953.18018649058
8843.04781128982 3919.22685466761
8854.35992522648 3912.10168830414
8837.66181339988 3895.78793698228
8989.80477864114 3932.5785121208
8853.3050509701 3880.15570849195
8888.21454475742 3881.97685998529
8887.22218000851 3869.71468980332
8938.74880418273 3876.00361527363
8918.72569526895 3858.93435079319
9065.97721422887 3891.82107079607
9012.59146165949 3866.48394769541
8964.32516380602 3839.75026183197
9005.92184432427 3842.0858611107
9014.50605630689 3835.71012062847
8961.95417013715 3807.58506291385
8939.11004168168 3790.97951596167
8944.45800308092 3782.36099942255
8975.23232061065 3780.81433013213
8861.19031033535 3733.54167963357
9010.03842217345 3771.48202920653
};
\addlegendentry{GP}
\end{axis}

\end{tikzpicture}